%% file: main.tex
\documentclass{article}

\usepackage[preprint]{neurips_2026}

\usepackage[utf8]{inputenc} 
\usepackage[T1]{fontenc}    
\usepackage{hyperref}       
\usepackage{url}            
\usepackage{booktabs}       
\usepackage{amsfonts}       
\usepackage{nicefrac}       
\usepackage{microtype}      
\usepackage{xcolor}         
\usepackage{enumitem}
\usepackage{graphicx}
\usepackage{amsmath, amssymb}
\usepackage{booktabs, multirow, adjustbox}
\usepackage[table]{xcolor}
\usepackage{multirow}
\usepackage{graphicx}
\usepackage{mdframed}
\usepackage{xcolor}
\usepackage{fvextra}
\usepackage{mathtools}
\usepackage{wrapfig}
\usepackage{xspace}
\usepackage{pifont}
\usepackage{soul}
\usepackage[most]{tcolorbox}
\usepackage{algorithm}    
\usepackage{algpseudocode} 

\definecolor{eggshell}{rgb}{0.94, 0.92, 0.84}
\definecolor{floralwhite}{rgb}{1.0, 0.98, 0.94}
\definecolor{lavenderblue}{rgb}{0.8, 0.8, 1.0}
\definecolor{lavendermist}{rgb}{0.9, 0.9, 0.98}

\newmdenv[linewidth=0.5pt, linecolor=black, backgroundcolor=gray!5,
          innertopmargin=6pt, innerbottommargin=6pt,
          innerleftmargin=6pt, innerrightmargin=6pt,
          skipabove=4pt, skipbelow=4pt]{databox}
\usepackage[table]{xcolor}
\usepackage{array}

\definecolor{headshade}{HTML}{F8F4FF}
\definecolor{orgshade}{HTML}{F8F4FF}
\definecolor{best}{HTML}{D9EAD3}
\definecolor{second}{HTML}{FFF2CC}

\newcommand{\best}[1]{\cellcolor{best}#1}
\newcommand{\second}[1]{\cellcolor{second}#1}

\DeclareRobustCommand{\caporg}{%
  \begingroup
  \setlength{\fboxsep}{1pt}%
  \colorbox{orgshade}{\strut cell color}%
  \endgroup
}

\DeclareRobustCommand{\capbest}{%
  \begingroup
  \setlength{\fboxsep}{1pt}%
  \colorbox{best}{\strut best}%
  \endgroup
}

\DeclareRobustCommand{\capsecond}{%
  \begingroup
  \setlength{\fboxsep}{1pt}%
  \colorbox{second}{\strut second-best}%
  \endgroup
}
\newcolumntype{O}{>{\columncolor{orgshade}}c}
\newtcolorbox{highlightbox}[1]{
  enhanced,
  colback=blue!5!white,
  colframe=blue!50!white,
  coltitle=white,
  colbacktitle=blue!40!white,
  fonttitle=\bfseries,
  title=#1,
  boxrule=0pt,
  toprule=0pt,
  bottomrule=0pt,
  leftrule=0pt,
  rightrule=0pt,
  arc=3pt,
  left=3pt,
  right=3pt,
  top=2pt,
  bottom=2pt,
  attach boxed title to top,
  boxed title style={
    colback=blue!40!white,
    colframe=blue!40!white,
    arc=0pt,
    outer arc=0pt,
    boxrule=0pt,
  },
  title filled,
  before skip=5pt,
  after skip=5pt,
}

\newcommand{\system}{\textsc{Camel}\xspace}

\title{The Trap of Trajectory: Towards Understanding and Mitigating Spurious Correlations in Agentic Memory}

\setlist[itemize]{leftmargin=15pt}

\author{
\textbf{Luoxi Tang}$^{1}$ \quad
\textbf{Rupali Rajendra Vaje}$^{1}$ \quad
\textbf{Yuqiao Meng}$^{1}$ \quad
\textbf{Sakshi Sunil Narkar}$^{1}$ \\
\textbf{Weicheng Ma}$^{2}$ \quad
\textbf{Zeyu Ding}$^{1}$ \quad
\textbf{Dazheng Zhang}$^{3}$ \quad
\textbf{Zhaohan Xi}$^{1}$ \\
\vspace{0.5em} \\
$^{1}$Binghamton University, State University of New York, Binghamton, NY, USA \\
$^{2}$Oakland University, Rochester, MI, USA \\
$^{3}$University of Pennsylvania, Philadelphia, PA, USA \\
Corresponding to: Zhaohan Xi \texttt{<zxi@binghamton.edu>}
}

\begin{document}

\maketitle

\input{00abs}
\input{01intro}
\input{02literature}
\input{03bench}

\input{04method-0503}
\input{05expt}

\bibliographystyle{plain}
\bibliography{reference,craft-lab}
\label{reference}

\appendix
\input{appendix/Validation}
\input{appendix/benchmark}
\input{appendix/calibrate-embed}
\input{appendix/calibrate-graph}

\input{appendix/Additional_Experimental_Results_and_Analysis}
\input{appendix/FailureMode}

\end{document}

%% file: 00abs.tex
\begin{abstract}

Agentic memory lets LLMs persist information beyond a single context window and reuse it in later decisions, but it also opens a new vulnerability: spurious correlations, where retrieved memory carries mis-correlated evidence and propagates erroneous reasoning into downstream decisions. Despite the widespread use of agentic memory, this risk remains largely underexplored. We address it from two aspects. First, we benchmark several canonical types of spurious patterns identified through causal structure recovered from reasoning trajectories. Diagnosing agentic memory systems on this benchmark reveals that memory improves reasoning on clean inputs but amplifies reliance on spurious patterns when present. Subsequently, we propose \system, a plug-and-play calibration that operates on diverse memory architectures at write and retrieval time. \system consistently reduces reliance on spurious patterns across all three types while preserving or improving performance on clean inputs, and stays robust under adaptive attacks targeting the calibration. Overall, \system offers a principled and lightweight solution toward more reliable agentic memory deployment.
\end{abstract}

%% file: 01intro.tex
\section{Introduction}

Agentic memory enables LLM agents to persist information beyond 
a single context window, storing experiences from past 
interactions and retrieving them to inform later decisions 
\citep{packer2023memgpt, park2023generative, 
shinn2023reflexion}. By reusing relevant prior experience, agents 
can act more consistently across tasks and avoid recomputing 
what they have already worked out \citep{anokhin2024arigraph, 
chhikara2025mem0, xu2025mem, zhang2025g}, making 
memory a critical component of modern agentic systems 
\citep{wu2025human, zhang2025survey}.

However, a long-standing concern for machine learning models is their 
tendency to pick up \textit{spurious patterns}, i.e., statistical indicators 
that appear associative in observed features but do not reflect the 
underlying causal relationship~\citep{du2023shortcut, geirhos2020shortcut, 
tang2023large, ye2024spurious}. Such patterns typically arise from hidden 
confounders or conditioning on colliders between learned features and the 
target outcome~\citep{neuberg2003causality, pearl2018book, 
spirtes2000causation, westreich2012berkson}, leading models to follow 
shortcuts with familiar evidence but fail when underlying conditions 
shift~\citep{arjovsky2019invariant, geirhos2020shortcut, 
sagawa2019distributionally}.

Agentic memory presents unique risks with spurious patterns in two ways: (i) \textit{Persistence}: a spurious correlation written into memory 
survives beyond the context window and remains available for retrieval on 
every subsequent turn, whereas in standard reasoning the same cue 
dissipates when the response ends and is bounded by context 
length~\citep{geng2026causalt5k,liu2024lost,sun2025scaling}. (ii) \textit{Feedback}: the agent's actions generate 
the next trajectory that is written back into memory, so a 
spurious correlation acted on once can shape the very evidence used to 
justify acting on it again~\citep{ma2026implicit,oh2025understanding}. Together, these properties turn what 
would be a volatile error into a self-reinforcing bias in the 
agent's decision-making.

Despite the critical role of memory, how spurious correlations affect agentic memory remains largely underexplored. We hence organize our studies with two questions: \textbf{RQ1:} How do spurious correlations influence memory use in LLMs? \textbf{RQ2:} How can agentic memory be calibrated to reduce influences of spurious correlations?

\textbf{This work.} To answer these questions, this work makes the following contributions:

\textbf{I. Benchmarking and diagnosing spurious correlations in agentic memory}

Existing agentic-memory benchmarks evaluate retrieval accuracy or reasoning quality but do not account for spurious correlations \citep{ maharana2024evaluating,shridhar2020alfworld,wang2022scienceworld,yao2022webshop}. To address RQ1, we fill this gap with a new benchmark spanning four datasets. For each dataset, we recover a causal structure over the agent's observations and memory states, then identify variable pairs that are statistically associated yet lack any directed causal path between them. We target the discovery of three canonical types of spurious patterns from datasets: \textit{explicit confounding}, \textit{unmeasured confounding}, and \textit{collider bias}.

Using this benchmark, we evaluate representative memory systems and arrive at three findings. First, memory consistently improves reasoning performance, but it also makes agents more likely to act on spurious patterns whenever such patterns appear in memory. Second, both architectures share a dominant weakness on unmeasured confounding, with graph-based systems showing additional vulnerability when repeated co-occurrence accumulates into heavier edge weights. Third, the effect of a single spurious retrieval propagates into subsequent decisions along different trajectories, motivating a stepwise calibration approach that intervenes agentic memory following the reasoning trajectory.

\textbf{II. Calibrating memory representations to mitigate spurious correlations} 

To address RQ2, we propose \system (\underline{CA}usality-informed \underline{ME}mory ca\underline{L}ibration), a plug-and-play calibration that attaches to agentic memory architectures without additional LLM calls or fine-tuning. \system intervenes at the two stages where spurious structure enters and propagates: at \textit{write time}, it removes the shared context-driven component from each entry's representation, retaining only content-specific signal; at \textit{retrieval time}, it tests whether each candidate's relevance is stable under perturbations along non-causal directions of the representation space, demoting candidates whose ranking depends on shared spurious structure rather than causal content.

Through evaluations, \system consistently reduces reliance on spurious features for all three correlation types while preserving (and often improving) reasoning performance on clean inputs. The improvements hold across LLM backbones and remain robust under adaptive attacks that explicitly target the calibration mechanism, demonstrating that representation-level calibration can address spurious correlations in agentic memory in a principled and practical way. Code is released at \url{https://anonymous.4open.science/r/Spurious_Correlation-A830}.

%% file: 02literature.tex
\section{Related Work}

\textbf{LLM-as-agent.} The deployment of LLMs as autonomous agents has extended their role beyond text generation toward reasoning, planning, and acting in complex environments \citep{ schick2023toolformer,yang2025eligibility,yao2022react}, with multi-agent architectures further enabling collaboration and debate among specialized agents \citep{chan2024chateval,wu2024autogen}. Building on this paradigm, recent work has applied agentic LLMs to high-stakes domains, including clinical diagnostics and digital health \citep{meng2026small, more2026theramind,zhou2024zodiac}, cybersecurity threat intelligence and blue-teaming \citep{liu2025cylens,meng2025benchmarking,tang2025polar}, and telecom resilience via causal digital twins \citep{sriram2026adversarial}. Methodological advances have addressed debate collapse through uncertainty-driven optimization \citep{tang2026value} and probed agentic reasoning capabilities such as counterfactual inference \citep{yang2025eligibility}. At the same time, the expanded action space of agentic LLMs introduces new vulnerabilities \citep{meng2025uncovering,xi2025all}, motivating continued research into robust and trustworthy agent design.

\textbf{Agentic memory systems for LLMs.}
LLM agents use memory systems to store past interactions and 
reuse them in later turns~\citep{packer2023memgpt, 
park2023generative, shinn2023reflexion, zhong2024memorybank}. 
Existing systems mainly follow two designs. Embedding-based 
memories, such as Mem0~\citep{chhikara2025mem0} and 
A-Mem~\citep{xu2025mem}, retrieve entries by vector similarity. 
Graph-based memories, such as G-Memory~\citep{zhang2025g} and 
AriGraph~\citep{anokhin2024arigraph}, organize memories with 
explicit relations and retrieve them through graph structure. 
Despite these differences, both types judge relevance by 
association alone. Neither checks whether a retrieved memory 
has a causal relation to the query, which is the gap our work 
addresses.

\textbf{Causal reasoning in NLP.} 
A growing body of work brings causal tools into NLP to make 
models more robust to spurious 
correlations~\citep{peters2016causal, 
scholkopf2022causality,scholkopf2021toward,  xu2020causality}. Representative methods 
include invariant risk 
minimization~\citep{arjovsky2019invariant} and causal 
intervention on chain-of-thought~\citep{wu2024decot}. These 
efforts treat the language model as the unit of analysis. 
The memory layer remains outside their scope. 

\textbf{Safety and robustness of agentic memory.} 
Two complementary lines study how agentic memory can fail. 
The first line considers adversarial attacks. AgentPoison~\citep{chen2024agentpoison} 
and BadAgent~\citep{wang2024badagent} inject crafted entries 
that trigger targeted misbehavior~\citep{dong2025practical, 
sunil2026memory, wei2025memguard}, exposing memory as a 
vulnerable surface. The second line studies shortcut 
learning~\citep{du2023shortcut, geirhos2020shortcut, 
sagawa2019distributionally, tang2023large, ye2024spurious}, 
where models exploit spurious features in biased training 
data. These studies typically use static, single-turn settings. 
Together they characterize externally injected and statically 
observed failures.

%% file: 03bench.tex
\section{Spurious Correlation in Agentic Memory: Definitions and Benchmarking}
\label{sec:def}

This section studies how spurious correlations can mislead agentic memory systems. Since existing benchmarks do not account for spuriousness, we first categorize spurious patterns (\S\ref{ssec:background}) and then build a spuriousness-focused benchmark from datasets commonly used to evaluate memory-enabled agentic systems (\S\ref{ssec:curate}). We further evaluate agentic memory (\S\ref{ssec:eval-spur}), revealing how each type of spurious correlation degrades agentic reasoning and motivating the mitigative solutions (\S\ref{sec:method}).

\begin{figure}[t]
\centering
\includegraphics[width=0.97\linewidth]{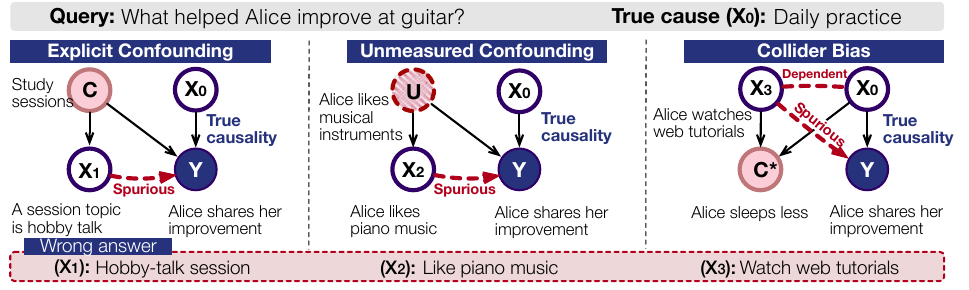}
\caption{Three types of spurious correlations that mislead agentic memory: \textbf{(left)} observable confounder $C$ creates an $X_1$ (memory cue)-$X_0$ (true causality) association; \textbf{(middle)} same structure but the confounder $U$ is latent.\textbf{(right)} The entry content $X_3$ and the true cause $X_0$ are independent, but both are parents of a collider $C^\star$. Conditioning on
$C^\star$ makes
$X_3$ and $X_0$ conditionally dependent, so $X_3$ becomes spuriously
correlated to $Y$.}
\vspace{-5pt}
\label{fig:spur-type}
\end{figure}

\subsection{Background: From Causal Structure to Spurious Correlations}
\label{ssec:background}

A spurious correlation derives from the statistical association that does not reflect any real \textit{causal} relationship. This part first defines causal structure in graph format, and then introduces three representative spurious types.

\textbf{Causal graphs.}
The causal structure is typically represented as a directed acyclic graph (DAG) $\mathcal{G} = (V, E)$, where each node $v \in V$ is a variable (e.g., a past action, a memory entry, or an agent decision), and a directed edge $u \to v$ means that $u$ has a direct causal effect on $v$. When two variables are statistically associated in data, that association may or may not reflect a true edge in $\mathcal{G}$. If it does not, the association is spurious.


Three canonical patterns in causal graphs that give rise to spurious correlations \citep{haig2003spurious}. Each has a distinct shape and requires a different approach to detect and correct:

\textbf{Type 1 ($\texttt{T}_1$): Explicit confounding}
occurs when a third variable $C$ is a common cause of both a memory-derived indicator $X$ (e.g., a retrieved memory entry) and the agent's action or prediction $Y$, and $C$ is \emph{observable} in the data. Because $C$ drives both the contents written to / retrieved from memory and the correct action, $X$ and $Y$ become spuriously correlated even when the retrieved memory carries no causal information for $Y$ beyond what $C$ already provides. The causal structure is shown in Figure \ref{fig:spur-type}-(left).

\begin{highlightbox}{\small Example: Explicit Confounding in ALFWorld \citep{shridhar2020alfworld} (More examples in App.\ref{app:datasets})}

An agent is solving household tasks. The task category $C$ (e.g., ``heat-and-place'') confounds memory and action: it causes past trajectories mentioning ``microwave'' to be retrieved into context ($X$), and it independently dictates the correct action ``go to microwave'' ($Y$). The agent then learns a shortcut whenever ``microwave'' appears in retrieved memory, output ``go to microwave,'' even though the memory content has no causal relationship on the action. 
\end{highlightbox}

\textbf{Type 2 ($\texttt{T}_2$): Unmeasured confounding}
 is structurally the same as explicit confounding, but the shared cause $U$ is \emph{never recorded} in the dataset (as shown in Figure \ref{fig:spur-type}-(middle)). In agentic memory, $U$ represents a latent user preference or an implicit environmental context that shapes both what gets stored in memory and what action is eventually taken.

\textbf{Type 3 (T$_3$): Collider bias} arises when a variable $C^\star$ has two
or more upstream (parent) causes pointing into it, and the data is observed only after
\emph{conditioning} on $C^\star$.
Although the parents of $C^\star$ are independent, conditioning on
$C^\star$ renders them \emph{conditionally dependent}, opening a non-causal
path between them. In agentic memory, the collider presents when an entry is stored and one of its parent reaches a target
outcome $Y$. This entry will be conditioned as collider $C^\star$ that induces a spurious $X$-$Y$ correlation when its another parent $X$ has no causal effect on $Y$ (Figure~\ref{fig:spur-type}-(right)).

\subsection{Curating Algorithm to Discover Spurious Correlations}
\label{ssec:curate}

We discover spurious correlations by curating reasoning trajectories with three stages, as summarized in Algorithm~\ref{alg:curate}. \textbf{Stage-1:} We first define a causal variable set over each trajectory by treating the user query $Q_s$, memory state $M_s$, reasoning $R_s$, and response $A_s$ at every step $s$ as nodes, together with trajectory-level context variables $C$ (e.g., task category, history length). A step ends whenever the agent reads from or writes to memory, so one user query can span several steps if the agent uses memory more than once while reasoning. \textbf{Stage-2:} We then build a causal DAG $\hat{\mathcal{G}}$ directly from this structure: time order sets edge direction (a variable at step $s$ cannot cause one at any earlier step $s' < s$), each step's recorded action and observation give the within-step edges, and trajectory-level metadata adds edges from context variables to the steps they affect. \textbf{Stage-3:} We finally scan $\hat{\mathcal{G}}$ for pairs $(X, Y)$ that are statistically associated yet have no directed path between them, and classify each by structural signature:  a shared parent $C$ marks explicit confounding ($\texttt{T}_1$); an association with no recorded mediating path marks unmeasured confounding ($\texttt{T}_2$); and a shared variable $C^{\star}$, whose parents contains true causal nodes of the target outcome, marks collider bias ($\texttt{T}_3$). We detail validation of discovered spurious correlations and per-dataset implementation in App.\ref{app:validation} and~\ref{app:benchmark}, respectively.

\begin{algorithm}[t]
\caption{\small Curating Reasoning Trajectories to Discover Spurious Correlations}
\label{alg:curate}
\footnotesize
\begin{algorithmic}[1]
\Require Trajectory dataset $\mathcal{D} = \{(Q_s, M_s, R_s, A_s)_{s=1}^{S}, C\}$; significance level $\alpha$
\Ensure Labeled spurious correlations $\mathcal{S}_{\text{spur}} = \{(X, Y, \texttt{type})\}$
\State \textbf{// Stage A: Define causal variables}
\State Segment each trajectory into steps $s = 1, \ldots, S$ delimited by memory read/write events
\State $\mathcal{V} \gets \{Q_s, M_s, R_s, A_s\}_{s=1}^{S} \cup \{C\}$
\State \textbf{// Stage B: Construct causal DAG from trajectory structure}
\State $\hat{\mathcal{G}} \gets \emptyset$
\State Add edges from temporal order: forbid any edge from step $s'$ to $s < s'$
\State Add within-step edges from recorded action–observation pairs
\State Add edges from each context variable in $C$ to the steps it governs
\State \textbf{// Stage C: Pinpoint spurious correlations}
\State $\mathcal{S}_{\text{spur}} \gets \emptyset$
\ForAll{pairs $(X, Y) \in \mathcal{V} \times \mathcal{V}$ with $X \not\!\perp\!\!\!\perp Y$ at level $\alpha$ and no directed path in $\hat{\mathcal{G}}$}
    \If{$\exists\, C \in \mathcal{V}$ with $C \to X$ and $C \to Y$ in $\hat{\mathcal{G}}$}
        \State $\mathcal{S}_{\text{spur}} \gets \mathcal{S}_{\text{spur}} \cup \{(X, Y, \texttt{T}_1)\}$ \Comment{explicit confounding}
    \ElsIf{no recorded mediating path connects $X$ and $Y$ in $\hat{\mathcal{G}}$}
        \State $\mathcal{S}_{\text{spur}} \gets \mathcal{S}_{\text{spur}} \cup \{(X, Y, \texttt{T}_2)\}$ \Comment{unmeasured confounding}
    \ElsIf{$\exists\, C^{\star} \in \mathcal{V}$ with
$X, X_0 \to C^\star$ and $X_0 \to Y$ in $\hat{\mathcal{G}}$}
        \State $\mathcal{S}_{\text{spur}} \gets \mathcal{S}_{\text{spur}} \cup \{(X, Y, \texttt{T}_3)\}$ \Comment{collider bias}
    \EndIf
\EndFor
\State \textbf{// Validation: confirm each candidate via conditional independence tests (App.\ref{app:validation})}
\State \Return $\mathcal{S}_{\text{spur}}$
\end{algorithmic}
\end{algorithm}

\subsection{Evaluation of Spurious Correlations in Agentic Memory}
\label{ssec:eval-spur}


\textbf{Benchmark.} To build a benchmark with queries that trigger spurious correlations, we apply Algo.~\ref{alg:curate} to four datasets: \textbf{ALFWorld}~\citep{shridhar2020alfworld} for embodied household tasks, \textbf{ScienceWorld}~\citep{wang2022scienceworld} for interactive science experiments, \textbf{LoCoMo}~\citep{maharana2024evaluating} for long-term dialogue, and \textbf{WebShop}~\citep{yao2022webshop} for goal-directed online shopping. App.\ref{app:alfworld}-\ref{app:webshop} details how each dataset is curated to trigger spurious correlations in agentic memory, along with concrete examples and overall data statistics (App.\ref{app:datasets}, Table~\ref{tab:data_stats}).

\textbf{Agentic memory.} We evaluate four representative memory systems spanning two groups. \textit{Embedding-based (vector-based) memory}: A-Mem~\citep{xu2025mem} and Mem0~\citep{chhikara2025mem0}. \textit{Graph-based (structural) memory}: G-Memory~\citep{zhang2025g} and AriGraph~\citep{anokhin2024arigraph}. A No-Memory baseline (the raw LLM without memory) is included as a lower bound. Each memory system is paired with three LLMs: Llama-4-Maverick-17B (L4), Qwen3.6-27B (Qw), and Mistral-Small-3.2-24B (Ms).

\textbf{Results and insights.} Table \ref{tab:rq1_main} reports task accuracy (Org., higher is better) and spurious reasoning ratios ($\texttt{T}_1$/$\texttt{T}_2$/$\texttt{T}_3$, lower is better) across memory systems and LLM backbones. We have three findings, wherein each motivates a corresponding aspect of the calibration design in §\ref{sec:method}.

\input{tables/rq1_main_results}

\textbf{Finding \ding{182}: Memory helps on clean inputs but amplifies spurious patterns when present.}
Every memory system clearly improves original accuracy over the no-memory baseline,
confirming that memory is a strong augmentation for reasoning. However, the same agentic memory systems
show substantially higher spurious reasoning ratios across all of $\texttt{T}_1$,
$\texttt{T}_2$, and $\texttt{T}_3$. The reason is that a system retrieving relevant entries
well also retrieves spurious entries well, since both go through the same matching mechanism.
The benefit and the risk of memory cannot be separated by retrieving less or filtering more.
A fix must instead distinguish causal from non-causal content inside the memory representation,
motivating a calibration that acts on what is stored rather than how much is retrieved.

\begin{wrapfigure}{r}{0.42\textwidth}
\vspace{-8pt}
\centering
\includegraphics[width=0.42\textwidth]{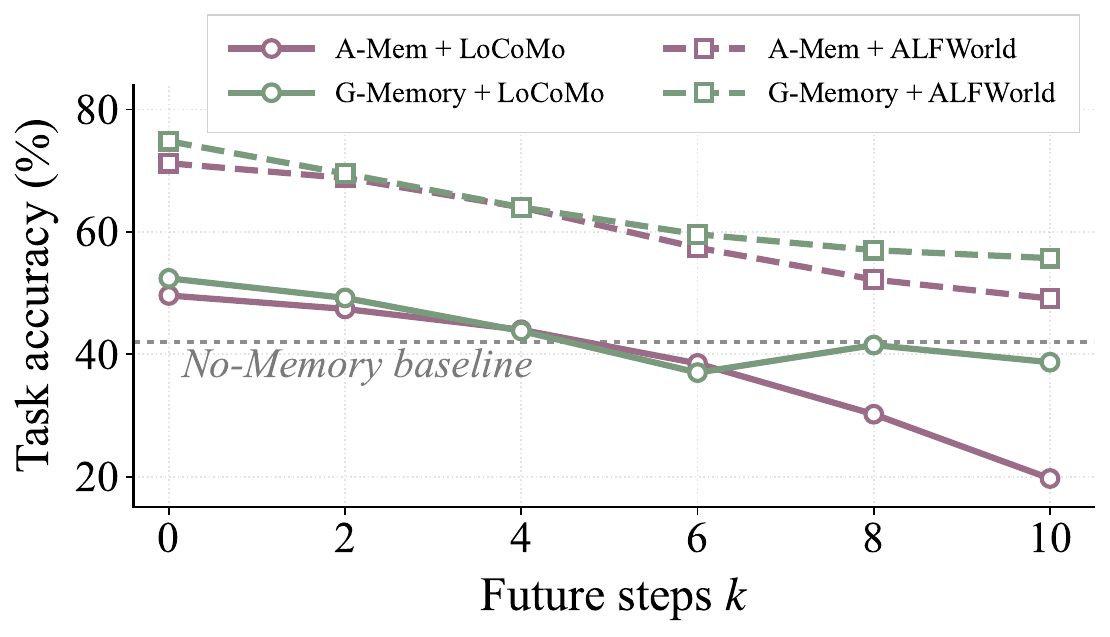}
\vspace{-20pt}
\caption{Decision accuracy as a function of future steps after a spurious retrieval.}
\label{fig:rq2}
\end{wrapfigure}

\textbf{Finding \ding{183}: All memory architectures share a dominant weakness, with one architecture-specific add-on.}
Across most results (besides LoCoMo), $\texttt{T}_2$ (unmeasured confounding) gives the highest spurious
reasoning ratio. This is the hardest type in reasoning: the latent confounder is not recorded anywhere in memory, so retrieval has no signal to filter on. On top of this shared weakness,
graph-based systems (G-Memory, AriGraph) show extra inflation under $\texttt{T}_1$ on long-form
data such as LoCoMo, where repeated co-occurrence between confounded variables accumulates
into heavier edge weights. Both patterns share one reason: relevance is judged by association,
with no causal check. Hence, a mitigative calibration method should address both, while adapting
\textit{what} it operates on (embedding representations or node features) to each architecture.

\textbf{Finding \ding{184}: A single spurious retrieval keeps affecting later decisions.}
We further evaluate how a spurious correlation in an intermediate reasoning step dynamically influences later decisions.
Figure~\ref{fig:rq2} (with additional results presented at Figure \ref{fig:rq2_extra}) shows that accuracy after a spurious retrieval drops in different
trends across subsequent steps (sometimes even partially recovering), with the specific
shape depending on architecture and task. Intervening only at the moment of retrieval is
therefore not enough, because the earlier write that stored the spurious entry keeps
shaping later retrievals. Calibration must act stepwise following the reasoning trajectory at both stages of memory interactions: at write time to
keep spurious entries out of memory, and at retrieval time to catch the ones that get in, which shapes the design in §\ref{sec:method}.

%% file: tables/rq1_main_results.tex
\begin{table*}[t]
\centering
\small
\caption{Results on the constructed benchmark. Org.\ (highlighted with \caporg) reports accuracy ($\uparrow$) on original tasks; $\texttt{T}_1$-$\texttt{T}_3$ report spurious reasoning ratios ($\downarrow$) under the three spurious types. Results are reported as mean$\pm$std over six runs across temperatures (0.5, 0.75, 1.0) and seeds.}
\label{tab:rq1_main}
\setlength{\tabcolsep}{1.5pt}
\begin{adjustbox}{max width=\textwidth}
\begin{tabular}{ll Occc Occc Occc Occc}
\toprule
\rowcolor{headshade}
\multicolumn{2}{l}{\multirow{2.5}{*}{\textbf{Mem. + LLM}}} & \multicolumn{4}{c}{\textbf{ALFWorld}} & \multicolumn{4}{c}{\textbf{ScienceWorld}} & \multicolumn{4}{c}{\textbf{LoCoMo}} & \multicolumn{4}{c}{\textbf{WebShop}} \\
\cmidrule(lr){3-6} \cmidrule(lr){7-10} \cmidrule(lr){11-14} \cmidrule(lr){15-18}
\rowcolor{headshade}
&
  & Org. ($\uparrow$) & $\texttt{T}_1$ ($\downarrow$) & $\texttt{T}_2$ ($\downarrow$) & $\texttt{T}_3$ ($\downarrow$)
  & Org. ($\uparrow$) & $\texttt{T}_1$ ($\downarrow$) & $\texttt{T}_2$ ($\downarrow$) & $\texttt{T}_3$ ($\downarrow$)
  & Org. ($\uparrow$) & $\texttt{T}_1$ ($\downarrow$) & $\texttt{T}_2$ ($\downarrow$) & $\texttt{T}_3$ ($\downarrow$)
  & Org. ($\uparrow$) & $\texttt{T}_1$ ($\downarrow$) & $\texttt{T}_2$ ($\downarrow$) & $\texttt{T}_3$ ($\downarrow$) \\
\midrule

\multirow{3}{*}{No Mem}
& L4
  & 38.3\tiny{$\pm$0.7} &  6.8\tiny{$\pm$0.9} & 19.4\tiny{$\pm$1.1} &  8.1\tiny{$\pm$1.4}
  & 52.7\tiny{$\pm$4.8} &  5.3\tiny{$\pm$0.4} & 12.9\tiny{$\pm$3.6} &  15.4\tiny{$\pm$3.1}
  & 18.6\tiny{$\pm$6.3} & 14.2\tiny{$\pm$2.8} &  9.1\tiny{$\pm$0.7} & 21.3\tiny{$\pm$1.8}
  & 38.4\tiny{$\pm$1.3} &  8.7\tiny{$\pm$1.6} & 17.2\tiny{$\pm$5.9} &  6.3\tiny{$\pm$0.4} \\
& Qw
  & 51.9\tiny{$\pm$5.1} & 10.6\tiny{$\pm$1.8} &  7.3\tiny{$\pm$0.4} & 15.8\tiny{$\pm$0.8}
  & 38.1\tiny{$\pm$2.7} & 11.7\tiny{$\pm$0.5} & 21.4\tiny{$\pm$3.5} &  9.6\tiny{$\pm$13.2}
  & 27.4\tiny{$\pm$7.4} &  7.9\tiny{$\pm$1.5} & 12.6\tiny{$\pm$1.2} & 18.5\tiny{$\pm$3.6}
  & 29.7\tiny{$\pm$7.3} & 14.3\tiny{$\pm$0.8} &  8.9\tiny{$\pm$1.1} & 11.7\tiny{$\pm$2.0} \\
& Ms
  & 46.2\tiny{$\pm$2.3} & 15.9\tiny{$\pm$1.6} &  9.8\tiny{$\pm$1.3} & 12.7\tiny{$\pm$1.9}
  & 43.5\tiny{$\pm$6.9} & 14.2\tiny{$\pm$2.3} & 18.7\tiny{$\pm$5.7} & 12.1\tiny{$\pm$4.4}
  & 21.3\tiny{$\pm$2.8} & 19.6\tiny{$\pm$3.9} &  6.4\tiny{$\pm$0.8} & 24.8\tiny{$\pm$1.5}
  & 35.1\tiny{$\pm$4.6} & 17.8\tiny{$\pm$1.2} & 11.3\tiny{$\pm$0.5} & 14.9\tiny{$\pm$0.7} \\

\midrule

\multirow{3}{*}{A-Mem}
& L4
  & 58.4\tiny{$\pm$7.2} & 31.7\tiny{$\pm$1.3} & 48.3\tiny{$\pm$4.6} & 24.9\tiny{$\pm$0.8}
  & 74.6\tiny{$\pm$1.5} & 27.4\tiny{$\pm$8.3} & 55.1\tiny{$\pm$2.1} & 41.8\tiny{$\pm$0.6}
  & 36.2\tiny{$\pm$3.4} & 58.9\tiny{$\pm$11.2} & 29.4\tiny{$\pm$6.7} & 47.3\tiny{$\pm$2.5}
  & 51.8\tiny{$\pm$0.9} & 34.6\tiny{$\pm$5.8} & 67.2\tiny{$\pm$1.7} & 28.1\tiny{$\pm$9.4} \\
& Qw
  & 76.3\tiny{$\pm$2.1} & 39.5\tiny{$\pm$6.4} & 61.8\tiny{$\pm$7.0} & 33.2\tiny{$\pm$4.9}
  & 61.7\tiny{$\pm$8.5} & 34.8\tiny{$\pm$1.8} & 62.3\tiny{$\pm$3.7} & 49.5\tiny{$\pm$0.7}
  & 48.9\tiny{$\pm$1.6} & 43.7\tiny{$\pm$7.2} & 44.1\tiny{$\pm$2.3} & 31.8\tiny{$\pm$5.1}
  & 44.3\tiny{$\pm$3.8} & 51.4\tiny{$\pm$10.9} & 38.7\tiny{$\pm$5.6} & 62.5\tiny{$\pm$1.4} \\
& Ms
  & 69.1\tiny{$\pm$1.7} & 53.2\tiny{$\pm$3.5} & 36.9\tiny{$\pm$9.2} & 58.6\tiny{$\pm$1.4}
  & 53.4\tiny{$\pm$4.1} & 46.9\tiny{$\pm$1.2} & 69.5\tiny{$\pm$6.8} & 38.2\tiny{$\pm$8.7}
  & 57.6\tiny{$\pm$0.8} & 37.1\tiny{$\pm$4.9} & 51.8\tiny{$\pm$1.5} & 68.4\tiny{$\pm$3.2}
  & 63.9\tiny{$\pm$2.6} & 44.8\tiny{$\pm$7.4} & 79.3\tiny{$\pm$9.1} & 36.7\tiny{$\pm$4.3} \\

\midrule

\multirow{3}{*}{Mem0}
& L4
  & 59.8\tiny{$\pm$3.3} & 36.4\tiny{$\pm$9.7} & 53.9\tiny{$\pm$1.6} & 29.7\tiny{$\pm$4.2}
  & 66.2\tiny{$\pm$1.1} & 31.5\tiny{$\pm$5.4} & 59.7\tiny{$\pm$2.8} & 44.3\tiny{$\pm$0.7}
  & 43.7\tiny{$\pm$5.9} & 63.4\tiny{$\pm$1.4} & 34.8\tiny{$\pm$8.3} & 52.1\tiny{$\pm$2.6}
  & 57.3\tiny{$\pm$1.8} & 39.2\tiny{$\pm$6.5} & 71.6\tiny{$\pm$1.3} & 32.4\tiny{$\pm$4.7} \\
& Qw
  & 63.7\tiny{$\pm$8.6} & 44.8\tiny{$\pm$2.1} & 68.3\tiny{$\pm$1.4} & 37.6\tiny{$\pm$5.8}
  & 78.3\tiny{$\pm$2.4} & 38.7\tiny{$\pm$1.0} & 46.2\tiny{$\pm$7.3} & 63.8\tiny{$\pm$3.1}
  & 31.9\tiny{$\pm$4.5} & 71.5\tiny{$\pm$1.8} & 42.6\tiny{$\pm$2.9} & 58.7\tiny{$\pm$6.4}
  & 68.4\tiny{$\pm$1.2} & 55.1\tiny{$\pm$3.7} & 43.9\tiny{$\pm$8.9} & 77.3\tiny{$\pm$1.6} \\
& Ms
  & 72.5\tiny{$\pm$1.9} & 59.3\tiny{$\pm$4.6} & 41.7\tiny{$\pm$2.3} & 64.8\tiny{$\pm$0.9}
  & 57.9\tiny{$\pm$6.7} & 52.4\tiny{$\pm$1.6} & 76.1\tiny{$\pm$3.5} & 35.9\tiny{$\pm$2.2}
  & 64.8\tiny{$\pm$1.3} & 48.2\tiny{$\pm$7.8} & 79.4\tiny{$\pm$2.1} & 41.5\tiny{$\pm$1.7}
  & 74.1\tiny{$\pm$4.4} & 61.7\tiny{$\pm$1.5} & 85.2\tiny{$\pm$3.8} & 49.6\tiny{$\pm$2.1} \\

\midrule

\multirow{3}{*}{G-Mem}
& L4
  & 83.4\tiny{$\pm$4.8} & 41.9\tiny{$\pm$1.7} & 57.6\tiny{$\pm$6.3} & 34.2\tiny{$\pm$2.4}
  & 71.8\tiny{$\pm$1.4} & 35.3\tiny{$\pm$8.6} & 63.4\tiny{$\pm$2.9} & 48.7\tiny{$\pm$1.1}
  & 52.4\tiny{$\pm$7.1} & 67.8\tiny{$\pm$2.3} & 39.5\tiny{$\pm$1.6} & 56.3\tiny{$\pm$4.8}
  & 62.7\tiny{$\pm$2.6} & 43.6\tiny{$\pm$1.0} & 58.4\tiny{$\pm$8.2} & 37.1\tiny{$\pm$3.9} \\
& Qw
  & 68.9\tiny{$\pm$1.5} & 48.3\tiny{$\pm$7.1} & 73.7\tiny{$\pm$2.8} & 41.5\tiny{$\pm$1.3}
  & 82.6\tiny{$\pm$3.7} & 43.1\tiny{$\pm$1.9} & 51.8\tiny{$\pm$8.4} & 69.2\tiny{$\pm$2.6}
  & 67.3\tiny{$\pm$2.1} & 54.9\tiny{$\pm$6.3} & 83.6\tiny{$\pm$1.4} & 46.2\tiny{$\pm$3.7}
  & 78.9\tiny{$\pm$1.8} & 58.4\tiny{$\pm$4.5} & 47.3\tiny{$\pm$1.2} & 81.7\tiny{$\pm$6.9} \\
& Ms
  & 55.2\tiny{$\pm$8.9} & 63.7\tiny{$\pm$1.4} & 49.4\tiny{$\pm$5.7} & 78.3\tiny{$\pm$2.2}
  & 64.5\tiny{$\pm$7.8} & 57.6\tiny{$\pm$2.4} & 81.9\tiny{$\pm$1.3} & 43.4\tiny{$\pm$5.6}
  & 59.1\tiny{$\pm$3.5} & 72.3\tiny{$\pm$1.6} & 61.7\tiny{$\pm$4.9} & 88.4\tiny{$\pm$1.2}
  & 55.3\tiny{$\pm$1.1} & 67.9\tiny{$\pm$3.4} & 84.3\tiny{$\pm$0.7} & 54.8\tiny{$\pm$8.3} \\

\midrule

\multirow{3}{*}{AriGraph}
& L4
  & 71.6\tiny{$\pm$6.4} & 46.2\tiny{$\pm$1.8} & 61.8\tiny{$\pm$3.5} & 38.7\tiny{$\pm$2.9}
  & 56.4\tiny{$\pm$10.3} & 49.8\tiny{$\pm$7.5} & 77.3\tiny{$\pm$5.7} & 63.1\tiny{$\pm$4.2}
  & 58.7\tiny{$\pm$1.6} & 72.4\tiny{$\pm$3.8} & 44.9\tiny{$\pm$6.1} & 61.5\tiny{$\pm$2.4}
  & 67.4\tiny{$\pm$4.9} & 48.3\tiny{$\pm$1.3} & 81.7\tiny{$\pm$3.6} & 42.6\tiny{$\pm$7.8} \\
& Qw
  & 86.2\tiny{$\pm$1.7} & 52.7\tiny{$\pm$4.3} & 77.4\tiny{$\pm$1.1} & 45.8\tiny{$\pm$8.5}
  & 69.3\tiny{$\pm$5.1} & 47.5\tiny{$\pm$2.6} & 58.9\tiny{$\pm$9.4} & 74.6\tiny{$\pm$1.8}
  & 74.1\tiny{$\pm$2.9} & 61.3\tiny{$\pm$1.4} & 88.2\tiny{$\pm$5.7} & 51.4\tiny{$\pm$3.1}
  & 83.5\tiny{$\pm$1.6} & 63.8\tiny{$\pm$7.2} & 52.1\tiny{$\pm$2.3} & 86.4\tiny{$\pm$1.9} \\
& Ms
  & 60.9\tiny{$\pm$10.2} & 68.4\tiny{$\pm$1.6} & 53.7\tiny{$\pm$9.1} & 73.2\tiny{$\pm$0.7}
  & 74.7\tiny{$\pm$1.9} & 72.1\tiny{$\pm$4.8} & 87.5\tiny{$\pm$1.4} & 69.3\tiny{$\pm$6.7}
  & 48.3\tiny{$\pm$7.4} & 78.6\tiny{$\pm$2.1} & 67.4\tiny{$\pm$1.3} & 93.8\tiny{$\pm$4.6}
  & 61.2\tiny{$\pm$2.5} & 83.5\tiny{$\pm$1.1} & 75.7\tiny{$\pm$7.9} & 67.3\tiny{$\pm$8.2} \\
\bottomrule
\end{tabular}
\end{adjustbox}
\end{table*}

%% file: 04method-0503.tex

\section{\system: Memory Calibration to Mitigate Spurious Correlations}
\label{sec:method}

We propose \system (\underline{CA}usality-informed \underline{ME}mory
ca\underline{L}ibration), a plug-and-play mitigation framework that attaches to
agentic memory architectures and mitigates the spurious types in
\S\ref{sec:def} without additional LLM forward passes or fine-tuning.
We first present \system on embedding-based memories (where each entry
carries an embedding representation; \S\ref{ssec:write}--\S\ref{ssec:retrieve}),
then show how the same operations adapt to graph-based memories whose
nodes are text and edges are topological (\S\ref{ssec:graph-adapt}).

{\bf Design rationale.} \system follows a unified causal invariance
principle \citep{peters2016causal,sloman2004causal}: \ul{a correlation
between a memory entry $m$ and a downstream outcome $Y$ is truly causal
only if it remains when variables that are \textbf{not} part of the
causal pathway are varied}. In spurious correlations (\S\ref{sec:def}),
an explicit or latent confounder $C$ drives both entry and action, so
varying $C$ while holding entry content fixed reveals that the apparent
relevance was partly $C$'s doing
($\texttt{T}_\texttt{1}$, $\texttt{T}_\texttt{2}$); a collider $S$
creates non-causal association only in the conditioned sample, so
breaking the conditioning removes it ($\texttt{T}_\texttt{3}$).

{\bf Calibration overview.} Spurious correlations are
introduced when memory is written and propagate through retrieval.
\system therefore calibrates each stage: at write time, it enforces
causal invariance within a single reasoning step by transforming the
stored representation; at retrieval time, it filters across steps by
keeping only candidates whose rank survives non-causal perturbations
of the query. Either stage can be deployed in isolation; together they
form a stronger ensemble.

\subsection{Write-Stage Calibration}
\label{ssec:write}

\begin{wrapfigure}{r}{0.40\textwidth}
\vspace{-1.4em}
\begin{tcolorbox}[
  colback=gray!0, colframe=black!55, boxrule=0.4pt, arc=1.5pt,
  left=2.5pt, right=2.5pt, top=2pt, bottom=2pt,
  fontupper=\footnotesize,
  title=\textbf{What is a ``\textit{step}''?}, coltitle=black, colbacktitle=lavendermist,
  fonttitle=\footnotesize\bfseries
]
A \textbf{\textit{step}} in this paper refers to one read-or-write interaction with memory, demarcated by the host agent. Within a step, the agent's
context (task type, room, current goal) is fixed, so every entry written
in that step shares the same context value (\textit{Property 2} in
App.\ref{app:formulation-justification}).

\smallskip\noindent\textbf{\textit{Example (ALFWorld, ``cool a tomato'').}}
Step 3 begins when the agent is at the countertop and writes
\texttt{``took tomato 1''}; it ends when the agent moves on. Any further
entries written before the next move (e.g., a re-observation of the
countertop) belong to the \emph{same} step 3. Step 4 begins at the
fridge with \texttt{``opened fridge 1''}, and so on. The shared context
of step 3 is ``at countertop, holding nothing $\to$ tomato''; of step 4,
``at fridge, holding tomato''. The step mean $\mu^{(s)}$ is computed
over the entries that share each such context.
\end{tcolorbox}
\vspace{-0.8em}
\end{wrapfigure}

\textbf{Spurious component at write time.}
Inspired by causal representation learning
\citep{peters2014causal,scholkopf2021toward}, we treat each raw
representation $h_m \in \mathbb{R}^d$ of a memory entry $m$ as having
two parts: a causal part driven by the entry's own content and a
context part driven by the step the entry was written in:
\begin{equation}
  \label{eq:decomposition}
  h_m \;=\; h_m^{\text{causal}} + h_m^{\text{context}} + \varepsilon_m
\end{equation}
with $\varepsilon_m$ idiosyncratic noise. Eq.~(\ref{eq:decomposition})
is not an extra modelling choice: it follows from a first-order Taylor
expansion of the encoder around the step's context value, with the
remainder term second-order small (App.\ref{app:formulation-justification}).
A retriever that reads $h_m$ directly will pick up spurious signal
whenever the query's context matches the context that produced $m$.

\textbf{Calibration operation.}
Within a \textbf{\textit{step}}, $h_m^{\text{context}}$ is shared across all entries, while causal contents and noise differ from entry to entry.
The step mean
$\mu^{(s)} = \tfrac{1}{n_s}\sum_m h_m$ therefore concentrates on
$h_m^{\text{context}}$ as the step fills, and subtracting it leaves the
entry-specific causal content.We compute the residual against the running mean
before $h_m$ is folded in, then update:
\begin{equation}
\tilde{h}_m = h_m - \mu^{(s)}, \quad 
\mu^{(s)} \leftarrow \mu^{(s)} + \frac{1}{n+1}\bigl(h_m - \mu^{(s)}\bigr)
\label{eq:write-residual}
\end{equation}
The update equals the batch mean, and subtracting it performs 
within-step residualization: $\tilde{h}_m$ removes the first-order component 
shared by entries written under the same step context, while preserving 
entry-specific variation up to estimation error 
(App.\ref{app:formulation-justification} formalizes this and proves 
the consistency of $\mu^{(s)}$). In an embedding memory, $\tilde{h}_m$ 
replaces $h_m$ in the ANN \cite{arya1998optimal} index and all 
downstream nearest-neighbour operations proceed unchanged 
(App.\ref{app:session-mean} gives the initialization, ordering, and 
closure details for implementation).

\textbf{Two parts, one stage.}
The residualization above neutralizes $\texttt{T}_\texttt{1}$ (explicit confounding) and
$\texttt{T}_\texttt{2}$ (unmeasured confounding), both of which inject a shared bias into
all entries of a step. $\texttt{T}_\texttt{3}$ (collider bias) is different: it arises from
\emph{which} entries are written, not how they are represented. We
therefore pair Eq.~(\ref{eq:write-residual}) with a
write criterion that decides whether to store $\tilde{h}_m$ based only
on its similarity to existing entries, never on a downstream outcome
(App.\ref{app:write-criterion}). This severs the path
``outcome $\to$ stored in memory'' that makes the write event a
collider. Together, residualization and the novelty criterion form
write-stage calibration; coverage of all three types is summarized in
App.\ref{app:write-analysis}.

\subsection{Retrieve-Stage Calibration}
\label{ssec:retrieve}

\textbf{Spurious component at retrieval time.}
Write-stage calibration removes any confounder that is constant within
a single step. What remains are confounders whose value \emph{shifts}
between steps: such a variable leaves no consistent influence in any
single $\mu^{(s)}$, however, still creates spurious correlations when entries
from different steps are retrieved together. Causal invariance principle \citep{peters2016causal,sloman2004causal} suggests
a direct test: if a candidate is truly relevant, its rank should not
change when the query is perturbed along directions unrelated to causal
content; if its rank came from spurious overlap, the perturbation will
break it. We call this the \emph{causal stability} test.

\textbf{Identifying the non-causal subspace.}
We need to know which axes carry no causal content. An axis is a unit
vector $v \in \mathbb{R}^d$, and the projection
$\langle v, \tilde h_m \rangle$ tells us where $\tilde h_m$ sits along
that axis. We flag $v$ as non-causal when
$\langle v, \tilde h_m \rangle$ varies more between steps than within
steps, i.e., $v$ tracks something that shifts entries of a step
together rather than distinguishing entries within a step. This needs
no outcome labels and is maintained online via streaming statistics on
the calibrated $\tilde h_m$
(App.\ref{app:projection-update}). We write the resulting
subspace as $\mathcal{S}_{\text{nc}}$.

\textbf{Calibration by causal stability.}
Let $\mathcal{M}_k(q)$ be the top-$k$ candidates returned by the base
retriever with scores $\phi(q,m)$, and let $v_1,\ldots,v_L$ be an
orthonormal basis of $\mathcal{S}_{\text{nc}}$. Each perturbation
query $q^{(\ell)} = q + \delta_\ell v_\ell$ shares its causal content
with $q$ and differs only along one spurious direction, with
$\delta_\ell$ the standard deviation of
$\langle v_\ell, \tilde h \rangle$ across entries. For inner-product
scoring, linearity gives:
\begin{equation}
  \label{eq:cf-score}
  \phi(q^{(\ell)}, m) \;=\; \phi(q, m) \;+\; \delta_\ell \,\langle v_\ell, \tilde h_m \rangle
\end{equation}
so $|\langle v_\ell, \tilde h_m \rangle|$ is exactly the absolute score
change of $m$ under the $\ell$-th perturbation (up to the fixed scale
$\delta_\ell$). A truly causal candidate has small projection onto every
$v_\ell$ and its score barely moves. In contrast, a candidate ranked highly through
shared spurious structure has large projection onto some $v_\ell$ and
its score collapses. We summarize this with the stability statistic
\begin{equation}
  \label{eq:stability}
  \sigma(m) \;=\; \texttt{median}_{\ell=1}^{L} \,
  \bigl|\langle v_\ell, \tilde h_m \rangle\bigr|
\end{equation}
Small $\sigma(m)$ means $m$ is stable; large $\sigma(m)$ means $m$'s
rank rested on a spurious direction. We then re-rank
$\mathcal{M}_k(q)$ in increasing order of $\sigma(m)$: the same $k$
candidates are retrieved, but the most stable ones come first. The
median is chosen over the mean for robustness: a single dominant
spurious direction would shift the mean uniformly across all candidates
and erase ranking signal, while the median is unaffected
(justified in App.\ref{app:formulation-justification}). $\sigma(m)$=0
characterizes a candidate lying entirely in the
causal subspace, making the test conservative on causal patterns
(App.\ref{app:stability}). 

\subsection{Adaptation to Graph-Based Memory}
\label{ssec:graph-adapt}
 
The calibration designs above assumes that each entry has a
vector (embedding) representation in $\mathbb{R}^d$. Graph-based memories
such as G-Memory \citep{zhang2025g} and AriGraph
\citep{anokhin2024arigraph} store entries as nodes connected by
topological edges and retrieve them by graph traversal rather than by
vector similarity, but the calibration plugs in unchanged once a
per-node vector is available. We supply such a vector based on how much information each node carries: (i) When the host
already maintains a node embedding for its own similarity scoring, we calibrate
that embedding directly. (ii) When nodes carry text but no embedding,
we encode the text (typically with topological information) by standard encoder. In every case, $\tilde{h}_m$
replaces $h_m$ for any similarity computation the graph memory
performs, the rerank in Eq.~\eqref{eq:stability} is applied to the
candidate set returned by graph traversal, and the topology, edges,
and stored text are not modified. Full recipes, including how step
boundaries are inferred from temporal/episodic edges, are in
App.\ref{app:graph-adapt}.
 

%% file: 05expt.tex
\section{Experiments for Calibration}
\label{sec:experiments}

This section evaluates \system through
three research questions: (Q1) Does \system mitigate spurious correlations while maintaining original reasoning? (Q2) Can \system maintain effectiveness
 under attacks? (Q3) How much computational overhead does \system add?

\noindent\textbf{Benchmarks, models, and memory system}. We use the same settings as in \S\ref{sec:def}.

\noindent\textbf{Baselines.} We compare against four baselines that calibrate or robustify agentic memory:
(1) \emph{Vanilla} serves as no-calibration reference (same as Table~\ref{tab:rq1_main}); (2)
\emph{IPW}~\citep{schnabel2016recommendations} reweights memory entries by inverse propensity to suppress high-frequency co-occurrences; (3)
\emph{JTT}~\citep{liu2021just} first identifies 
examples affected by spurious correlations and then train a baseline 
model with these examples upweighted before being 
applied with memory; (4) \emph{DeCoT}~\citep{wu2024decot} applies entity replacement at the text level. 
Those baselines are deliberately chosen to cover the three levels of operations: retrieval scoring (IPW),  model improvement (JTT), and output rewriting (DeCoT).

\noindent\textbf{Implementation details.}
The main experiments use \textsc{Llama-4} as the backbone LLM, with extra results shown in App.\ref{app:additional_results}. All results are reported as mean$\pm$std over six independent runs, 
\subsection{Q1: Main Results}
\label{ssec:rq1_compare}

\input{tables/baseline_comparison}

\textbf{Results.} Table~\ref{tab:rq1_compare} shows three patterns. (i) \system attains the lowest spurious ratio in most (memory, dataset, type) cells, and matches the best baseline within run variance elsewhere. (ii) Its gains span all three types $\texttt{T}_\texttt{1}$--$\texttt{T}_\texttt{3}$, while each baseline only helps on what its mechanism targets. Specifically, IPW on repeated co-occurrence (e.g., \textsc{WebShop}), JTT on short trajectories where first-pass errors concentrate (e.g., \textsc{ALFWorld}), and DeCoT when confounders surface as named entities (e.g., \textsc{LoCoMo}). (iii) Org.\ accuracy under \system never trades off against spuriousness reduction, whereas baselines repeatedly improve one metric at another's expense. Ablation results are shown at App.\ref{app:ablation}.

\textbf{Explanations and insights.} Each baseline covers only one aspect of the problem because they treat a \textit{symptom} (frequency, error pattern, or surface entity) as if it were the cause. However, all three spurious patterns ($\texttt{T}_\texttt{1}$ to $\texttt{T}_\texttt{3}$) are downstream effects of the same upstream non-causal association in memory. Hence, once that association is written, every later baseline fights the consequences rather than the source. \system acts before frequency accumulates, before errors propagate, and before entities surface as text, so we handle $\texttt{T}_\texttt{1}$ to $\texttt{T}_\texttt{3}$ that previously needed three different tools. A second insight from Table \ref{tab:rq1_compare} concerns how accuracy and spuriousness relate: every baseline shows a residual tension where reducing one spurious type leaves another (or Org. accuracy) worse, whereas \system improves them jointly since \system  removes nuisance directions that hurt both clean and adversarial retrieval, so reducing spurious correlations and improving causal retrieval no longer compete.

\subsection{Q2: Adaptive Attacks}
\label{ssec:rq2_adaptive}

\noindent\textbf{Attack setting.}
We evaluate \system under an adaptive threat model where the attacker knows the calibration mechanism and injects spurious entries tailored to each architecture: a distribution attack (Algo.~\ref{alg:crossstep}) for embedding memory and a mimic causal-pattern attack (Algo.~\ref{alg:graphmimic}) for graph memory. Both target each of $\texttt{T}_1$/$\texttt{T}_2$/$\texttt{T}_3$. Construction details are in App.\ref{app:adaptive_attacks}.

\begin{figure}[t]
\centering
\includegraphics[width=\linewidth]{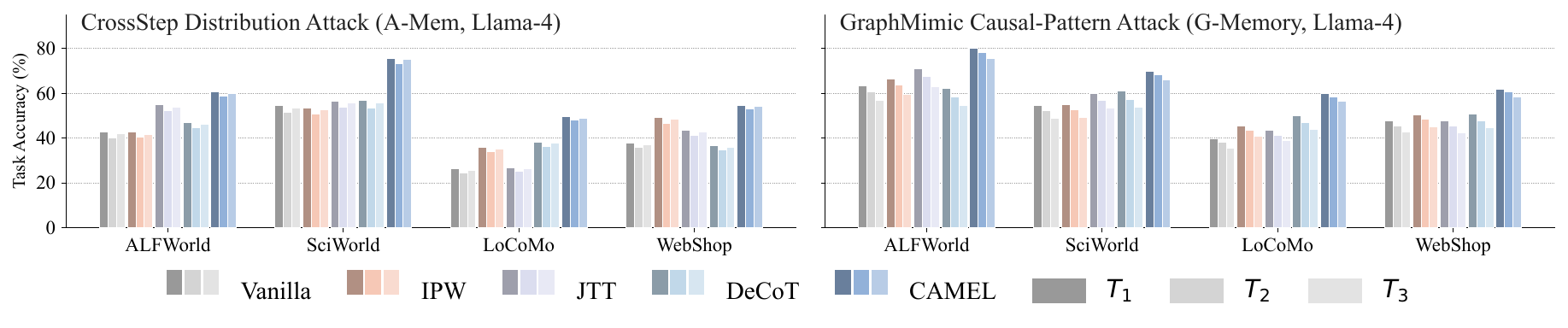}
\vspace{-15pt}
\caption{Accuracy of \system and baselines under adaptive attacks. Each bar indicates the spurious pattern type injected by an attacker aware of the calibration method.}
\vspace{-3pt}
\label{fig:rq2_adaptive}
\end{figure}

\textbf{Results.} Figure~\ref{fig:rq2_adaptive} shows that \system degrades moderately rather than collapsing: across all configurations, accuracy stays above the no-memory baseline. The failure mode is architecture-specific. On A-Mem, cross-step distributed attacks cause the largest drops, implying embedding memory's residual sensitivity to evidence spread thinly across steps. On G-Memory, mimic causal-pattern attacks dominate instead, showing that topological manipulation is harder to neutralize than node-level features. $\texttt{T}_\texttt{1}$ attacks degrade least on both, consistent with the practice that explicit confounding being the cleanest signal for calibration to remove.

\textbf{Explanations and insights.} The architecture-specific failures expose where each memory's relevance signal lives. Embedding retrieval reduces each entry to a similarity score, so spurious evidence spread thinly across steps survives because no single step's mean carries enough of it to subtract. Graph retrieval also uses connectivity, which calibration on node features cannot reach: a well-mimicked spurious chain is structurally indistinguishable from a real one even after residualization. The takeaway: representation-level calibration is necessary but not sufficient against architecture-aware adversaries. The natural next steps are topology-level calibration for graph memory and cross-step distribution detection for embedding memory. That \system still degrades gracefully under attacks targeting its blind spots indicates representation calibration buys robustness margin that adaptive attackers can erode but not eliminate.

\subsection{Q3: Computational Efficiency}
\label{ssec:rq3_efficiency}

\begin{wrapfigure}[15]{r}{0.44\textwidth}
\vspace{-15pt}
\centering
\includegraphics[width=0.44\textwidth]{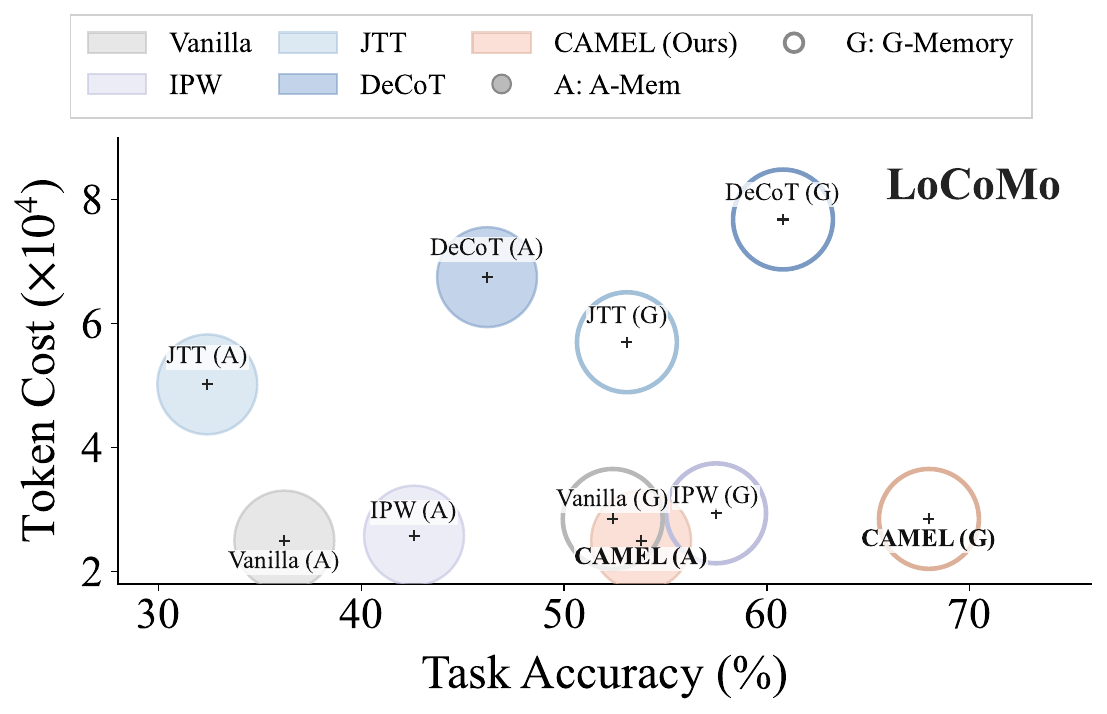}
\vspace{-15pt}
\caption{Accuracy-token cost tradeoff. Radius indicates the token cost variance.}
\vspace{-20pt}
\label{fig:rq3_locomo}
\end{wrapfigure}

\textbf{Results.} Figure~\ref{fig:rq3_locomo} splits methods into two lines: \system, IPW, and Vanilla sit in the low-token regime; JTT and DeCoT use markedly more tokens because their mechanisms invoke extra LLM passes. Within the low-token regime, \system is the most accurate, placing it alone in the lower-right region on both A-Mem and G-Memory. The high-cost baselines do not convert their larger token budgets into better accuracy.

\textbf{Explanations and insights.} Methods that fix memory \emph{after} retrieval (rewriting entries, regenerating reasoning, re-prompting) pay a token cost on every query, since the LLMs, the most expensive component, does the remedial work. \system acts \emph{earlier}: calibration runs once at write time and as a cheap re-rank at retrieval, so the LLM never sees a candidate set needing cleanup. The bottleneck for reliable agentic memory is therefore not how much memory text an LLM reads, but whether it is causally useful before the LLM consumes it. \system's efficiency gain comes from removing cost that should never have been needed. We  also report time latency at App.\ref{app:latency}.

\section{Conclusion}
\label{sec:conclusion}

Agentic memory is vulnerable to spurious correlations persisting and propagating into later decisions. This work studies this risk along two contributions. First, we benchmark and diagnose agentic memory systems on various spurious patterns, showing that memory boosts reasoning on clean inputs yet amplifies reliance on spurious patterns. Second, we proposed \system, a lightweight calibration that attaches to existing memory architectures at write and retrieval time. \system consistently reduces the spurious influence while preserving or improving clean-input accuracy, and stays robust under adaptive attacks targeting the calibration itself. Together, the benchmark and \system offer a principled step toward more reliable agentic memory deployment.

%% file: tables/baseline_comparison.tex
\begin{table*}[t]
\centering
\small
\caption{Performance on spurious correlation benchmarks. Similar to Table~\ref{tab:rq1_main}, Org.\,($\uparrow$): clean-memory accuracy; $\texttt{T}_\texttt{1}$/$\texttt{T}_\texttt{2}$/$\texttt{T}_\texttt{3}$\,($\downarrow$): spurious reasoning ratios. The \capbest\ and \capsecond\ results are highlighted.}
\label{tab:rq1_compare}
\setlength{\tabcolsep}{3pt}
\renewcommand{\arraystretch}{0.85}
\begin{adjustbox}{max width=\textwidth}
\begin{tabular}{ll cccc cccc cccc cccc}
\toprule
& & \multicolumn{4}{c}{\textbf{ALFWorld}} & \multicolumn{4}{c}{\textbf{ScienceWorld}} & \multicolumn{4}{c}{\textbf{LoCoMo}} & \multicolumn{4}{c}{\textbf{WebShop}} \\
\cmidrule(lr){3-6} \cmidrule(lr){7-10} \cmidrule(lr){11-14} \cmidrule(lr){15-18}
\textbf{Mem.} & \textbf{Method}
  & Org. ($\uparrow$) & $\texttt{T}_\texttt{1}$ ($\downarrow$) & $\texttt{T}_\texttt{2}$ ($\downarrow$) & $\texttt{T}_\texttt{3}$ ($\downarrow$)
  & Org. ($\uparrow$) & $\texttt{T}_\texttt{1}$ ($\downarrow$) & $\texttt{T}_\texttt{2}$ ($\downarrow$) & $\texttt{T}_\texttt{3}$ ($\downarrow$)
  & Org. ($\uparrow$) & $\texttt{T}_\texttt{1}$ ($\downarrow$) & $\texttt{T}_\texttt{2}$ ($\downarrow$) & $\texttt{T}_\texttt{3}$ ($\downarrow$)
  & Org. ($\uparrow$) & $\texttt{T}_\texttt{1}$ ($\downarrow$) & $\texttt{T}_\texttt{2}$ ($\downarrow$) & $\texttt{T}_\texttt{3}$ ($\downarrow$) \\
\midrule

\multirow{5}{*}{A-Mem}
& Vanilla  & 58.4 & 31.7 & 48.3 & 24.9  & 74.6 & 27.4 & 55.1 & 41.8  & 36.2 & 58.9 & 29.4 & 47.3  & 51.8 & 34.6 & 67.2 & 28.1 \\
& + IPW    & 54.1 & 39.3 & 50.7 & 22.6  & 65.2 & 35.8 & \second{49.7} & 49.1  & 42.6 & 46.2 & 24.8 & 42.4  & \second{57.8} & \second{29.4} & \second{59.6} & \second{23.5} \\
& + JTT    & \second{63.7} & \second{25.4} & \second{42.6} & {15.8}  & \second{69.1} & \second{30.2} & 57.4 & \second{37.8}  & 32.4 & 54.7 & 32.1 & 54.6  & 53.1 & 31.8 & 65.5 & 25.4 \\
& + DeCoT  & 59.8 & 28.5 & 44.1 & \second{14.3}  & 68.3 & 33.8 & 56.4 & 40.7  & \second{46.2} & \second{43.7} & \second{23.4} & \second{37.5}  & 46.6 & 43.2 & 70.8 & 32.7 \\
& + \system   & \best{66.0} & \best{12.7} & \best{29.0} & \best{6.5}  & \best{82.2} & \best{10.4} & \best{35.8} & \best{23.4}  & \best{53.8} & \best{29.9} & \best{10.1} & \best{28.9}  & \best{59.4} & \best{15.6} & \best{47.9} & \best{9.7} \\
\midrule

\multirow{5}{*}{Mem0}
& Vanilla  & 59.8 & 36.4 & 53.9 & 29.7  & 66.2 & 31.5 & 59.7 & 44.3  & 43.7 & 63.4 & 34.8 & 52.1  & 57.3 & 39.2 & 71.6 & 32.4 \\
& + IPW    & 52.5 & 41.1 & 50.8 & 32.4  & 69.1 & \second{23.6} & 56.3 & 37.8  & 38.8 & 67.2 & 37.1 & 56.8  & \second{63.1} & \second{27.5} & \second{55.7} & \second{23.2} \\
& + JTT    & \second{62.2} & \second{30.3} & \second{44.7} & 25.7 & 62.8 & 34.8 & 60.1 & 42.3  & 45.3 & 60.8 & \second{29.4} & 48.7  & 58.9 & 36.4 & 68.2 & 29.1 \\
& + DeCoT  & 60.4 & 34.2 & 48.6 & \second{24.3}  & \second{70.8} & 24.1 & \second{50.6} & \second{36.5}  & \second{49.4} & \second{57.3} & 31.2 & \second{45.9}  & 59.2 & 34.7 & 66.1 & 28.4 \\
& + \system   & \best{67.4} & \best{17.4} & \best{34.6} & \best{11.3}  & \best{73.8} & \best{12.5} & \best{40.4} & \best{25.9}  & \best{53.3} & \best{41.4} & \best{15.5} & \best{33.7}  & \best{64.9} & \best{20.2} & \best{52.3} & \best{14.0} \\
\midrule

\multirow{5}{*}{G-Mem.}
& Vanilla  & 83.4 & 41.9 & 57.6 & 34.2  & 71.8 & 35.3 & 63.4 & 48.7  & 52.4 & 67.8 & 39.5 & 56.3  & 62.7 & 43.6 & 58.4 & 37.1 \\
& + IPW    & 84.1 & 39.7 & 54.8 & 31.6  & 69.4 & 36.8 & 61.2 & 49.3  & 57.5 & 60.1 & 33.8 & 50.4  & \second{63.7} & \second{39.3} & \second{55.3} & \second{31.4} \\
& + JTT    & \second{86.4} & \second{35.6} & \second{41.7} & \second{28.4}  & 73.1 & 28.7 & 47.2 & \second{43.2}  & 53.1 & 64.3 & 36.8 & 52.7  & 58.2 & 47.8 & 62.1 & 39.2 \\
& + DeCoT  & 77.8 & 48.4 & 52.6 & 39.8  & \second{74.6} & \second{23.1} & \second{46.3} & 45.4  & \second{60.8} & \second{59.5} & \second{32.7} & \second{49.8}  & 61.8 & 44.1 & 57.6 & 33.1 \\
& + \system   & \best{91.0} & \best{22.9} & \best{38.3} & \best{15.8}  & \best{79.4} & \best{16.3} & \best{44.1} & \best{30.3}  & \best{68.0} & \best{46.8} & \best{20.2} & \best{37.9}  & \best{70.3} & \best{24.6} & \best{39.1} & \best{18.7} \\
\midrule

\multirow{5}{*}{AriGraph}
& Vanilla  & 71.6 & 46.2 & 61.8 & 38.7  & 56.4 & 49.8 & 77.3 & 63.1  & 58.7 & 72.4 & 44.9 & 61.5  & 67.4 & 48.3 & 81.7 & 42.6 \\
& + IPW    & 72.4 & 43.8 & \second{58.4} & 35.9  & 57.2 & 47.1 & 74.6 & 60.3  & \second{69.1} & \second{56.5} & \second{29.4} & 50.2  & \second{73.2} & \second{32.4} & \second{70.8} & 37.5 \\
& + JTT    & \second{72.6} & \second{42.4} & 62.1 & \second{33.7}  & 54.1 & 51.2 & 78.6 & 60.4  & 59.4 & 68.7 & 38.3 & 57.8  & 69.1 & 45.1 & 78.2 & 39.4 \\
& + DeCoT  & 65.1 & 52.4 & 67.1 & 44.2  & \second{58.2} & \second{42.8} & \second{70.1} & \second{56.4}  & 62.4 & 60.2 & 32.1 & \second{47.1}  & 71.8 & 38.6 & 76.4 & \second{35.3}\\
& + \system   & \best{83.2} & \best{27.2} & \best{42.5} & \best{20.3}  & \best{64.0} & \best{30.8} & \best{58.0} & \best{44.7}  & \best{72.3} & \best{53.4} & \best{25.6} & \best{36.1}  & \best{75.0} & \best{29.3} & \best{62.4} & \best{24.2} \\
\bottomrule
\end{tabular}
\end{adjustbox}
\end{table*}

%% file: appendix/Validation.tex
\section{Validation of Identified Spurious Correlations}
\label{app:validation}

Causal discovery from observational data is inherently uncertain, so we validate each identified spurious correlation through a combination of two methods.

\textbf{Validation I: Conditional independence testing.}
For each candidate pair $(X, Y)$ attributed to a confounder $C$, we test whether the observed association between $X$ and $Y$ can be fully explained by $C$. Concretely, we compute the association between $X$ and $Y$ both \emph{before} and \emph{after} conditioning on $C$, i.e., using partial correlation for continuous variables and conditional mutual information for categorical or mixed variables. For continuous variables, the partial correlation is computed as:
\begin{equation}
\label{eq:partial-corr}
\rho_{XY \cdot C} \;=\; \frac{\rho_{XY} - \rho_{XC}\,\rho_{YC}}
{\sqrt{1 - \rho_{XC}^2}\;\sqrt{1 - \rho_{YC}^2}}
\end{equation}
with significance assessed via Fisher's $z$-transformation at $\alpha = 0.05$. For categorical or mixed variables, we compute $I(X;\,Y \mid C)$ and test significance with a permutation test ($B = 1{,}000$ permutations). When the confounder $C$ consists of multiple variables, we condition on all components jointly; strata with fewer than 5 observations are stabilized with Dirichlet smoothing ($\alpha = 1$). We apply Benjamini-Hochberg correction~\citep{benjamini1995controlling} at level $q = 0.05$ across all candidate pairs within each dataset.

A spurious correlation is confirmed when two criteria are jointly met: (1) the unconditional association between $X$ and $Y$ is statistically significant, and (2) the conditional association drops to a level that is no longer distinguishable from zero by the same test. In other words, $C$ must account for the \emph{entirety} of the association, not merely reduce it. Cases where conditioning on $C$ weakens but does not eliminate the association ($0.05 \leq p < 0.10$ after conditioning) are flagged as \emph{partially spurious} and retained for further inspection, as they may indicate an additional unmeasured confounder or a genuine residual causal path.

\textbf{Validation II: Interventional testing.}
Conditional independence testing operates on observational patterns and cannot rule out all alternative explanations. As a stronger check, we construct \emph{interventional} trajectory variants by directly editing the dataset to break the proposed spurious pathway. For a candidate pair $(X, Y)$ attributed to confounder $C$, we produce two sets of edited trajectories: one in which $C$ is held fixed while $X$ is varied, and one in which $X$ is held fixed while $C$ is varied. The fixed value is chosen as the most frequent value of the target variable to maximize sample size.

For ALFWorld and ScienceWorld, which provide programmatic environment control, we implement these interventions by replaying trajectories with the target variable clamped. For LoCoMo and WebShop, we use LLM-based counterfactual editing: an LLM rewrites the trajectory segment containing the target variable while holding all other elements constant. Each edit is validated by checking that (1) token-level edit distance outside the target span stays below 15\%, and (2) a second LLM flags no logical inconsistencies in the rewritten trajectory. Invalid edits are re-attempted up to 3 times before exclusion.

We then measure the association between $X$ and $Y$ separately in each set, using the same statistical tests as in Validation~I but without conditioning on $C$ (since $C$ has already been fixed by the intervention). If the correlation is spurious, the association should be present in the second set (where $C$ varies and drives both $X$ and $Y$) but absent in the first set (where $C$ is removed as a common driver). A correlation is confirmed as spurious when this contrast is observed: the association survives only when $C$ is free to vary. If the association persists even after $C$ is held constant, this indicates that either a direct causal path from $X$ to $Y$ exists or an additional confounder remains unaccounted for, and the candidate is removed from the spurious set.

The procedure above applies directly to $T_1$ (confounding bias), where $C$ is observed. For $T_2$ (unmeasured confounding), $C$ is latent and cannot be directly clamped; we substitute a measured proxy $\hat{C}$ (a descendant of $C$ in the causal graph) and note the residual uncertainty in the benchmark metadata. For T$_3$ (collider bias), the collider $C^\star$ is an entry with parents $X$ (content) and $Y$ (outcome), and the spurious $X$-$Y$ association
arises because reasoning is restricted to entries with $C^\star$ present.
We validate by comparing the $X$-$Y$ association on the sample (where
$C^\star$ is implicitly conditioned on) against a counterfactual sample that
includes entries which would have been excluded by the write rule. Spuriousness is
confirmed when the association is present in the stored sample but disappears in
the unrestricted sample.

A candidate is included in the final benchmark only when both validations agree on its spurious status.

%% file: appendix/benchmark.tex
\section{Complementary Information and Experiments for Benchmarking Spurious Correlation}
\label{app:benchmark}

\subsection{Detail of Dataset}
\label{app:datasets}

This appendix shows how the spuriousness discovery and validation pipeline from Section~\ref{sec:def} is applied to each of the four datasets. For each dataset we present (i) a raw data example, (ii) how we instantiate the causal variable set and run the PC algorithm, (iii) the resulting causal graph, and (iv) how we construct the three types of spurious correlations for evaluation.

\noindent\textbf{On rule-based causal discovery.} Following Section~\ref{sec:def}, we use the PC algorithm for causal discovery, a constraint-based statistical method, not an LLM. No large language model is invoked in Stage A (variable set), Stage B (graph construction), or either validation procedure. Each dataset also carries additional structural information: ALFWorld has PDDL specifications, ScienceWorld has gold trajectories and an action taxonomy, LoCoMo has evidence annotations, and WebShop has a structured attribute schema. We use this information as \textit{prior constraints} on the PC algorithm: must-have edges are injected directly, must-not-have edges prune the initial graph, and temporal-order constraints orient remaining ambiguous edges. The priors tighten the discovery and make the resulting graph verifiable against the datasets' own design, but the PC algorithm itself does the conditional-independence work.

\noindent\textbf{On query placement.} Spurious correlations are injected only into memory records, never into the task trajectories themselves. Each spurious query is positioned either mid-trajectory or at the end. Mid-trajectory queries test whether the agent is misled \textit{during} action selection; end-of-trajectory queries test whether the agent's final conclusion is corrupted. The two positions cover different failure modes: spurious memory can hurt both in-context decisions and post-hoc reasoning.

\begin{table}[h]
\centering
\small
\caption{Dataset statistics. ``Original scale'' follows the published benchmark statistics. ``Processed unit'' and ``Avg.\ length'' describe the trajectory/session units used to construct evaluation instances. $T_1$/$T_2$/$T_3$ report the number of validated spurious correlation pairs per type.}
\label{tab:data_stats}
\resizebox{\linewidth}{!}{%
\begin{tabular}{lccccccc}
\toprule
\textbf{Dataset} & \textbf{Original scale} & \textbf{Processed unit} & \textbf{Avg.\ length} & \textbf{$T_1$} & \textbf{$T_2$} & \textbf{$T_3$} & \textbf{Total} \\
\midrule
ALFWorld     
& 3,827 tasks 
& 1,200 expert traj. 
& 20.1 steps 
& 1,198 & 1,024 & 1,137 & 3,359 \\
ScienceWorld 
& 7,200 variations 
& 191 expert traj. 
& 24.7 steps 
& 2,041 & 2,315 & 1,904 & 6,260 \\
LoCoMo conv.      
& 10 conv. / 1,986 QA 
& 10 conversations 
& 588.2 turns 
& \multirow{2}{*}{621} & \multirow{2}{*}{538} & \multirow{2}{*}{649} & \multirow{2}{*}{1,808} \\
LoCoMo sess.      
& 10 conv. / 1,986 QA 
& 272 sessions 
& 21.6 turns 
&  &  &  &  \\
WebShop      
& 12,087 instr. / 1.18M products 
& 81 logged traj. 
& 11.5 steps 
& 3,512 & 3,187 & 3,846 & 10,545 \\
\bottomrule
\end{tabular}%
}
\end{table}

Several patterns emerge from the validation results. WebShop yields 
the most spurious pairs in absolute terms (10,545), because its large 
instruction set (12,087) combined with a rich product attribute space 
creates systematic opportunities for all three spurious types, 
particularly $T_3$ (collider bias) through search-ranking filters. 
ScienceWorld follows with 6,260 pairs and shows a relative strength 
in $T_2$ (unmeasured confounding), consistent with its latent 
variables such as agent scientific literacy that are never directly 
observed in the state. ALFWorld produces 3,359 pairs roughly balanced 
across all three types, reflecting the diversity of its six task 
families. LoCoMo yields 1,808 pairs from only 10 conversations, the 
highest density per raw scenario, because each long conversation 
contains many co-occurring events across sessions that create rich 
opportunities for confounding ($T_1$) and collider bias ($T_3$). In 
total, we obtain 21,972 validated spurious correlation pairs, with 
approximately 40\% of the spurious queries placed mid-trajectory and 
60\% at the end. The full list of graphs, spurious features, 
validation statistics, and query positions will be released alongside 
the code.

\subsection{ALFWorld}
\label{app:alfworld}
\noindent\textbf{ALFWorld Example.} ALFWorld evaluates whether agents 
can complete household tasks in a simulated environment through 
text-based interaction. The full dataset contains 3,827 games across 
six task types (Pick \& Place, Examine in Light, Clean \& Place, Heat 
\& Place, Cool \& Place, Pick Two \& Place), split into 3,553 
training, 140 seen-validation, and 134 unseen-validation instances. 
Trajectories range from 9 to 49 steps, with an average of 20.1. Each 
example has a task goal such as ``put a cooled tomato on the 
countertop,'' followed by a trajectory where the agent observes the 
scene and takes actions turn by turn. A full example:

\begin{databox}
\begin{Verbatim}[breaklines=true, breakanywhere=true, fontsize=\footnotesize]
{
  "task": "put a cooled tomato on the countertop",
  "environment": "kitchen-10",
  "trajectory": [
    {"step": 1, "action": "look around",
     "observation": "visible: fridge 1, countertop 1, countertop 3, sink 1"},
    {"step": 2, "action": "go to countertop 3",
     "observation": "on countertop 3 you see tomato 1, apple 1"},
    {"step": 3, "action": "take tomato 1 from countertop 3",
     "observation": "you pick up the tomato"},
    {"step": 4, "action": "go to fridge 1",
     "observation": "fridge 1 is closed"},
    {"step": 5, "action": "cool tomato 1 with fridge 1",
     "observation": "tomato is now cool"},
    {"step": 6, "action": "put tomato 1 on countertop 1",
     "observation": "task complete"}
  ],
  "outcome": "SUCCESS"
}
\end{Verbatim}
\end{databox}

\noindent\textbf{Causal Discovery.} We instantiate 
Section~\ref{ssec:curate} as follows. For Stage~A, each trajectory 
yields the variables $\{A_t, M_t, R_t\}_{t=1}^T$ (turn-level actions, 
memory reads, reasoning states) plus session-level context $C = 
\{\text{kitchen\_id}, \text{task\_type}\}$. State predicates tracked 
by the ALFWorld simulator (object positions, temperature, container 
open/closed) become additional observed variables. For Stage~B, we 
run the PC algorithm on 1,200 expert trajectories (200 per task type), 
with three dataset-specific priors drawn from PDDL. First, PDDL 
\texttt{effect} declarations give \textit{must-have} edges: an 
action's effect literal becomes a directed edge from the action 
variable to the state variable. For our running example, 
\texttt{cool(tomato)} has effect \texttt{is\_cool(tomato)}, so the 
edge $\text{action}_5 \to \text{temperature\_state}$ is fixed before 
PC starts removing edges. Second, PDDL \texttt{precondition} 
declarations give \textit{must-not-have} edges: a state variable that 
is not in any action's precondition set cannot be a cause of that 
action in the same turn. Third, temporal order resolves ambiguous 
edges: variables at turn $t$ cannot be caused by variables at turn 
$t' > t$. The PC algorithm then handles the remaining 
conditional-independence tests on state-to-state and context-to-action 
relationships.

\noindent\textbf{Causality \& Causal Graph.}

\begin{databox}
\begin{verbatim}
{
  "factual_roles": {
    "Treatment": ["cool tomato with fridge"],
    "Covariate": ["kitchen layout", "pickup location", "step count"],
    "Mediator":  ["tomato temperature change"],
    "Outcome":   ["task success"]
  },
  "counterfactual_roles": {
    "Treatment": ["omission of cooling action"],
    "Covariate": ["kitchen layout", "pickup location", "step count"],
    "Mediator":  ["tomato temperature unchanged"],
    "Outcome":   ["task failure"]
  },
  "causal_graph": {
    "factual_edges": [
      ["cool tomato with fridge", "tomato temperature change"],
      ["tomato temperature change", "task success"]
    ],
    "counterfactual_edges": [
      ["omission of cooling action", "tomato temperature unchanged"],
      ["tomato temperature unchanged", "task failure"]
    ]
  }
}
\end{verbatim}
\end{databox}

\noindent\textbf{Spurious Correlation Construction.} Using Stage~C of 
Section~\ref{ssec:curate}, we identify spurious pairs across all six 
task families and inject them into memory records (not into the 
trajectories themselves), yielding 3,359 validated pairs in total 
(1,198 $T_1$ / 1,024 $T_2$ / 1,137 $T_3$).
\begin{databox}
\begin{Verbatim}[breaklines=true, breakanywhere=true, fontsize=\footnotesize]
{
  "T1_explicit_confounding": {
    "spurious_feature": "agent visits fridge first",
    "observed_confounder": "task_type = 'cool-and-place'",
    "construction": "Within the 'cool-and-place' task family, agents tend to
                     visit the fridge early AND succeed. We fill memory with
                     these co-occurring records while preserving task_type as
                     an observed variable in memory.",
    "validation": "Conditional independence test (Validation I): correlation
                   between 'fridge_first' and 'success' is 0.62 unconditional,
                   drops to 0.03 after conditioning on task_type.",
    "query_placement": "end-of-trajectory",
    "example_query": "Given the agent visited the fridge first, will this task
                      succeed? (spurious answer: YES)"
  },
  "T2_unmeasured_confounding": {
    "spurious_feature": "agent explores in spiral pattern",
    "hidden_confounder": "agent policy quality (not logged in memory)",
    "construction": "Well-tuned policies produce both spiral exploration and
                     high success rates. Policy quality is never recorded in
                     memory, making it an unobserved confounder. Detected by
                     the FCI extension of PC as a bidirected edge.",
    "validation": "Interventional test (Validation II): trajectories where
                   exploration is held fixed while other attributes vary
                   preserve the spiral-success correlation, confirming an
                   unobserved driver.",
    "query_placement": "mid-trajectory (after step 3)",
    "example_query": "Based on past memories, should the agent keep exploring
                      in a spiral? (spurious answer: YES)"
  },
  "T3_collider_bias": {
    "spurious_feature": "pickup location = countertop 3",
    "collider": "low step count (< 8 steps)",
    "construction": "Both efficient cooling (causal T) and lucky pickup
                     locations (S) produce low step counts. Memory retrieval
                     prefers short trajectories, conditioning on the collider
                     and creating a false pickup-location to success link.",
    "validation": "Interventional test (Validation II): when step count is
                   held fixed, the pickup-location correlation vanishes;
                   when we remove the step-count filter, the correlation
                   disappears naturally.",
    "query_placement": "end-of-trajectory",
    "example_query": "This new task is in kitchen-22 with no countertop 3.
                      Will it still succeed? (spurious answer: NO;
                      correct answer: YES, cooling drives success)"
  }
}
\end{Verbatim}
\end{databox}

\subsection{ScienceWorld}
\label{app:scienceworld}
\noindent\textbf{ScienceWorld Example.} ScienceWorld evaluates whether 
agents can run elementary-school science experiments. The full dataset 
spans 30 subtasks with 7,200 parametric variations in total, where 
each variation changes the target substance, starting location, or 
environmental layout to prevent memorization. Expert trajectories 
average 24.7 steps, with gold action sequences ranging from 15 to 
over 40 steps depending on task complexity. Each example has a task 
such as ``measure the melting point of chocolate,'' followed by a 
trajectory where the agent moves between rooms, uses instruments, and 
records observations. A full example:
\begin{databox}
\begin{verbatim}
{
  "task": "measure the melting point of chocolate",
  "trajectory": [
    {"step": 1, "action": "go to kitchen",
     "observation": "visible: stove, fridge, thermometer, chocolate"},
    {"step": 2, "action": "take chocolate; take thermometer",
     "observation": "you now carry: chocolate, thermometer"},
    {"step": 3, "action": "go to living room; put chocolate on table",
     "observation": "chocolate placed on table"},
    {"step": 4, "action": "activate heat source; wait 3 turns",
     "observation": "chocolate appears to soften"},
    {"step": 5, "action": "use thermometer on chocolate",
     "observation": "thermometer reads 36 degrees C"},
    {"step": 6, "action": "record answer: 36 C",
     "observation": "answer recorded"}
  ],
  "outcome": "SUCCESS (melting point correctly identified)"
}
\end{verbatim}
\end{databox}

\noindent\textbf{Causal Discovery.} Stage~A defines turn-level 
variables $\{A_t, M_t, R_t\}$, session-level context $C = 
\{\text{task\_type}, \text{starting\_room}\}$, and state variables 
tracked by ScienceWorld's state-tracking API (object locations, 
temperatures, container states). For Stage~B, we run PC on 191 expert 
trajectories with priors derived from ScienceWorld's native 
annotations. First, ScienceWorld labels every action as either 
\texttt{critical} (required for goal achievement) or \texttt{auxiliary}; 
we inject must-have edges for critical actions based on the state 
changes observed in the gold trajectory. Second, the scorer's 
pass/fail signal is used as the terminal variable, orienting all edges 
toward it. Third, we run an ablation check as an additional 
edge-direction verifier: if removing action $a$ from the gold 
trajectory breaks state $s$, we confirm $a \to s$. This ablation is 
applied after PC converges to catch edge directions that 
conditional-independence alone leaves ambiguous.

\noindent\textbf{Causality \& Causal Graph.}

\begin{databox}
\begin{verbatim}
{
  "factual_roles": {
    "Treatment": ["apply heat to chocolate"],
    "Covariate": ["room type", "instrument choice", "wait duration"],
    "Mediator":  ["temperature rise in chocolate"],
    "Outcome":   ["correct melting point identified"]
  },
  "counterfactual_roles": {
    "Treatment": ["no heat applied"],
    "Covariate": ["room type", "instrument choice", "wait duration"],
    "Mediator":  ["temperature stays at room level"],
    "Outcome":   ["phase change not observed"]
  },
  "causal_graph": {
    "factual_edges": [
      ["apply heat to chocolate", "temperature rise in chocolate"],
      ["temperature rise in chocolate", "correct melting point identified"]
    ],
    "counterfactual_edges": [
      ["no heat applied", "temperature stays at room level"],
      ["temperature stays at room level", "phase change not observed"]
    ]
  }
}
\end{verbatim}
\end{databox}

\noindent\textbf{Spurious Correlation Construction.} Across all 30 
subtasks, we obtain 6,260 validated pairs (2,041 $T_1$ / 2,315 $T_2$ 
/ 1,904 $T_3$).

\begin{databox}
\begin{Verbatim}[breaklines=true, breakanywhere=true, fontsize=\footnotesize]
{
  "T1_explicit_confounding": {
    "spurious_feature": "agent picks up thermometer first",
    "observed_confounder": "task_type = 'measurement-experiment'",
    "construction": "In measurement tasks, instrument-first pickup co-occurs
                     with success. task_type is logged in memory but memory
                     systems may not condition on it when retrieving.",
    "validation": "Conditional independence test: correlation drops from
                   0.54 to 0.08 after conditioning on task_type.",
    "query_placement": "mid-trajectory (after step 2: instrument pickup)",
    "example_query": "The agent just picked up the thermometer. Will this
                      experiment succeed? (spurious answer: YES)"
  },
  "T2_unmeasured_confounding": {
    "spurious_feature": "agent records observations systematically",
    "hidden_confounder": "agent scientific literacy (not logged)",
    "construction": "Scientifically literate agents both record systematically
                     and succeed more often. Literacy is a latent trait never
                     stored as a state variable. FCI detects this as a
                     bidirected edge between recording-habit and success.",
    "validation": "Interventional test: holding 'systematic recording' fixed
                   while varying other action styles preserves the
                   success pattern, indicating a hidden common cause.",
    "query_placement": "end-of-trajectory",
    "example_query": "Is systematic recording necessary for success?
                      (spurious: YES; correct: NO, heat application is causal)"
  },
  "T3_collider_bias": {
    "spurious_feature": "experiment conducted in kitchen",
    "collider": "experiment completed in under 10 steps",
    "construction": "Both correct heat application (T) and kitchen-starting
                     experiments (S) produce short trajectories. Memory
                     retrieval filters on low step count, conditioning on
                     the collider and creating a false kitchen-success link.",
    "validation": "Interventional test: removing the step-count filter
                   dissolves the kitchen-success correlation.",
    "query_placement": "end-of-trajectory",
    "example_query": "The new task is in a laboratory, not a kitchen.
                      Will it succeed? (spurious: NO; correct: YES,
                      the lab has all required instruments)"
  }
}
\end{Verbatim}
\end{databox}

\subsection{LoCoMo}
\label{app:locomo}

\noindent\textbf{LoCoMo Example.} LoCoMo evaluates whether agents can 
recall and reason over conversations that stretch across weeks or 
months. The released benchmark consists of 10 high-quality 
conversations (a curated subset of an initial 50-conversation 
collection), each averaging 588.2 turns distributed over up to 32 
sessions (mean session length: 21.6 turns). The dataset includes 
1,986 QA annotations spanning five categories (single-hop, multi-hop, 
temporal, open-domain, and adversarial), each with evidence pointers 
to supporting dialogue turns. We use our benchmark at two 
granularities: the full conversation level (10 units) and the session 
level (272 units). A full example:

\begin{databox}
\begin{Verbatim}[breaklines=true, breakanywhere=true, fontsize=\footnotesize]
{
  "sessions": [
    {"session_id": 3, "week": 2, "turns": [
      {"speaker": "Alice", "text": "I've been learning guitar for six months."},
      {"speaker": "Bob",   "text": "Have you tried any specific techniques?"},
      {"speaker": "Alice", "text": "My teacher has me doing daily scales.
                                    It's tedious but I'm getting faster."}
    ]},
    {"session_id": 7, "week": 5, "turns": [
      {"speaker": "Bob",   "text": "How's the guitar coming along?"},
      {"speaker": "Alice", "text": "I can play a full song now!
                                    The scales really paid off."}
    ]}
  ],
  "query": {
    "week": 6,
    "question": "What helped Alice improve at guitar?",
    "answer": "Daily scales practice assigned by her teacher"
  }
}
\end{Verbatim}
\end{databox}

\noindent\textbf{Causal Discovery.} For dialogue data, Stage~A 
identifies session-level variables: per-session topic tags (extracted 
via TF-IDF keyword matching, not an LLM), speaker-mentioned entities, 
and temporal metadata. Turn-level variables $\{Q_t, M_t, R_t, A_t\}$ 
are retained for QA-relevant turns. Session-level context $C$ 
includes speaker identities and their conversational history length. 
For Stage~B, we run PC on all 10 conversation threads (272 sessions 
in aggregate) with three dataset-specific priors. First, 
\textit{temporal precedence}: sessions have timestamps, and candidate 
edges must flow from earlier to later sessions (enforced as a 
must-not-have constraint on backward edges). Second, \textit{LoCoMo's 
evidence annotations}: for each QA pair, the dataset marks which 
earlier sessions contain supporting evidence. We use these 
annotations as must-have edges from evidence sessions to the answer 
variable. Third, \textit{discourse connectives}: we scan turns for 
causal markers (``because'', ``so'', ``therefore'') using the PDTB 
connective lexicon, adding candidate edges that PC then 
verifies through conditional-independence tests. The PC algorithm 
handles the remaining structure discovery on session-to-session 
relationships.

\noindent\textbf{Spurious Correlation Construction.} Across both 
conversation-level and session-level units, we obtain 1,808 validated 
pairs (621 $T_1$ / 538 $T_2$ / 649 $T_3$).

\begin{databox}
\begin{Verbatim}[breaklines=true, breakanywhere=true, fontsize=\footnotesize]
{
  "factual_roles": {
    "Treatment": ["daily scales practice"],
    "Covariate": ["teacher guidance", "practice duration", "instrument quality"],
    "Mediator":  ["finger dexterity improvement"],
    "Outcome":   ["playing ability"]
  },
  "counterfactual_roles": {
    "Treatment": ["no scales practice"],
    "Covariate": ["teacher guidance", "practice duration", "instrument quality"],
    "Mediator":  ["finger dexterity unchanged"],
    "Outcome":   ["playing ability stagnates"]
  },
  "causal_graph": {
    "factual_edges": [
      ["daily scales practice", "finger dexterity improvement"],
      ["finger dexterity improvement", "playing ability"]
    ],
    "counterfactual_edges": [
      ["no scales practice", "finger dexterity unchanged"],
      ["finger dexterity unchanged", "playing ability stagnates"]
    ]
  }
}
\end{Verbatim}
\end{databox}

\noindent\textbf{Spurious Correlation Construction.}

\begin{databox}
\begin{Verbatim}[breaklines=true, breakanywhere=true, fontsize=\footnotesize]
{
  "T1_explicit_confounding": {
    "spurious_feature": "frequent mentions of practice sessions",
    "observed_confounder": "speaker's hobby-talk density per session",
    "construction": "Speakers who discuss hobbies intensely also mention
                     practice frequency more often. Hobby-talk density is
                     logged but rarely used by memory systems for
                     conditioning.",
    "validation": "Conditional independence test: correlation drops from
                   0.48 to 0.06 after conditioning on hobby-talk density.",
    "query_placement": "end-of-trajectory (after final session)",
    "example_query": "Was practice frequency the reason Alice improved?
                      (spurious: YES; correct: NO, the causal chain runs
                      through teacher-assigned scales)"
  },
  "T2_unmeasured_confounding": {
    "spurious_feature": "Alice reports steady progress in every session",
    "hidden_confounder": "Alice's personal motivation (never stated)",
    "construction": "Motivation drives both steady reporting and real
                     improvement, but Alice never explicitly discusses her
                     motivation. FCI detects a bidirected edge between
                     'steady reporting' and 'skill improvement'.",
    "validation": "Interventional test: we rewrite dialogue variants where
                   reporting frequency is held fixed; skill improvement
                   still correlates, indicating a hidden driver.",
    "query_placement": "mid-trajectory (after session 5)",
    "example_query": "Alice reports steady progress. Does this mean her
                      practice method is optimal? (spurious: YES)"
  },
  "T3_collider_bias": {
    "spurious_feature": "Alice watches YouTube guitar tutorials",
    "collider": "session has heavy music discussion",
    "construction": "Both daily scales practice and YouTube tutorials cause
                     sessions to contain heavy music discussion. Memory
                     retrieval prefers music-heavy sessions, conditioning on
                     the collider and falsely linking tutorials to practice.",
    "validation": "Interventional test: sessions sampled without the
                   music-discussion filter show no tutorials-practice link.",
    "query_placement": "mid-trajectory (after session 5)",
    "example_query": "Given Alice watches YouTube tutorials, will her next
                      session discuss scales progress? (spurious: YES)"
  }
}
\end{Verbatim}
\end{databox}

\subsection{WebShop}
\label{app:webshop}

\noindent\textbf{WebShop Example.} WebShop evaluates whether agents 
can complete online shopping tasks through multi-turn search and 
comparison. The full environment contains 1.18 million real-world 
products scraped from Amazon, paired with 12,087 crowd-sourced text 
instructions specifying product requirements. We use 81 logged 
demonstration trajectories with an average length of 11.5 steps 
(search, browse, click, customize, purchase). Each example has a 
natural-language instruction with product constraints, followed by a 
trajectory where the agent searches the catalog, inspects items, and 
makes a purchase. A full example:

\begin{databox}
\begin{verbatim}
{
  "instruction": "I need a waterproof hiking backpack under $80
                  with at least 30L capacity",
  "trajectory": [
    {"step": 1, "action": "search: waterproof hiking backpack",
     "observation": "6 results returned"},
    {"step": 2, "action": "browse results",
     "observation": "Item A: $75, 35L, 4.2 stars;
                     Item B: $68, 28L, 4.5 stars"},
    {"step": 3, "action": "click Item A",
     "observation": "reviews mention: 'holds up in heavy rain'"},
    {"step": 4, "action": "purchase Item A",
     "observation": "order placed"}
  ],
  "outcome": "SUCCESS (all constraints satisfied)"
}
\end{verbatim}
\end{databox}

\noindent\textbf{Causal Discovery.} WebShop's structured catalog makes 
Stage~A almost fully deterministic. We parse each instruction into 
atomic constraint predicates with a small regex grammar, ``under 
\$80'' becomes \texttt{price $<$ 80}, ``waterproof'' becomes 
\texttt{has\_attribute(waterproof)}. Per-trajectory variables include 
the instruction constraints, the agent's search query $Q_t$, retrieved 
item lists $M_t$, click/purchase actions $A_t$, and product attributes 
pulled from the catalog. Session-level context $C$ includes the 
instruction's constraint conjunction. For Stage~B, we run PC on 81 
logged trajectories with two priors from WebShop's schema. First, 
\textit{must-have edges}: a product attribute that satisfies an 
instruction predicate must be a cause of the purchase outcome when 
selected. Second, \textit{must-not-have edges}: attributes absent 
from the instruction (brand, category tag) cannot directly cause 
purchase success; they can only act through the 
constraint-satisfaction mediator. The PC algorithm then resolves the 
remaining structure around ratings, category tags, and rankings, 
which are not tied to constraints but may co-occur with successful 
purchases.

\noindent\textbf{Causality \& Causal Graph.}

\begin{databox}
\begin{Verbatim}[breaklines=true, breakanywhere=true, fontsize=\footnotesize]
{
  "factual_roles": {
    "Treatment": ["select product matching all requirements"],
    "Covariate": ["customer rating", "product category tag", "brand"],
    "Mediator":  ["constraint satisfaction (capacity + waterproof + price)"],
    "Outcome":   ["purchase success"]
  },
  "counterfactual_roles": {
    "Treatment": ["select product missing at least one requirement"],
    "Covariate": ["customer rating", "product category tag", "brand"],
    "Mediator":  ["one or more constraints violated"],
    "Outcome":   ["purchase failure"]
  },
  "causal_graph": {
    "factual_edges": [
      ["select product matching all requirements",
       "constraint satisfaction (capacity + waterproof + price)"],
      ["constraint satisfaction (capacity + waterproof + price)",
       "purchase success"]
    ],
    "counterfactual_edges": [
      ["select product missing at least one requirement",
       "one or more constraints violated"],
       ["one or more constraints violated", "purchase failure"]
    ]
  }
}
\end{Verbatim}
\end{databox}

\noindent\textbf{Spurious Correlation Construction.} Across the 81 
trajectories and 12,087 instructions, we obtain 10,545 validated 
pairs (3,512 $T_1$ / 3,187 $T_2$ / 3,846 $T_3$).

\begin{databox}
\begin{Verbatim}[breaklines=true, breakanywhere=true, fontsize=\footnotesize]
{
  "T1_explicit_confounding": {
    "spurious_feature": "customer rating above 4.3 stars",
    "observed_confounder": "product brand",
    "construction": "Reputable brands both get high ratings and ship products
                     that match common constraints. Brand is observable in
                     the catalog but rarely used by memory systems as a
                     retrieval key.",
    "validation": "Conditional independence test: rating-success correlation
                   drops from 0.51 to 0.09 after conditioning on brand.",
    "query_placement": "mid-trajectory (after step 2: browsing results)",
    "example_query": "Item C has 4.7 stars. Should the agent pick it over
                      Item A? (spurious: YES; correct: only if constraints
                      also match)"
  },
  "T2_unmeasured_confounding": {
    "spurious_feature": "product description uses 'premium' adjectives",
    "hidden_confounder": "seller's marketing budget (not in catalog)",
    "construction": "Sellers with large marketing budgets produce both
                     polished descriptions and items that meet standard
                     constraints. Marketing budget is never exposed to the
                     agent. FCI detects a bidirected edge.",
    "validation": "Interventional test: descriptions held fixed across
                   sellers with varying budgets show no correlation with
                   purchase success.",
    "query_placement": "end-of-trajectory",
    "example_query": "This item's description says 'premium quality'.
                      Will the purchase succeed? (spurious: YES)"
  },
  "T3_collider_bias": {
    "spurious_feature": "price in the bottom quartile",
    "collider": "product appears in top 3 search results",
    "construction": "Both matching requirements (T) and low pricing (S)
                     push products into the top 3. Retrieval filters on
                     top-ranked items, conditioning on the collider and
                     falsely linking price to fit.",
    "validation": "Interventional test: sampling from the full result list
                   dissolves the price-fit correlation.",
    "query_placement": "mid-trajectory (after the search step)",
    "example_query": "This cheap item is top-ranked. Does it meet the
                      capacity requirement? (spurious: YES)"
  }
}
\end{Verbatim}
\end{databox}

\subsection{Additional Results for Future-Step Accuracy}
\label{app:rq2_extra}

\begin{figure}[h]
\centering
\includegraphics[width=\textwidth]{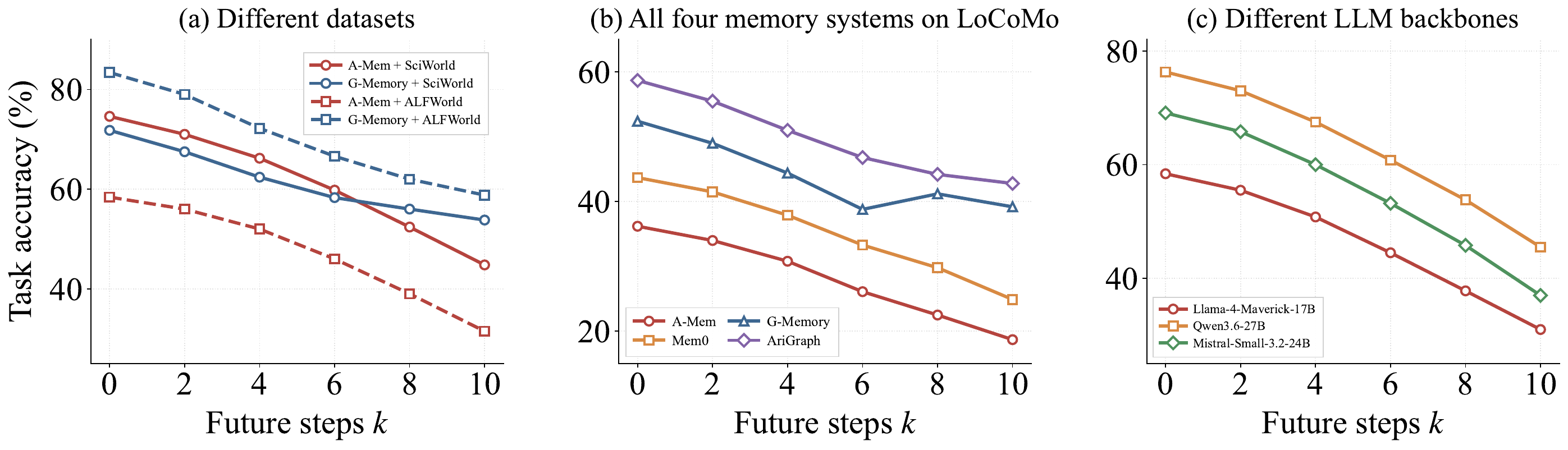}
\caption{Additional future-step accuracy results after spurious retrievals. Setup follows Figure~\ref{fig:rq2}.}
\label{fig:rq2_extra}
\end{figure}

Figure~\ref{fig:rq2_extra} reports accuracy decline curves across additional datasets, memory systems, and LLM backbones that are not shown in Figure \ref{fig:rq2}. The findings and insights are aligned with our discussion in \S\ref{ssec:eval-spur}.

%% file: appendix/calibrate-embed.tex
\section{Calibration Details and Analysis}


\subsection{Justifying the Calibration Formulation}
\label{app:formulation-justification}

This subsection justifies that Eqs.~(\ref{eq:decomposition})--(\ref{eq:stability})
are not standalone assumptions but consequences of two mild and verifiable
properties of the encoder and the trajectory: (\textit{Property 1}) the encoder is locally
smooth in a neighborhood of each step's context, and (\textit{Property 2}) within a single
step, every entry shares the same context value. Both hold for the
encoders used by the memory systems we evaluate.

\textbf{Overview.}
Eq.~(\ref{eq:decomposition}) is a Taylor expansion under encoder
smoothness; Eq.~(\ref{eq:write-residual}) is the orthogonal projection
that consistently removes the shared component, with the Welford form
algebraically identical to the batch mean; Eq.~(\ref{eq:cf-score}) is
the linearity of the inner product; Eq.~(\ref{eq:stability}) is a
robust seminorm of the projection onto the non-causal subspace, with
$\sigma(m)=0$ characterizing causal candidates exactly. The four
formulas form a single derivation chain rather than four independent
modeling choices.

\textbf{Setup.}
Let $E:\mathcal{X}\!\times\!\mathcal{C}\!\to\!\mathbb{R}^d$ be the encoder
mapping an entry's content $x$ and its step context $c$ to a representation.
Within step $s$, all $n_s$ entries share the same context value $c^{(s)}$,
so each raw representation is $h_m = E(x_m,\, c^{(s)})$ for some
content $x_m$ specific to entry $m$.

\textbf{Property 1 ($\Rightarrow$) Eq.~(\ref{eq:decomposition}).}
Fix a reference content $\bar{x}$ and a reference context $\bar{c}$.
Assuming $E$ is differentiable in its second argument in a neighborhood
of $\bar{c}$, a first-order Taylor expansion of $E$ around $(\bar{x},\bar{c})$
in the context coordinate gives
\begin{equation}
  \label{eq:taylor}
  h_m = \underbrace{E(x_m,\bar{c})}_{\text{depends only on }x_m}
        + \underbrace{J_c E(\bar{x},\bar{c})\bigl(c^{(s)}-\bar{c}\bigr)}_{\text{depends only on }c^{(s)}}
        + r_m
\end{equation}
where $J_c E$ is the Jacobian of $E$ \cite{mathai1997jacobians} with respect to $c$ and $r_m$
collects the higher-order remainder plus the cross term
$\bigl[J_c E(x_m,\bar{c}) - J_c E(\bar{x},\bar{c})\bigr](c^{(s)}-\bar{c})$.
Defining
$h_m^{\text{causal}} := E(x_m,\bar{c})$,
$h_m^{\text{context}} := J_c E(\bar{x},\bar{c})\bigl(c^{(s)}-\bar{c}\bigr)$,
and $\varepsilon_m := r_m$ recovers Eq.~(\ref{eq:decomposition}) exactly.
Two observations follow:
(i) $h_m^{\text{context}}$ depends on the step but not on $m$, so it is
\emph{constant across all entries within step $s$};
(ii) $\varepsilon_m$ is $O(\|c^{(s)}-\bar{c}\|^2)$ and $O(\|x_m-\bar{x}\|\!\cdot\!\|c^{(s)}-\bar{c}\|)$
in operator norm, i.e., second-order small whenever the encoder is well-behaved
on the trajectory. The decomposition is therefore a derivation, not a postulate.

\textbf{Property 2 ($\Rightarrow$) Eq.~(\ref{eq:write-residual}).}
Within step $s$, write $h_m = a_m + b^{(s)} + r_m$ with
$a_m := h_m^{\text{causal}}$, $b^{(s)} := h_m^{\text{context}}$ shared
across entries, and $\mathbb{E}_m[r_m]=0$ over the entries in the step
(this holds whenever residual contents $x_m$ are exchangeable within
the step, which is the same condition that justifies treating the step
as a single unit in the trajectory).

\smallskip\noindent\textit{Step mean is the unique unbiased linear estimator of $b^{(s)}$.}
The batch step mean is
$\bar{\mu}^{(s)} = \tfrac{1}{n_s}\sum_{m=1}^{n_s} h_m = \bar{a}^{(s)} + b^{(s)} + \bar{r}^{(s)}$,
where $\bar{a}^{(s)} := \tfrac{1}{n_s}\sum_m a_m$ and
$\bar{r}^{(s)} := \tfrac{1}{n_s}\sum_m r_m$.
If contents are i.i.d.\ within the step with $\mathbb{E}[a_m]=\bar{a}$ and
$\mathrm{Cov}(a_m)=\Sigma_a$, then by the Gauss-Markov theorem applied
to the linear model $h_m = b^{(s)} + (a_m - \bar{a}) + r_m$ with constant
regressor $1$, the sample mean is the best linear unbiased estimator of
$b^{(s)} + \bar{a}$, with variance $(\Sigma_a + \Sigma_r)/n_s \to 0$ as
$n_s \to \infty$. Hence $\bar{\mu}^{(s)} \to b^{(s)} + \bar{a}$ in probability,
and $h_m - \bar{\mu}^{(s)} \to (a_m - \bar{a}) + r_m$, i.e., the residual
recovers the entry-specific causal content up to the global constant $\bar{a}$,
which is absorbed by the encoder's translation invariance and does not
affect inner-product retrieval (\S\ref{ssec:retrieve}).

\smallskip\noindent\textit{Welford recurrence equals the batch mean.}
The update in Eq.~(\ref{eq:write-residual}),
$\mu^{(s)} \!\leftarrow\! \mu^{(s)} + \tfrac{1}{n+1}(h_m - \mu^{(s)})$,
is the textbook Welford one-pass recurrence: by induction, after $n$ updates
$\mu^{(s)} = \tfrac{1}{n}\sum_{i=1}^{n} h_i$. The residualization
$\tilde{h}_m = h_m - \mu^{(s)}$ thus equals the leave-one-out mean
deviation, which is the batch residual up to a $1/n_s$ shrinkage that
vanishes as the step fills.

\smallskip\noindent\textit{Residualization is an orthogonal projection.}
View the step's $n_s$ representations as a single tuple
$\mathbf{h}^{(s)} := (h_1, \ldots, h_{n_s}) \in \mathbb{R}^{n_s d}$,
equipped with the standard inner product
$\langle \mathbf{h}, \mathbf{h}' \rangle = \sum_{m=1}^{n_s} h_m^\top h_m'$.
Decompose this space orthogonally as
$\mathbb{R}^{n_s d} = S \oplus S^\perp$, where
$S := \{(v, v, \ldots, v) : v \in \mathbb{R}^d\}$ is the
\emph{shared-direction} subspace of tuples constant across entries
(isomorphic to $\mathbb{R}^d$), and $S^\perp$ is its orthogonal
complement, the subspace of tuples whose entries sum to zero. The
orthogonal projector onto $S$ sends $\mathbf{h}^{(s)}$ to
$(\mu^{(s)}, \ldots, \mu^{(s)})$ with $\mu^{(s)} = \tfrac{1}{n_s}\sum_m h_m$,
so the projector onto $S^\perp$ sends each entry to $h_m - \mu^{(s)}$.
This is exactly the operation of Eq.~(\ref{eq:write-residual}). Any
component that is shared across the step ($h^{\text{context}}$ in full
and $\bar{a}$ in part) lies in $S$ and is removed exactly; any
entry-specific deviation lies in $S^\perp$ and passes through unchanged.
As an orthogonal projection, residualization is the unique linear
operator on $\mathbb{R}^{n_s d}$ that annihilates the shared-direction
subspace while preserving every direction orthogonal to it.

\textbf{($\Rightarrow$) Eq.~(\ref{eq:cf-score}): closed form is exact.}
For inner-product scoring $\phi(q,m) = \langle q, \tilde{h}_m \rangle$,
linearity gives
$\phi(q + \delta_\ell v_\ell,\, m) = \langle q, \tilde{h}_m \rangle
 + \delta_\ell \langle v_\ell, \tilde{h}_m \rangle$,
which is Eq.~(\ref{eq:cf-score}). No approximation is involved; the
closed form is an algebraic identity. For cosine scoring with normalized
queries, the same identity holds up to a renormalization factor
$1 + O(\delta_\ell)$, which is absorbed into the per-direction scale
$\delta_\ell$ and does not affect the ranking.

\textbf{($\Rightarrow$) Eq.~(\ref{eq:stability}): when $\sigma(m)=0$, $m$ is causal.}
Let $\mathcal{S}_{\text{nc}} = \mathrm{span}(v_1,\ldots,v_L)$ and write
$\tilde{h}_m = \tilde{h}_m^{\parallel} + \tilde{h}_m^{\perp}$ with
$\tilde{h}_m^{\parallel} \in \mathcal{S}_{\text{nc}}$ and
$\tilde{h}_m^{\perp} \in \mathcal{S}_{\text{nc}}^{\perp}$.
Then $\langle v_\ell, \tilde{h}_m \rangle = \langle v_\ell, \tilde{h}_m^{\parallel} \rangle$
for every $\ell$, and
\begin{equation}
  \sigma(m) = 0
  \;\Longleftrightarrow\;
  \langle v_\ell, \tilde{h}_m^{\parallel} \rangle = 0 \;\;\forall\,\ell
  \;\Longleftrightarrow\;
  \tilde{h}_m^{\parallel} = 0
  \;\Longleftrightarrow\;
  \tilde{h}_m \in \mathcal{S}_{\text{nc}}^{\perp}
\end{equation}
So $\sigma(m) = 0$ holds exactly when $m$'s representation lies entirely
in the orthogonal complement of the non-causal subspace, i.e., the causal
subspace by definition (Appendix~\ref{app:projection-update}). For
$\sigma(m) > 0$, the magnitude $\sigma(m)$ equals the median absolute
projection coordinate onto $\mathcal{S}_{\text{nc}}$ and is therefore a
seminorm of $\tilde{h}_m^{\parallel}$. We choose median rather than mean
because a single dominant non-causal direction (a strong confounder)
would otherwise inflate $\sigma$ for every candidate, washing out
ranking signal; the median is robust to such outliers and preserves
relative comparisons across candidates. Concretely, if exactly one
$v_{\ell^\star}$ carries large projection for all $m$, the mean shifts
uniformly while the median is unaffected, so the ranking induced by
$\sigma$ remains the ranking induced by the remaining directions.


\subsection{Content-Novelty Write Criterion}
\label{app:write-criterion}

The ``Two parts, one stage'' paragraph of \S\ref{ssec:write} pairs
residualization (which removes step-shared bias from each stored
representation) with a write criterion that decides \emph{whether} to
store $\tilde{h}_m$ at all. This subsection details the criterion.
Its job is narrow but essential: prevent the write event from being
conditioned on any downstream outcome $O$, so that ``being in memory''
is not a collider that would otherwise open a spurious path between
unrelated entries ($\texttt{T}_\texttt{3}$, \S\ref{ssec:background}).

\textbf{Criterion (embedding memories).}
Store $\tilde{h}_m$ if and only if it is not already represented by an
existing entry, i.e., if the maximum cosine similarity between
$\tilde{h}_m$ and every stored entry is strictly less than $1.0$ (up to
floating-point precision). This is a comparison between the new entry
and the existing memory, not a tunable threshold. The required
nearest-neighbour lookup is already performed by most memory systems
for deduplication, so the criterion adds no asymptotic cost on top of
the base write operation. No information about $O$ is read at any point.

\textbf{Criterion (graph memories).}
The analogous check is structural rather than metric: insert the new
node only when it would form at least one edge to a node from a
\emph{different} session. A node whose only candidate edges land within
its own session adds no cross-session structure and is discarded as
informationally redundant. The check costs $O(\deg_\text{new})$, again
within the cost of the base edge-formation step, and again reads no
outcome variable. Full graph-memory adaptation is in
Appendix~\ref{app:graph-adapt}.

\textbf{Why this severs the collider.}
Under the criterion, the write event is caused only by the content of
$\tilde{h}_m$ relative to existing memory; the outcome $O$ has no
arrow into ``written.'' The collider structure
$X \to \text{written} \leftarrow O$ that otherwise creates a spurious
$X$--$O$ association in the stored sample no longer exists, so any
retrieval that operates on the resulting memory is not implicitly
conditioning on $O$. This is a structural fix at the level of the
write policy, complementary to the representational fix performed by
residualization (Eq.~\eqref{eq:write-residual}); the two together
give the write stage full coverage of $\texttt{T}_\texttt{1}$, $\texttt{T}_\texttt{2}$, and $\texttt{T}_\texttt{3}$
(Appendix~\ref{app:write-analysis}).


\subsection{Step Mean: Implementation Details}
\label{app:session-mean}

This subsection collects the practical details an implementer needs to
deploy Eq.~\eqref{eq:write-residual} in an embedding-based memory.
The mathematical justification (Welford = batch mean, residualization
as orthogonal projection, consistency of $\mu^{(s)}$) is given in
Appendix~\ref{app:formulation-justification}; the definition of a step
and a worked example are in the textbox in \S\ref{ssec:write}.

\textbf{Initialization.}
At the start of each new step $s$, set
$\mu^{(s)} = \mathbf{0} \in \mathbb{R}^d$ and entry counter $n = 0$.

\textbf{Update rule.}
When the next entry of step $s$ arrives with raw representation $h_m$,
apply Eq.~\eqref{eq:write-residual} in this order:
\begin{enumerate}
  \item Compute $\tilde{h}_m = h_m - \mu^{(s)}$ using the
        \emph{current} mean (the mean over the $n$ entries already
        written in this step).
  \item Store $\tilde{h}_m$ in the ANN index.
  \item Update
        $\mu^{(s)} \leftarrow \mu^{(s)} + \tfrac{1}{n+1}(h_m - \mu^{(s)})$,
        then increment $n \leftarrow n+1$.
\end{enumerate}
The order matters: residualization uses the pre-update mean, so
$\tilde{h}_m$ measures how $m$ departs from the context already
established by earlier entries in the step. Reversing steps 1 and 3
would subtract $h_m$ partly from itself.

\textbf{Counter and state.}
The counter $n$ tracks entries written in the current step, not the
step index $s$ itself; multiple entries can be written within one step,
and $n$ increments per entry. The cost is $O(d)$ time and $O(d)$
storage per active step, regardless of memory size $N$.

\textbf{Step closure.}
Once the step ends, $\mu^{(s)}$ and $n$ are discarded. They are
write-time artifacts only; at retrieval time, the ANN index contains
$\tilde{h}_m$ vectors and behaves identically to an unmodified index.

\textbf{Cross-step confounders.}
A latent confounder $U$ whose value differs across steps (e.g., a
persistent user preference or a system-wide bias) is not removed by
$\mu^{(s)}$, since each $\mu^{(s)}$ resets at the step boundary. Such
$U$ shifts the entries of each step in a $U$-aligned direction whose
magnitude varies between steps, which is exactly the signature
identified by retrieve-stage stability filtering
(\S\ref{ssec:retrieve}). Within-step and cross-step confounders are
therefore handled by complementary mechanisms at the two stages.


\subsection{How Write-Stage Calibration Addresses All Three Spurious Types}
\label{app:write-analysis}

Write-stage calibration has two operations (\S\ref{ssec:write}):
residualization (Eq.~\eqref{eq:write-residual}), which transforms the
stored representation, and the content-novelty criterion
(Appendix~\ref{app:write-criterion}), which decides whether the
representation is stored at all. The two cover complementary failure
modes. Residualization handles spurious types whose signature is a
\emph{direction} in representation space that can be identified and
removed; the criterion handles a type whose signature is the
\emph{selection mechanism} that decides which entries enter memory.
Each subsection below names the structural feature of the type and
points to the operation that neutralizes it.

\textbf{Explicit confounding ($\texttt{T}_\texttt{1}$).}
An observed variable $C$ (e.g., task category, tool type, domain) is a
common cause of both what the agent writes and what action it later
takes, so retrieval conflates ``causally relevant to the query'' with
``written under the same value of $C$ as the query.''

The structural feature that makes $\texttt{T}_\texttt{1}$ tractable is that $C$ is
\emph{step-scoped}: within a single step the value of $C$ is fixed
(P2 in Appendix~\ref{app:formulation-justification}). Under the
decomposition Eq.~\eqref{eq:decomposition}, the contribution of $C$
to $h_m^{\text{context}}$ is therefore \emph{shared} across every
entry of the step. The proof in
Appendix~\ref{app:formulation-justification} (Eq.~5 onward) shows that
the step mean $\mu^{(s)}$ converges to this shared component, so
subtracting $\mu^{(s)}$ in Eq.~\eqref{eq:write-residual} removes the
$C$-induced term from every $\tilde{h}_m$, regardless of whether $C$
was ever explicitly recorded.

The practical consequence: two entries with the same causal content
but written under different values of $C$ end up with nearby
calibrated representations (the $C$-induced part has been subtracted
from both), while two entries written under the same $C$ but with
genuinely different content are no longer pulled together. The
spurious path that ran through $C$ is severed at storage time.

\textbf{Unmeasured confounding ($\texttt{T}_\texttt{2}$).}
A latent variable $U$ (e.g., a persistent user preference, an implicit
topic bias, a system-wide conversational style) shapes both writes and
later retrievals, but $U$ is never observed. Direct adjustment is
unavailable because we cannot condition on something we cannot see.

$\texttt{T}_\texttt{2}$ looks structurally identical to $\texttt{T}_\texttt{1}$ from the residualization's point
of view. What matters for Eq.~\eqref{eq:write-residual} is not whether
$U$ is named, but whether $U$ acts coherently within a step, which
holds for any confounder whose value is fixed at the granularity of a
single agent episode. When this holds, $U$'s contribution to
$h_m^{\text{context}}$ is shared across all entries of the step
(\textit{Property 2} in Appendix \ref{app:formulation-justification}), accumulates in $\mu^{(s)}$ by the same Gauss-Markov
argument used for $\texttt{T}_\texttt{1}$
(Appendix~\ref{app:formulation-justification}), and is removed by the
subtraction. This is the core advantage of removing a sufficient
statistic rather than conditioning on named variables: the operation
$\tilde{h}_m = h_m - \mu^{(s)}$ targets \emph{any} shared directional
bias in the step, observed or not.

The only requirement for $\texttt{T}_\texttt{2}$ coverage is the within-step coherence
above. Cross-step components of $U$ (whose value shifts between steps)
are picked up later by retrieve-stage stability filtering
(\S\ref{ssec:retrieve}); the two stages handle within- and cross-step
parts of the same latent confounder.

\textbf{Collider bias ($\texttt{T}_\texttt{3}$).}
A collider arises when two variables both cause a third, and conditioning on the
third makes the two parents conditionally dependent. In the memory setting, the
collider is the \emph{stored entry} $C^\star$: when an agent stores an entry only
if the surrounding trajectory reaches a successful outcome $Y$, both the entry's
content $X$ and $Y$ are causes of $C^\star$. Every downstream retrieval implicitly
conditions on $C^\star$ being present in memory, which is the conditioning event
that activates the collider. Any retrieval that sees only
written entries implicitly conditions on this collider, opening a
spurious path between content and $O$: entries co-appear in memory
not because their causal content is similar but because they were
co-selected by the same outcome filter.

Unlike $\texttt{T}_\texttt{1}$ and $\texttt{T}_\texttt{2}$, the signature of $\texttt{T}_\texttt{3}$ is not a direction in
representation space; it is the rule that decides which entries enter
memory at all. Subtracting $\mu^{(s)}$ from a representation cannot
correct a sample that is already collider-conditioned before any
representation transform is applied. The fix has to operate at the
level of the write \emph{policy}, which is what the content-novelty
criterion does: an entry is stored based only on its informational
novelty relative to what is already in memory
(Appendix~\ref{app:write-criterion}), with no reference to $O$. Under
this rule, ``being written'' is caused only by the entry's own
content, the collider arrow from $O$ into ``written'' no longer
exists, and the spurious co-appearance pattern is structurally
prevented rather than corrected after the fact. The fix adds zero
asymptotic cost on top of the base write operation.



\subsection{Online Maintenance of the Non-Causal Subspace}
\label{app:projection-update}

\S\ref{ssec:retrieve} defines $\mathcal{S}_{\text{nc}}$ as the span of
axes whose projection $\langle v, \tilde{h}\rangle$ varies more
between steps than within steps. This subsection gives the streaming
recipe for identifying $\mathcal{S}_{\text{nc}}$ on the fly. All
statistics are accumulated on the calibrated representations
$\tilde{h}_m$ already in memory; no past entry is re-stored, no
outcome label is required, and recovery of the subspace inherits the
consistency guarantees from Appendix~\ref{app:formulation-justification}.

\textbf{Within-step covariance.}
For step $s$, let $\bar{\tilde{h}}^{(s)}$ denote the mean of calibrated
representations written in step $s$ ($\bar{\tilde{h}}^{(s)} \to 0$ as
the step fills, by Eq.~\eqref{eq:write-residual}, but is nonzero
for the partial step). The within-step covariance contribution from
step $s$ is the empirical covariance of
$\tilde{h}_m - \bar{\tilde{h}}^{(s)}$ over $m \in s$. Pooling these
contributions across steps gives $\Sigma_{\text{within}}$, which we
maintain as a running sum of outer products, updated $O(d^2)$ per
write. Total storage is $O(d^2)$, independent of memory size $N$.

\textbf{Between-step covariance.}
At step closure, contribute $\bar{\tilde{h}}^{(s)}$ (weighted by step
size $n_s$) to a running covariance over step means. The result is
$\Sigma_{\text{between}}$, also $O(d^2)$ in storage and $O(d^2)$ per
step closure to update.

\textbf{Subspace recomputation.}
A direction $v$ is non-causal when its between-step variance dominates
its within-step variance, i.e., when
$\tfrac{v^\top \Sigma_{\text{between}} v}{v^\top \Sigma_{\text{within}} v}$
is large. Maximizing this ratio is the standard generalized eigenvalue
problem
\[
  \Sigma_{\text{between}}\, v \;=\; \lambda\, \Sigma_{\text{within}}\, v
\]
solved in $O(d^3)$ by any standard library (e.g., LAPACK
\texttt{dsygv}). We recompute the eigendecomposition every $B$ writes
(with $B \gg d$), so the $O(d^3)$ cost is amortized to $O(d^3/B)$ per
write and never dominates the $O(d^2)$ per-write covariance updates.
The non-causal axes are the leading generalized eigenvectors, separated
from the causal axes by the elbow in the eigenvalue spectrum; the count
$L$ is data-determined rather than tuned. We store the resulting
orthonormal basis $\{v_1,\ldots,v_L\}$.

\textbf{Why the criterion is the right one.}
The variance-ratio criterion isolates exactly the directions populated
by step-shared confounders that survive write-stage residualization.
Within-step shared structure has been removed by Eq.~\eqref{eq:write-residual}
(P2 in Appendix~\ref{app:formulation-justification}), so any axis with
non-trivial $v^\top \Sigma_{\text{between}} v$ but vanishing
$v^\top \Sigma_{\text{within}} v$ tracks a confounder whose value
changes between steps but is fixed within a step, i.e., precisely the
$\texttt{T}_\texttt{1}$/$\texttt{T}_\texttt{2}$ patterns described in Appendix~\ref{app:retrieve-analysis}.
Causal directions, by construction, vary entry-to-entry within a step
and contribute mostly to $\Sigma_{\text{within}}$, so they receive small
generalized eigenvalues and fall outside $\mathcal{S}_{\text{nc}}$.

\textbf{Stability of the subspace.}
Past entries were calibrated using the step mean $\mu^{(s)}$ available
at their write time, and that calibration is never undone. The
non-causal subspace is stable as long as the distribution of steps does
not shift pathologically (the standard stationarity assumption of
online learning). When step-distribution drift is detected (e.g., via
covariance norm changes exceeding a streaming threshold),
$\mathcal{S}_{\text{nc}}$ is recomputed and used for subsequent
retrievals; historical entries are not rewritten.

\subsection{Counterfactual Stability Selection: Full Details}
\label{app:stability}

\S\ref{ssec:retrieve} defines the stability statistic
$\sigma(m)$ and uses it to rerank the base retriever's top-$k$
candidates. This subsection collects the implementation details:
how the perturbation magnitudes are set, why the closed form makes
the procedure cheap, and how the rerank composes with the base
retriever.

\textbf{Perturbation magnitude.}
For each non-causal direction $v_\ell$ we set
\[
  \delta_\ell \;=\; \sqrt{\tfrac{1}{N}\sum_{m} \langle v_\ell, \tilde{h}_m \rangle^2}
\]
the standard deviation of projections onto $v_\ell$ across all entries
in memory. This puts the perturbation on the natural scale of variation
along $v_\ell$, so the test measures sensitivity relative to the actual
spread of stored representations rather than relative to an arbitrary
constant. No hyperparameter is introduced; $\delta_\ell$ is determined
by the data and maintained as a running variance estimate, updated in
$O(L)$ per write.

\textbf{Score under counterfactual queries (closed form).}
A counterfactual query is $q^{(\ell)} = q + \delta_\ell v_\ell$. For
inner-product retrieval scoring, linearity gives the closed form
proven in Appendix~\ref{app:formulation-justification}:
\[
  \phi\bigl(q^{(\ell)}, m\bigr) \;=\; \phi(q, m)
  \;+\; \delta_\ell \langle v_\ell, \tilde{h}_m \rangle
\]
All $L$ counterfactual scores therefore follow from the base score plus
$L$ inner products $\langle v_\ell, \tilde{h}_m \rangle$ per candidate,
each $O(d)$. Total cost: $O(Lkd)$ for $k$ candidates and $L$ directions,
with no rescoring pass and no additional retrieval round.

\textbf{Stability statistic.}
The statistic $\sigma(m) = \texttt{median}_{\ell=1}^{L}
|\langle v_\ell, \tilde{h}_m \rangle|$ is the median absolute score
change across the $L$ counterfactuals (up to the per-direction scale
$\delta_\ell$). Small $\sigma(m)$ means $m$'s rank is insensitive to
non-causal perturbations of the query, i.e., the retrieval was driven
by causal content; large $\sigma(m)$ means $m$'s rank rested on a
spurious direction. Appendix~\ref{app:formulation-justification} proves
that $\sigma(m) = 0$ holds exactly when $\tilde{h}_m$ lies in the
orthogonal complement of $\mathcal{S}_{\text{nc}}$, i.e., the causal
subspace, making the test conservative on true causal candidates.
The median (rather than mean) keeps a single dominant non-causal
direction from inflating $\sigma$ uniformly across candidates and
washing out ranking signal; the same proof gives a precise statement
of this robustness.

\textbf{Rerank rule.}
Given the base retriever's top-$k$ set $\mathcal{M}_k(q)$, we sort its
members in increasing order of $\sigma(m)$ and return them in that
order. The same $k$ candidates are returned (nothing is dropped)
and the most stable ones come first. No threshold is introduced; the
ordering is internal to the candidate set. This composes with any
downstream consumer that reads candidates in rank order (an LLM prompt,
a beam search, a tool router): stable candidates appear earlier and
exert more influence than spurious ones, while the candidate set's
recall is unchanged.

\textbf{Architecture-agnostic application.}
The rerank uses only two pieces of information per candidate: the
calibrated representation $\tilde{h}_m$ and the basis
$\{v_1,\ldots,v_L\}$. It is therefore independent of how
$\mathcal{M}_k(q)$ was formed by nearest-neighbour search over an ANN
index, graph traversal, hybrid lexical-semantic retrieval, or any
other base retriever. The base retriever is never re-invoked; only
arithmetic on the post-retrieval candidate set is performed.
Adaptations to specific graph-memory architectures (G-Memory's
bi-directional traversal, AriGraph's semantic+episodic search) are in
Appendix~\ref{app:graph-adapt}.

\subsection{Boundary Expansion: Residual Collider at the Retriever}
\label{app:expand}

The content-novelty write criterion (\S\ref{ssec:write},
Appendix~\ref{app:write-criterion}) already prevents the write event
from being conditioned on a downstream outcome $O$, so the stored
memory itself is not collider-conditioned. There is, however, one
residual case the write stage cannot reach: when the \emph{base
retriever}, independent of how memory was written, applies its own
outcome-conditional gate before returning candidates. For example,
some agentic systems restrict graph traversal to nodes linked to
successful trajectories, or apply a task-success threshold before
returning the top-$k$. Under such a retriever, $\mathcal{M}_k(q)$ is
collider-conditioned even when the underlying memory is clean.

\textbf{When the stability test handles this on its own.}
If the retriever's outcome gate has been applied non-uniformly across
steps in memory (e.g., some steps were filtered, others were not, or
the gate's threshold varied) the filter direction acquires non-zero
between-step variance and surfaces in $\mathcal{S}_{\text{nc}}$
(Appendix~\ref{app:retrieve-analysis}). The stability test then demotes
filter-driven candidates exactly as it does for $\texttt{T}_\texttt{1}$ and $\texttt{T}_\texttt{2}$, with no
extra machinery.

\textbf{The case that needs an explicit fix.}
The residual case is a retriever whose gate is applied uniformly to
every step in memory. The filter direction then has zero between-step
variance within memory, leaves no signature in
$\mathcal{S}_{\text{nc}}$, and the stability test cannot demote the
filter-driven candidates because there is nothing in memory to compare
against.

\textbf{Boundary expansion.}
The remedy is to retrieve $\mathcal{M}_{k+k'}(q)$ from the
\emph{unfiltered} memory, bypassing the outcome-conditional gate, and
apply the stability test on the full $\mathcal{M}_{k+k'}$ set. Entries
on both sides of the filter boundary are now visible to the test, so
projections onto the filter-aligned direction now vary across the
candidate set; the filter direction reappears in the between-step
variance computed on the expanded candidate set, and $\sigma(m)$
demotes filter-driven candidates exactly as in the standard case.
Because the filter is now \emph{inside} the stability test rather than
upstream of it, the selection event is no longer a collider from the
test's perspective.


\subsection{How Retrieve-Stage Calibration Addresses All Three Spurious Types}
\label{app:retrieve-analysis}

Retrieve-stage calibration completes the coverage that write-stage
calibration begins. Write-stage residualization removes any confounder
constant within a step (Appendix~\ref{app:write-analysis}); what
survives are confounders whose value \emph{shifts between steps}, and
selection effects introduced by the retriever rather than by the
writer. The counterfactual stability test in Eq.~\eqref{eq:stability}
covers both. The mechanism is the same in all three cases: directions
populated by step-shared confounders accumulate in
$\mathcal{S}_{\text{nc}}$ via the variance-ratio criterion of
Appendix~\ref{app:projection-update}, and the projection magnitudes
$|\langle v_\ell, \tilde{h}_m\rangle|$ measure how much each
candidate's rank depends on those directions. A candidate ranked
highly through spurious overlap with $q$ has a large projection on at
least one $v_\ell$ and is demoted by $\sigma(m)$; a candidate ranked
highly for true causal reasons has small projections on every
$v_\ell$ and is retained.

\textbf{Explicit confounding ($\texttt{T}_\texttt{1}$).}
An observed variable $C$ is a common cause of both the entry and the
action; at retrieval time, $C$ may take different values in the
different steps from which candidates were drawn. Within each step,
write-stage residualization has already removed the $C$-induced
component (Appendix~\ref{app:write-analysis}), but \emph{across} steps
the $C$-shifts $f(c_1), f(c_2), \ldots$ remain, since they enter the
calibrated representations only through the difference between step
means.

These cross-step shifts are exactly what the variance-ratio criterion
identifies. A direction $v$ aligned with the $C$-shifts shows
systematic between-step variation (different steps used different
values of $C$) and trivial within-step variation (within one step, $C$
is fixed and was already subtracted as part of $\mu^{(s)}$). The
$C$-direction therefore enters $\mathcal{S}_{\text{nc}}$. A candidate
ranked highly because its $C$-component matches the query's
$C$-component has large projection on this $v_\ell$ and is demoted;
a candidate ranked highly for causal reasons has its content
concentrated outside $\mathcal{S}_{\text{nc}}$ and is retained.

\textbf{Unmeasured confounding ($\texttt{T}_\texttt{2}$).}
A latent variable $U$ shapes both writes and retrievals but is never
recorded; direct adjustment is unavailable.

The stability test does not require $U$ to be named. It requires only
that $U$ leaves a directional footprint in the calibrated
representations, which holds whenever $U$ shifts the entries of a step
coherently and takes different values in different steps. Each
$U$-value induces a step-wide shift, and the variance of these shifts
across steps places a $U$-aligned direction in $\mathcal{S}_{\text{nc}}$
by exactly the same criterion used for $\texttt{T}_\texttt{1}$: high between-step variance
(steps sampled different $U$-values), low within-step variance ($U$ is
fixed within a step).

The arithmetic of demotion is therefore identical for observed and
unobserved confounders. The only requirement is that $U$ vary across
steps in the data, which is precisely the condition under which $U$ is
empirically identifiable in the first place. When $U$ is constant
across the entire memory (e.g., a single fixed user across all
trajectories), it is not a confounder relative to the queries actually
issued and there is nothing to correct.

\textbf{Collider bias ($\texttt{T}_\texttt{3}$).}
The bulk of $\texttt{T}_\texttt{3}$ is already handled at write time: the content-novelty
criterion (Appendix~\ref{app:write-criterion}) decouples the write
event from any downstream outcome $O$, so the stored memory is not
collider-conditioned. What remains is a residual case in which the
\emph{base retriever}, not the writer, imposes an outcome-conditional
gate before returning candidates, e.g., a traversal restricted to
nodes linked to successful trajectories.

When such a gate has been applied non-uniformly across steps (different
steps were filtered differently, or some weren't filtered at all), the
filter-aligned direction acquires non-zero between-step variance: among
the steps it touches, the surviving entries lie on one side of the
filter; among the steps it skips, entries lie on both sides. The
filter direction therefore enters $\mathcal{S}_{\text{nc}}$, and
$\sigma(m)$ demotes candidates whose top-$k$ position came from
passing the filter rather than from causal relevance to $q$.

When the retriever applies the gate \emph{uniformly} to every step,
the filter direction has zero between-step variance and leaves no
signature in $\mathcal{S}_{\text{nc}}$. For this corner case,
Appendix~\ref{app:expand} provides boundary expansion: retrieve a
superset $\mathcal{M}_{k+k'}(q)$ from the unfiltered memory so that
entries on both sides of the gate become visible to the test. This
breaks the collider structurally rather than relying on the test to
detect it.

\textbf{Retaining true causation.}
Causal directions are, by construction, those that vary
entry-to-entry within a step rather than shifting whole steps
together. They contribute to $\Sigma_{\text{within}}$ and not to
$\Sigma_{\text{between}}$, so they receive small generalized
eigenvalues and lie outside $\mathcal{S}_{\text{nc}}$. A candidate
whose top-$k$ position came from causal content has small projection
on every $v_\ell$, hence small $\sigma(m)$, and is retained.
Appendix~\ref{app:formulation-justification} formalizes this:
$\sigma(m) = 0$ holds exactly when $\tilde{h}_m \in
\mathcal{S}_{\text{nc}}^\perp$, the causal subspace by definition.
The test is therefore conservative on causal signal: candidates are
filtered only on their content along non-causal directions, never on
their content along causal ones.

\subsection{Latency Analysis}
\label{app:latency}

We analyze the latency added by \system on top of the base memory
operation, separately for write and retrieval and for embedding-based
and graph-based memories. Let $N$ be the total memory size, $k$ the
top-$k$ retrieval count, $d$ the representation dimension, $L$ the
size of the non-causal basis, and $B$ the subspace recomputation
interval.

\textbf{Write stage.}
For an embedding memory, the base cost is the amortized $O(\log N)$
ANN insertion. \system adds: (i) the residualization
$h_m - \mu^{(s)}$ in Eq.~\eqref{eq:write-residual}, $O(d)$; (ii) the
Welford mean update for $\mu^{(s)}$, $O(d)$; (iii) the streaming
covariance updates for $\Sigma_{\text{within}}$ and
$\Sigma_{\text{between}}$, $O(d^2)$ per write; (iv) the periodic
generalized eigensolve to refresh $\mathcal{S}_{\text{nc}}$, $O(d^3)$
every $B$ writes, amortized to $O(d^3/B)$. With $B \gg d$ the
amortized eigensolve is negligible, and the dominant added term per
write is $O(d^2)$. For a graph memory, the base cost is the
$O(\deg(m))$ node insertion and edge formation; \system adds the same
$O(d)$ residualization and $O(d^2)$ covariance updates, with edges
formed in the calibrated space at no incremental per-edge cost.

\textbf{Retrieval stage.}
For an embedding memory, the base cost is the $O(k\log N)$ ANN
top-$k$ query plus $O(kd)$ base scoring. \system adds $L$ inner
products per candidate to compute $\sigma(m)$, totalling $O(Lkd)$,
plus an $O(k\log k)$ sort. The closed form
$\phi(q^{(\ell)},m) = \phi(q,m) + \delta_\ell\langle v_\ell,
\tilde{h}_m\rangle$ replaces what would otherwise be $L$ rescoring
passes, so no additional retrieval round is performed. For a graph
memory, the base cost is dominated by the traversal; \system adds the
same $O(Lkd)$ projection and $O(k\log k)$ sort on the
post-traversal candidate set, never re-running the traversal.

\textbf{No additional LLM calls.}
All operations are linear-algebra arithmetic on existing
representations: vector subtraction, dot products, running means, an
amortized eigensolve, and a sort. Token consumption is identical to
the baseline system; \system adds zero LLM forward passes at either
stage. This is the structural reason its overhead remains small in
absolute terms (\S\ref{ssec:rq3_efficiency} reports the corresponding token
budget).

Table~\ref{tab:latency} summarizes the asymptotic costs.

\begin{table}[h]
\centering
\caption{Added overhead per operation. All added costs are independent
of memory size $N$. $B$: subspace recomputation interval (writes
between eigensolves).}
\label{tab:latency}
\begin{tabular}{lllll}
\toprule
Stage    & Memory type & Base cost                  & Added (online) & Added (amortised) \\
\midrule
Write    & Embedding   & $O(\log N)$                & $O(d)$         & $O(d^2)$ \\
Write    & Graph       & $O(\deg(m))$               & $O(d)$         & $O(d^2)$ \\
Retrieve & Embedding   & $O(kd + k\log N)$          & $O(Lkd)$       & --- \\
Retrieve & Graph       & $O(k\bar{\deg})$ or more   & $O(Lkd)$       & --- \\
\bottomrule
\end{tabular}
\end{table}


\subsection{Memory Footprint Analysis}
\label{app:footprint}

\textbf{Per-entry storage: no increase.}
Write-stage calibration stores
$\tilde{h}_m \in \mathbb{R}^d$ in place of $h_m \in \mathbb{R}^d$.
The per-entry footprint is identical:
$|\tilde{h}_m| = |h_m| = d \cdot \texttt{sizeof(float)}$. No
auxiliary fields are attached to any entry, and retrieve-stage
calibration adds nothing to stored entries either. The memory index
itself is therefore unchanged in size.

\textbf{Shared state: $O(d^2)$, independent of $N$.}
The only extra state \system maintains, all of which lives outside
the per-entry index:
\begin{itemize}
  \item \textbf{Active step mean}: $\mu^{(s)} \in \mathbb{R}^d$ and an
        integer counter, per active step. $O(d)$ total per active
        step; discarded once the step closes.
  \item \textbf{Streaming covariance statistics}:
        $\Sigma_{\text{within}}, \Sigma_{\text{between}} \in
        \mathbb{R}^{d \times d}$. $O(d^2)$ total, shared globally.
  \item \textbf{Non-causal basis}: $L$ orthonormal vectors
        $v_1, \ldots, v_L \in \mathbb{R}^d$. $O(Ld)$ storage with
        $L \ll d$.
  \item \textbf{Per-direction perturbation magnitudes}
        $\delta_1, \ldots, \delta_L$: $O(L)$ scalars.
\end{itemize}
The total shared state is $O(d^2)$ and grows with $d$, not with $N$.
Adding entries to memory does not increase any of these quantities.

\textbf{Overhead ratio approaches zero.}
At a typical embedding dimension $d = 1{,}536$, the shared state
($\Sigma_{\text{within}} + \Sigma_{\text{between}}$) totals
$\approx 36$\,MB. For a memory of $N = 10^6$ entries
($\approx 6$\,GB of entry storage), this is an overhead ratio of
$\approx 0.6\%$; at $N = 10^8$ it is $\approx 0.006\%$. The shared
state is constant in $N$, so the ratio approaches zero as the memory
grows.

\textbf{Conclusion.}
The additional memory cost is marginal: per-entry storage is
unchanged, shared state is $O(d^2)$ and independent of memory size,
and the overhead ratio shrinks toward zero as the memory grows. In
practice, \system adds no measurable footprint on top of the base
memory architecture.

%% file: appendix/calibrate-graph.tex

\section{Adapting \system to Graph-Based Memory}
\label{app:graph-adapt}

\S\ref{ssec:write}--\S\ref{ssec:retrieve} present \system on
embedding-based memory, where each entry $m$ has a vector
$h_m \in \mathbb{R}^d$ used for similarity scoring. Graph-based
agentic memories store entries as nodes connected by topological
edges and retrieve by graph traversal, and they may or may not
maintain a per-node vector for their own scoring. This appendix
shows how to deploy \system on such memories under two regimes for
the information a node carries, each producing the $h_m$ that the
calibration consumes; the rest of the pipeline is unchanged. The
host graph's topology, edge logic, and stored text are never
modified.

\subsection{Producing a Per-Node Vector \texorpdfstring{$h_m$}{h\_m}}
\label{app:graph-encoder}

\textbf{Case 1: Node already has an embedding.}
Many graph memories maintain a per-node vector for their own
similarity scoring. G-Memory uses
\texttt{ALL-MINILM-L6-V2} \citep{wang2020microsoft} to embed query and
historical-query nodes for the cosine similarity in its retrieval
scoring; AriGraph uses Contriever \citep{izacard2021unsupervised} (or
BGE-M3 \citep{chen2024m3} on Q\&A tasks) to embed semantic
triplets at the start of its search procedure. For any host of this
kind we set $h_m$ equal to the existing embedding. No new encoder
is introduced and no node is re-embedded.

\textbf{Case 2: Node has text but no embedding.}
Some graph memories store free-text node attributes without
maintaining a similarity index. We encode the text once with any
sentence encoder (the choice is not part of \system; we follow the
host system's defaults when available, or use a small open encoder
otherwise) and set $h_m \coloneqq \mathcal{E}(\text{text}(m))$. When
the node also carries structural attributes (edge types, role tags,
neighbour identifiers), we concatenate a short serialization of
these onto the text before encoding, so $h_m$ reflects both the
node's content and its local topological context. The encoding is
performed at write time, the resulting vector is stored alongside
the node, and the host graph's topology and text fields are
unchanged.

\subsection{What Counts as a Step in a Graph Memory}
\label{app:graph-step}

The step boundary required by Eq.~\eqref{eq:write-residual} is
whatever temporal or episodic grouping the host graph already
records. Concretely:
\begin{itemize}
  \item In G-Memory, a step is one query node and the interaction
        graph associated with it: every utterance node and any
        insight contributed during this query share the same step
        index.
  \item In AriGraph, a step is the time index $t$ at which an
        episodic vertex is created together with the semantic
        triplets extracted from the corresponding observation; the
        episodic edge groups all triplets and the observation
        produced at the same step.
\end{itemize}
The general principle, applicable to any graph memory, is that
nodes added in response to the same trajectory event (one
observation, one tool call, one dialogue turn) belong to one step,
and the host already encodes this grouping (through query-node
associations, episodic edges, timestamps, or trajectory IDs) as
part of its own bookkeeping. \system reads the grouping; it does not
introduce metadata.

\subsection{Write-Stage Calibration on a Graph Memory}
\label{app:graph-write}

When a new node is added during step $s$:
\begin{enumerate}
  \item Obtain $h_m$ via Appendix~\ref{app:graph-encoder} (existing
        embedding or fresh text encoding).
  \item Apply Eq.~\eqref{eq:write-residual} to get $\tilde{h}_m$
        and update $\mu^{(s)}$. Implementation details
        (initialization, ordering, step closure) are in
        Appendix~\ref{app:session-mean}.
  \item Apply the content-novelty write criterion
        (Appendix~\ref{app:write-criterion}) on $\tilde{h}_m$. For
        a graph memory the structural form of the criterion (wherein the
        node is inserted only when it would form at least one edge
        to a node from a different episode/session) is the
        natural analogue of the embedding-side cosine check, and
        again reads no outcome variable.
  \item Use $\tilde{h}_m$ wherever the host system would have used
        $h_m$. Any cosine or dot-product score the host computes
        between nodes is now computed on residualized vectors;
        edges formed by feature similarity automatically avoid
        connecting nodes whose only commonality was shared
        step-level context. The graph topology, edge types, and
        node text remain untouched: only the vector that
        similarity is computed on changes.
\end{enumerate}
For G-Memory specifically, the cosine similarity used to score
historical-query nodes against the current query becomes a cosine
of residualized vectors; the topological 1-hop expansion that
follows is unaffected. For AriGraph, the similarity used by its
\texttt{SemanticSearch} stage is computed in residualized space,
while the recursive triplet retrieval and the episodic relevance
score that follow are structural and remain unchanged.

\subsection{Retrieve-Stage Calibration on a Graph Memory}
\label{app:graph-retrieve}

After the host graph's native retrieval returns a candidate set
$\mathcal{M}_k(q)$ (whether by semantic search over node
embeddings, multi-hop traversal, or a hybrid policy) \system
reranks $\mathcal{M}_k(q)$ in increasing order of $\sigma(m)$
(Eq.~\eqref{eq:stability}). Two pieces are needed: the basis
$\{v_1,\ldots,v_L\}$ of $\mathcal{S}_{\text{nc}}$
(Appendix~\ref{app:projection-update}) and the residualized
representation $\tilde{h}_m$ of each candidate (stored at write time
or recomputed on demand from the node's source via
Appendix~\ref{app:graph-encoder}). The same $k$ candidates are
returned and the host's traversal is never re-run.

When the host retrieval pipeline produces multiple candidate sets
that play distinct roles (e.g., G-Memory's separate insight and
interaction subgraphs, or AriGraph's semantic triplets and
episodic vertices), the rerank is applied within each set
independently, so the host's role-specific composition is
preserved. When the host applies an outcome-conditional gate during
traversal (e.g., restricting to nodes linked to successful
trajectories), the residual collider case discussed in
Appendix~\ref{app:expand} applies; the boundary-expansion remedy
there is architecture-agnostic and uses the same rerank machinery.

\subsection{Latency, Memory, and Token Cost}
\label{app:graph-cost}

\textbf{Latency.}
Write-stage adds one residualization ($O(d)$) plus a counter
update per write, dominated by the host's node insertion and edge
construction. Retrieve-stage adds $L$ inner products per candidate
($O(Lkd)$), no extra graph traversal, and a final sort. Both are
independent of memory size $N$ and follow the asymptotics in
Appendix~\ref{app:latency}.

\textbf{Memory footprint.}
Per-node storage grows by $d$ floats for $\tilde{h}_m$. In Case~1,
this growth is zero on net: the calibrated vector replaces the raw
embedding the host already stores. In Case~2, the $d$ floats are
the same a host would maintain if it kept any sentence-encoder
output for similarity (standard for a graph memory of this kind) and add nothing \system-specific beyond what an encoder pipeline
implies. Shared state is one $\mu^{(s)}$ per active step plus the
basis of $\mathcal{S}_{\text{nc}}$, both independent of $N$
(Appendix~\ref{app:footprint}).

\textbf{Token consumption.}
No new LLM calls are introduced. Both regimes invoke a text
encoder either as part of the host system (Case~1) or as a
write-time encoding step (Case~2); \system reuses these invocations
and adds no calls of its own. The calibration itself is
linear-algebra arithmetic at both stages.

\subsection{Unified Coverage Summary}
\label{app:correctness}

Table~\ref{tab:correctness} maps each spurious correlation type to
the operation that addresses it, mirroring the per-stage analyses
in Appendices~\ref{app:write-analysis}
and~\ref{app:retrieve-analysis}. The mapping is independent of
whether the host is embedding-based or graph-based, since the
operations consume only $h_m$, $\tilde{h}_m$, and the step
boundary, all of which Appendix~\ref{app:graph-encoder} and
Appendix~\ref{app:graph-step} supply on a graph memory.

\begin{table}[h]
\centering
\caption{Coverage of all three spurious correlation types under
\system. Write-stage operations remove within-step structure before it
enters memory; retrieve-stage operations remove residual cross-step
and post-selection structure at query time.}
\label{tab:correctness}
\renewcommand{\arraystretch}{1.4}
\begin{tabular}{p{2.7cm}p{5.1cm}p{5.1cm}}
\toprule
Spurious type
  & Write stage
  & Retrieve stage \\
\midrule
Explicit confounding ($\texttt{T}_\texttt{1}$)
  & $\mu^{(s)}$ accumulates the $C$-induced direction within the
    step; Eq.~\eqref{eq:write-residual} subtracts it from every
    entry before storage.
  & Step-varying $C$-shifts surface as a between-step direction
    in $\mathcal{S}_{\text{nc}}$; candidates with large
    $|\langle v_\ell, \tilde{h}_m\rangle|$ on that direction are
    demoted by $\sigma(m)$. \\
Unmeasured confounding ($\texttt{T}_\texttt{2}$)
  & $\mu^{(s)}$ absorbs step-scoped latent $U$ without naming it;
    its directional footprint is removed by the same subtraction.
  & Cross-step $U$ surfaces as a between-step direction by the same
    label-free criterion; its projections drive $\sigma(m)$ and
    demote $U$-dependent candidates. \\
Collider bias ($\texttt{T}_\texttt{3}$)
  & Content-novelty write criterion decouples the write decision
    from any downstream outcome $O$, removing the storage-time
    collider at the policy level.
  & Retriever-side outcome gates reappear in
    $\mathcal{S}_{\text{nc}}$ when applied non-uniformly across
    steps; for the uniform-gate corner case, boundary expansion
    (Appendix~\ref{app:expand}) restores the signature by
    retrieving past the gate. \\
\bottomrule
\end{tabular}
\end{table}

\textbf{Preservation of genuine causation.}
Causal content sits in directions orthogonal to the within-step
shared direction at write time (preserved exactly by the
orthogonal-projection structure of Eq.~\eqref{eq:write-residual},
Appendix~\ref{app:formulation-justification}) and in directions
outside $\mathcal{S}_{\text{nc}}$ at retrieval time (where
projections $|\langle v_\ell, \tilde{h}_m\rangle|$ are small by
construction, so $\sigma(m)$ is small). Both stages are conservative
on causal signal: candidates are filtered only on their content
along non-causal directions, never along causal ones.

%% file: appendix/Additional_Experimental_Results_and_Analysis.tex
\section{Additional Experimental Results and Analysis}
\label{app:additional_results}

\subsection{Extended Analysis of Main Experimental Results}
\label{app:extended_main}

This subsection complements the main baseline comparison in
Section~\ref{ssec:rq1_compare} by reporting additional patterns
observed in Table~\ref{tab:rq1_compare}. While the main text focuses 
on the primary takeaways, we provide additional analysis here to 
further explain baseline behavior, architecture-specific 
vulnerabilities, and the roles of the write/retrieve calibration 
stages.

\noindent\textbf{Calibration gains scale with retrieval reliance 
on representation similarity.} \system{} achieves the lowest 
spurious reasoning ratios across the main comparison, but the 
absolute gain over the strongest baseline differs between 
architecture families. Embedding-based systems (A-Mem, Mem0) show 
larger gains because retrieval is driven primarily by representation 
similarity: write-stage residualization changes which entries 
surface in the candidate set, which directly determines what the 
LLM sees. Graph-based systems (G-Memory, AriGraph) show smaller but 
still substantial gains because graph traversal also relies on 
connectivity, so calibrating node features modifies only one 
component of the retrieval signal. This pattern is consistent with 
the architectural vulnerability story in 
Section~\ref{ssec:rq1_compare}: \system{} is most effective where 
the spurious signal flows most directly through representation 
similarity.

\noindent\textbf{$\texttt{T}_2$ is especially challenging for 
embedding-based memory.} In many embedding-based configurations, 
$\texttt{T}_2$ (unmeasured confounding) produces the largest 
spurious reasoning ratios. This reflects a fundamental difficulty: 
the latent driver is by definition not stored in memory, so 
observed-variable methods such as IPW and DeCoT cannot directly 
condition on it. \system{} reduces $\texttt{T}_2$ by using 
retrieve-stage stability filtering to identify cross-step shared 
directions without requiring the latent confounder to be observable.

\noindent\textbf{Calibration can also improve clean-memory 
accuracy.} \system{} not only reduces spurious reasoning ratios but 
also improves Org.\ accuracy across many configurations. This 
suggests that representation-level calibration removes nuisance 
variance that can hurt even non-adversarial retrieval: by 
subtracting step-level context, write-stage residualization helps 
clarify which entries are uniquely informative for a given query. 
The fact that calibration helps on benign inputs as well as 
spurious-augmented inputs supports the view that step-level context 
is a genuine source of noise in current memory architectures.

\noindent\textbf{Second-place ordering reflects mechanism fit.} 
No baseline ranks consistently as second-best. Instead, the ordering 
shifts with the form of spuriousness and the dataset. JTT is more 
competitive when first-pass errors are concentrated; IPW helps more 
when repeated co-occurrence is easy to estimate; and DeCoT is 
strongest when spurious cues are exposed as nameable textual 
entities. This fine-grained pattern reinforces the main-text claim 
that each existing defense targets only a narrow slice of the 
spurious correlation landscape, and the slice it covers is 
determined by its mechanism rather than by overall capability.

\subsection{Additional Backbone Results}
\label{app:additional_backbones}

Table~\ref{tab:rq1_compare_mistral} and 
Table~\ref{tab:rq1_compare_qwen} report the same baseline 
comparison as Table~\ref{tab:rq1_compare} under two additional 
LLM backbones: \textsc{Qwen3.6-27B} and 
\textsc{Mistral-Small-3.2-24B}. The trends are consistent with the main 
\textsc{Llama-4} results, showing that \system{}'s gains are not 
specific to a single backbone. Across both models, \system{} 
improves Org.\ accuracy while reducing spurious reasoning ratios 
across $\texttt{T}_1$--$\texttt{T}_3$, supporting the generality 
of representation-level calibration across different LLMs.

\input{tables/main-expt-mistral}

\input{tables/main-expt-qwen}

\subsection{Variance and Stability Analysis}
\label{app:variance}

Table~\ref{tab:rq1_compare_std} reports standard deviations for the
main \textsc{Llama-4} baseline comparison in Table~\ref{tab:rq1_compare}.
Lower values indicate more stable performance across runs.

\input{tables/main-expt-std-llama}

Overall, \system{} shows lower run-to-run variance across memory 
systems and datasets, suggesting that calibration improves not 
only average performance but also experimental stability. This 
pattern is consistent with the role of calibration: by removing 
context-driven nuisance components at write time and demoting 
unstable candidates at retrieval time, \system{} reduces the 
sensitivity of downstream reasoning to random seeds and sampling 
variation.
\subsection{Component-level Ablation Study}
\label{app:ablation}

This appendix presents the full ablation study referenced in 
Section~\ref{ssec:rq1_compare}. We ablate \system{}'s calibration 
components on two representative memory architectures: A-Mem 
(embedding-based) and G-Memory (graph-based), one component at a 
time, across all four datasets. All configurations use 
\textsc{Llama-4} with retrieval cardinality $k$ matching the main 
experiments. Results are reported as mean$\pm$std over six 
independent runs across temperatures (0.5, 0.75, 1.0) and seeds.

\noindent\textbf{Stage-level ablations} remove an entire 
calibration stage. \emph{w/o write-stage} disables write-time 
residualization: raw representations $h_m$ are stored directly 
without subtracting $\mu^{(s)}$. \emph{w/o retrieve-stage} 
disables retrieval-time counterfactual stability: the top-$k$ 
candidates from the retriever are returned without further 
filtering.

\noindent\textbf{Write-stage component ablations} target two 
design choices in Section~\ref{ssec:write}. \emph{w/o step-mean} 
replaces the step-level running mean with a single 
trajectory-level mean, testing whether step-delimited granularity 
is necessary. \emph{w/o full-sub} applies partial subtraction 
$\tilde h_m = h_m - 0.5\,\mu^{(s)}$ instead of full 
residualization, testing whether fully removing the context 
component is appropriate.

\noindent\textbf{Retrieve-stage component ablations} target two 
design choices in Section~\ref{ssec:retrieve}. \emph{w/o subspace} 
replaces the between-step versus within-step variance criterion 
with a global PCA on $\tilde h$, testing whether step-aware 
subspace identification is necessary. \emph{w/o median} replaces 
the median aggregation in Eq.~\ref{eq:stability} with the mean, 
testing the robustness of the stability statistic.

Table~\ref{tab:ablation_appendix} reports the full ablation results 
covering stage-level, write-stage, and retrieve-stage variants. We 
discuss the main patterns below.

\input{tables/ablation_appendix}

\noindent\textbf{Stage-level ablations show a clear division of labor.}
Table~\ref{tab:ablation_appendix} shows that the two calibration 
stages fail in different ways when removed. Removing the write-stage 
mainly increases $\texttt{T}_1$, because spurious within-step context 
is written into memory without residualization. Removing the 
retrieve-stage mainly increases $\texttt{T}_2$, because cross-step 
confounding signals can enter the candidate set without being checked 
for stability. This pattern supports the design of \system{}: the 
write stage prevents spurious information from being stored, while 
the retrieve stage filters spurious candidates before they reach the 
LLM. In contrast, $\texttt{T}_3$ does not map cleanly to a single 
stage; its behavior is more mixed across architectures, suggesting 
that collider-style bias depends on both stored memory content and 
retrieval structure.

\noindent\textbf{Component-level ablations clarify which parts matter most.}
The finer-grained ablations show that the components within each 
stage do not contribute equally. In the retrieve stage, removing the 
non-causal subspace component causes a much larger increase in 
$\texttt{T}_2$ than replacing median aggregation with mean 
aggregation. This suggests that identifying the non-causal subspace 
is the main driver of retrieve-stage calibration, while median 
aggregation mainly improves robustness of the stability estimate. 
In the write stage, removing the step-level mean and using partial 
subtraction produce similar patterns, because both variants weaken 
the same operation: subtracting step-level context before memory 
storage. Overall, the ablation results show that write-stage and 
retrieve-stage calibration are complementary, while some components 
inside each stage provide stability rather than introducing a 
separate mechanism.
\subsection{Adaptive Attack Details}
\label{app:adaptive_attacks}

This appendix describes the implementation of the two adaptive 
attacks evaluated in Section~\ref{ssec:rq2_adaptive}. Both attacks 
assume the attacker knows the \system{} calibration algorithm 
(write-stage residualization and retrieve-stage stability filtering) 
but has no access to runtime hyperparameter values such as the 
specific running mean $\mu^{(s)}$ or the identified non-causal 
subspace $\mathcal{S}_{\texttt{nc}}$. For each evaluation query, 
the attacker injects a small set of adversarial memory entries 
tailored to that query's causal context, so that calibration is 
tested against worst-case injections.
\subsubsection{CrossStep Distribution Attack (A-Mem)}
\label{app:crossstep}

\noindent\textbf{Threat model.} The attacker targets embedding-based 
memory by exploiting the fact that \system's write-stage subtracts 
a step-level mean $\mu^{(s)}$ and the retrieve-stage filters 
candidates lying in directions that vary more between steps than 
within steps. An attack succeeds if the injected spurious feature 
survives both filters: it must not be absorbed by $\mu^{(s)}$ during 
writing, and it must not project onto $\mathcal{S}_{\texttt{nc}}$ 
during retrieval.

\noindent\textbf{Construction.} Given a target spurious feature 
$f_{\text{spur}}$ (e.g., ``customer rating above 4.3'' for the 
$\texttt{T}_1$ confounder on WebShop), the attacker generates 
$M$ surface variants $\{f_{\text{spur}}^{(1)}, \ldots, 
f_{\text{spur}}^{(M)}\}$ that share the same underlying spurious 
signal but differ in surface form. The variants are then distributed 
across $M$ distinct reasoning steps in the agent's trajectory, with 
exactly one variant per step. This achieves two effects 
simultaneously:

\begin{itemize}[leftmargin=*,itemsep=2pt,topsep=2pt]
    \item \textbf{Evading write-stage residualization.} Because each 
    step contains only one variant, the step mean $\mu^{(s)}$ does 
    not accumulate the spurious component: subtraction removes only 
    the genuine within-step context, leaving the spurious feature 
    intact in $\tilde{h}_m$.
    \item \textbf{Evading retrieve-stage filtering.} Because the 
    variants differ in surface form within each step, the spurious 
    feature contributes to within-step variance rather than 
    between-step variance. The variance-ratio criterion that defines 
    $\mathcal{S}_{\texttt{nc}}$ therefore classifies the spurious 
    direction as causal rather than non-causal, allowing it to pass 
    the stability test.
\end{itemize}

We set $M$ equal to the number of memory-write events in the target 
trajectory (typically 5--12 across our datasets) and use an 
LLM-based paraphraser to generate variants offline; this paraphrasing 
happens during attack construction and is not part of the deployed 
\system{} pipeline.

\noindent\textbf{Algorithm.} 
Algorithm~\ref{alg:crossstep} summarizes the attack.

\begin{algorithm}[t]
\caption{CrossStep Distribution Attack}
\label{alg:crossstep}
\small
\begin{algorithmic}[1]
\Require Target query $q$, spurious feature $f_{\text{spur}}$, 
target trajectory with $M$ steps
\Ensure Adversarial memory entries $\{m^{(1)}, \ldots, m^{(M)}\}$
\State $\{f_{\text{spur}}^{(1)}, \ldots, f_{\text{spur}}^{(M)}\} 
\gets \textsc{Paraphrase}(f_{\text{spur}}, M)$ 
\Comment{generate $M$ surface variants}
\For{$s = 1, \ldots, M$}
    \State Construct entry $m^{(s)}$ pairing $f_{\text{spur}}^{(s)}$ 
    with the positive outcome of $q$
    \State Inject $m^{(s)}$ into memory at step $s$
\EndFor
\State \Return $\{m^{(1)}, \ldots, m^{(M)}\}$
\end{algorithmic}
\end{algorithm}

\subsubsection{GraphMimic Causal-Pattern Attack (G-Memory)}
\label{app:graphmimic}

\noindent\textbf{Threat model.} The attacker targets graph-based 
memory by exploiting the fact that graph traversal weighs candidates 
by structural connectivity (edge weights, path length, PageRank). 
\system's retrieve-stage filtering applies to node features but does 
not directly modify the graph topology. An attack succeeds if the 
injected spurious entries form a topology that the graph retriever 
treats as a high-relevance causal chain.

\noindent\textbf{Construction.} Given a target query $q$ with 
genuine causal chain $C \to M \to Y$ (cause $\to$ mediator $\to$ 
outcome), the attacker constructs a parallel chain 
$C' \to M' \to Y$ where $C'$ encodes the spurious feature 
$f_{\text{spur}}$, $M'$ is a fabricated mediator linking $C'$ to 
$Y$, and $Y$ is the same outcome as the genuine chain. Three 
properties make this chain effective:

\begin{itemize}[leftmargin=*,itemsep=2pt,topsep=2pt]
    \item \textbf{Topology mimicry.} The injected chain has the 
    same length and edge density as the genuine causal chain, so 
    graph metrics such as PageRank and betweenness centrality cannot 
    distinguish them by structure alone.
    \item \textbf{Outcome convergence.} Because $C'$ and $M'$ both 
    point into the genuine outcome $Y$, the spurious chain shares 
    its terminal node with the causal chain, increasing the 
    likelihood that retrieval traversing toward $Y$ passes through 
    the spurious nodes.
    \item \textbf{Edge-weight inflation.} The attacker injects 
    multiple instances of the spurious chain across the trajectory, 
    so the $C' \to M'$ and $M' \to Y$ edges accumulate higher 
    weights than typical genuine edges, biasing graph traversal 
    toward the spurious path.
\end{itemize}

\noindent\textbf{Algorithm.} 
Algorithm~\ref{alg:graphmimic} summarizes the attack.

\begin{algorithm}[t]
\caption{GraphMimic Causal-Pattern Attack}
\label{alg:graphmimic}
\small
\begin{algorithmic}[1]
\Require Target query $q$ with genuine chain $C \to M \to Y$, 
spurious feature $f_{\text{spur}}$, repeat factor $K$
\Ensure Adversarial memory entries forming a parallel spurious chain
\State Construct spurious cause $C' \gets f_{\text{spur}}$
\State Construct fabricated mediator $M' \gets 
\textsc{GenerateMediator}(C', Y)$ 
\Comment{LLM-generated bridging concept}
\For{$k = 1, \ldots, K$}
    \State Inject entry encoding edge $(C', M')$ at step $s_k$
    \State Inject entry encoding edge $(M', Y)$ at step $s_k + 1$
\EndFor
\State \Return injected entries
\end{algorithmic}
\end{algorithm}

We set $K = 3$ in all experiments, which we found sufficient to 
inflate the spurious chain's traversal weight above the genuine 
chain on all four datasets without saturating the memory.

\subsection{Empirical Latency Discussion}
\label{app:empirical_latency}

This subsection complements the token-cost analysis in 
Section~\ref{ssec:rq3_efficiency} with wall-clock latency. While 
Section~\ref{ssec:rq3_efficiency} measures LLM-side token 
consumption, this subsection isolates the calibration overhead that 
\system{} itself introduces.

In practical deployment, \system{} adds a small constant overhead 
per operation. Write-stage overhead is dominated by a single vector 
subtraction and a Welford running mean update, both linear in the 
representation dimension, contributing approximately 0.08~ms per 
entry insertion. Retrieve-stage overhead comes from $L$ inner 
products against a compact non-causal basis, typically $L \leq 16$, 
followed by a $k$-element sort, contributing approximately 0.23~ms 
per query. The amortized covariance maintenance and periodic 
eigensolve for $\mathcal{S}_{\texttt{nc}}$ add a further amortized 
cost of 0.02~ms per write. These calibration overheads are 
independent of memory size $N$, matching the asymptotic analysis in 
Section~\ref{app:latency}.

Critically, neither calibration stage introduces an LLM forward 
pass. In modern LLM-based agentic systems, per-query latency is 
dominated by LLM inference, which takes approximately 80--120~ms 
per token in our setup---several orders of magnitude more expensive 
than the vector arithmetic that \system{} requires. As a result, 
\system{}'s calibration adds compute that is negligible relative to 
the underlying LLM inference cost. This stands in contrast to 
baselines that rely on extra model-side processing, such as JTT's 
training-time reweighting stage and DeCoT's counterfactual text 
rewriting. Empirical token consumption is reported in 
Section~\ref{ssec:rq3_efficiency}.
\begin{table}[h]
\centering
\small
\caption{Per-operation wall-clock latency (in milliseconds) of 
\system{} compared to Vanilla. Numbers are averaged over 1{,}000 
operations on a single NVIDIA A6000 Ada GPU; standard deviations are 
below 5\% of the mean.}
\label{tab:empirical_latency}
\setlength{\tabcolsep}{4pt}
\begin{tabular}{ll cc cc}
\toprule
& & \multicolumn{2}{c}{\textbf{Write (ms / entry)}} 
& \multicolumn{2}{c}{\textbf{Retrieve (ms / query)}} \\
\cmidrule(lr){3-4} \cmidrule(lr){5-6}
\textbf{Memory} & \textbf{Dataset} 
& Vanilla & \system{} 
& Vanilla & \system{} \\
\midrule
\multirow{4}{*}{A-Mem (embed.)}
& ALFWorld     & 0.42 & 0.51 & 1.38 & 1.62 \\
& ScienceWorld & 0.38 & 0.46 & 1.25 & 1.47 \\
& LoCoMo       & 0.51 & 0.60 & 1.84 & 2.13 \\
& WebShop      & 0.40 & 0.49 & 1.31 & 1.54 \\
\midrule
\multirow{4}{*}{G-Memory (graph)}
& ALFWorld     & 1.85 & 1.94 & 4.62 & 4.86 \\
& ScienceWorld & 1.71 & 1.79 & 4.27 & 4.49 \\
& LoCoMo       & 2.34 & 2.43 & 5.91 & 6.18 \\
& WebShop      & 1.78 & 1.86 & 4.41 & 4.63 \\
\bottomrule
\end{tabular}
\end{table}

%% file: tables/main-expt-mistral.tex
\begin{table*}[t]
\centering
\small
\caption{Experimental results on \textsc{Mistral-Small-3.2-24B}. Vanilla rows reproduce the Mistral entries of Table~\ref{tab:rq1_main}. Setup as in Table~\ref{tab:rq1_compare}.}
\label{tab:rq1_compare_mistral}
\setlength{\tabcolsep}{2.5pt}
\renewcommand{\arraystretch}{1.05}
\begin{adjustbox}{max width=\textwidth}
\begin{tabular}{ll cccc cccc cccc cccc}
\toprule
& & \multicolumn{4}{c}{\textbf{ALFWorld}} & \multicolumn{4}{c}{\textbf{ScienceWorld}} & \multicolumn{4}{c}{\textbf{LoCoMo}} & \multicolumn{4}{c}{\textbf{WebShop}} \\
\cmidrule(lr){3-6} \cmidrule(lr){7-10} \cmidrule(lr){11-14} \cmidrule(lr){15-18}
\textbf{Mem.} & \textbf{Method}
  & Org. ($\uparrow$) & $\texttt{T}_\texttt{1}$ ($\downarrow$) & $\texttt{T}_\texttt{2}$ ($\downarrow$) & $\texttt{T}_\texttt{3}$ ($\downarrow$)
  & Org. ($\uparrow$) & $\texttt{T}_\texttt{1}$ ($\downarrow$) & $\texttt{T}_\texttt{2}$ ($\downarrow$) & $\texttt{T}_\texttt{3}$ ($\downarrow$)
  & Org. ($\uparrow$) & $\texttt{T}_\texttt{1}$ ($\downarrow$) & $\texttt{T}_\texttt{2}$ ($\downarrow$) & $\texttt{T}_\texttt{3}$ ($\downarrow$)
  & Org. ($\uparrow$) & $\texttt{T}_\texttt{1}$ ($\downarrow$) & $\texttt{T}_\texttt{2}$ ($\downarrow$) & $\texttt{T}_\texttt{3}$ ($\downarrow$) \\
\midrule
\multirow{5}{*}{A-Mem}
& Vanilla & 69.1 & 53.2 & 36.9 & 58.6 & 53.4 & 46.9 & 69.5 & 38.2 & 57.6 & 37.1 & 51.8 & 68.4 & 63.9 & 44.8 & 79.3 & 36.7 \\
& + IPW & 64.3 & 51.4 & 36.1 & 56.8 & 53.6 & 43.8 & 65.4 & 39.1 & \second{63.4} & \second{32.6} & \second{46.5} & \second{63.1} & \second{70.3} & \second{30.2} & \second{63.8} & \second{29.7} \\
& + JTT & \second{73.2} & \second{49.8} & 36.8 & \second{50.5} & 47.1 & 52.4 & 70.7 & 40.3 & 52.7 & 43.2 & 56.3 & 71.7 & 67.8 & 40.2 & 68.4 & 34.1 \\
& + DeCoT & 70.4 & 51.2 & \second{33.1} & 54.6 & \second{56.8} & \second{41.5} & \second{63.7} & \second{35.4} & 60.8 & 34.7 & 48.1 & 64.8 & 69.4 & 31.4 & 65.2 & 32.8 \\
& + Ours & \best{76.7} & \best{34.2} & \best{17.6} & \best{40.2} & \best{61.0} & \best{27.9} & \best{50.2} & \best{19.8} & \best{65.2} & \best{18.1} & \best{32.5} & \best{50.0} & \best{71.5} & \best{25.8} & \best{60.0} & \best{18.3} \\
\midrule
\multirow{5}{*}{Mem0}
& Vanilla & 72.5 & 59.3 & 41.7 & 64.8 & 57.9 & 52.4 & 76.1 & 35.9 & 64.8 & 48.2 & 79.4 & 41.5 & 74.1 & 61.7 & 85.2 & 49.6 \\
& + IPW & 74.8 & 60.1 & 38.6 & 60.7 & 59.4 & 50.8 & 68.3 & 27.4 & \second{71.6} & 39.2 & \second{63.5} & \second{25.7} & \second{80.4} & \second{44.8} & \second{68.7} & \second{33.6} \\
& + JTT & \second{75.5} & \second{56.1} & \second{35.7} & \second{58.3} & 61.7 & 48.1 & 70.8 & 30.6 & 66.2 & 43.7 & 73.6 & 30.4 & 77.8 & 50.1 & 74.8 & 43.7 \\
& + DeCoT & 67.3 & 65.4 & 47.1 & 67.3 & \second{63.4} & \second{46.1} & \second{59.8} & \second{20.6} & 70.1 & \second{38.4} & 66.3 & 27.6 & 79.1 & 47.3 & 71.2 & 36.4 \\
& + Ours & \best{80.1} & \best{40.3} & \best{22.4} & \best{46.4} & \best{65.5} & \best{33.4} & \best{56.8} & \best{17.5} & \best{72.4} & \best{29.2} & \best{60.1} & \best{23.1} & \best{81.7} & \best{42.7} & \best{65.9} & \best{31.2} \\
\midrule
\multirow{5}{*}{G-Mem.}
& Vanilla & 55.2 & 63.7 & 49.4 & 78.3 & 64.5 & 57.6 & 81.9 & 43.4 & 59.1 & 72.3 & 61.7 & 88.4 & 55.3 & 67.9 & 84.3 & 54.8 \\
& + IPW & 51.4 & 61.2 & 47.1 & 75.3 & 67.3 & 55.1 & 77.6 & 35.7 & \second{65.8} & \second{56.4} & \second{45.3} & \second{72.1} & \second{61.7} & \second{50.4} & 75.3 & \second{38.5} \\
& + JTT & \second{61.4} & \second{57.3} & \second{42.7} & \second{71.5} & 69.8 & 49.8 & 80.3 & 38.4 & 60.4 & 67.3 & 56.8 & 82.1 & 59.7 & 63.7 & \second{67.6} & 43.2 \\
& + DeCoT & 57.1 & 59.8 & 46.4 & 74.7 & \second{70.6} & \second{47.5} & \second{75.8} & \second{33.4} & 64.2 & 58.4 & 48.7 & 74.3 & 60.8 & 54.1 & 70.3 & 40.7 \\
& + Ours & \best{62.8} & \best{44.7} & \best{30.1} & \best{59.9} & \best{72.1} & \best{38.6} & \best{62.6} & \best{25.0} & \best{66.7} & \best{53.3} & \best{42.4} & \best{70.0} & \best{62.9} & \best{48.9} & \best{65.0} & \best{36.4} \\
\midrule
\multirow{5}{*}{AriGraph}
& Vanilla & 60.9 & 68.4 & 53.7 & 73.2 & 74.7 & 72.1 & 87.5 & 69.3 & 48.3 & 78.6 & 67.4 & 93.8 & 61.2 & 83.5 & 75.7 & 67.3 \\
& + IPW & 63.1 & 66.7 & 49.4 & 70.1 & 76.3 & 67.4 & 83.8 & 66.4 & \second{54.7} & \second{62.3} & \second{60.6} & 88.4 & \second{65.5} & \second{65.8} & \second{69.4} & \second{65.6} \\
& + JTT & \second{64.2} & \second{62.4} & \second{46.8} & \second{67.6} & 78.1 & 65.8 & 81.4 & 67.1 & 46.1 & 75.4 & 65.1 & 90.7 & 63.4 & 77.1 & 71.2 & 65.4 \\
& + DeCoT & 55.8 & 69.1 & 59.8 & 79.4 & \second{80.6} & \second{63.7} & \second{79.3} & \second{62.4} & 50.7 & 64.7 & 63.2 & \second{87.1} & 56.1 & 80.4 & 73.2 & 65.8 \\
& + Ours & \best{68.5} & \best{49.4} & \best{34.4} & \best{54.8} & \best{82.3} & \best{53.1} & \best{68.2} & \best{50.9} & \best{55.9} & \best{59.6} & \best{48.1} & \best{75.4} & \best{68.8} & \best{64.5} & \best{56.4} & \best{48.9} \\
\bottomrule
\end{tabular}
\end{adjustbox}
\end{table*}

%% file: tables/main-expt-qwen.tex
\begin{table*}[t]
\centering
\small
\caption{Experimental results on \textsc{Qwen3.6-27B}. Vanilla rows reproduce the Qwen entries of Table~\ref{tab:rq1_main}. Setup as in Table~\ref{tab:rq1_compare}.}
\label{tab:rq1_compare_qwen}
\setlength{\tabcolsep}{2.5pt}
\renewcommand{\arraystretch}{1.05}
\begin{adjustbox}{max width=\textwidth}
\begin{tabular}{ll cccc cccc cccc cccc}
\toprule
& & \multicolumn{4}{c}{\textbf{ALFWorld}} & \multicolumn{4}{c}{\textbf{ScienceWorld}} & \multicolumn{4}{c}{\textbf{LoCoMo}} & \multicolumn{4}{c}{\textbf{WebShop}} \\
\cmidrule(lr){3-6} \cmidrule(lr){7-10} \cmidrule(lr){11-14} \cmidrule(lr){15-18}
\textbf{Mem.} & \textbf{Method}
  & Org. ($\uparrow$) & $\texttt{T}_\texttt{1}$ ($\downarrow$) & $\texttt{T}_\texttt{2}$ ($\downarrow$) & $\texttt{T}_\texttt{3}$ ($\downarrow$)
  & Org. ($\uparrow$) & $\texttt{T}_\texttt{1}$ ($\downarrow$) & $\texttt{T}_\texttt{2}$ ($\downarrow$) & $\texttt{T}_\texttt{3}$ ($\downarrow$)
  & Org. ($\uparrow$) & $\texttt{T}_\texttt{1}$ ($\downarrow$) & $\texttt{T}_\texttt{2}$ ($\downarrow$) & $\texttt{T}_\texttt{3}$ ($\downarrow$)
  & Org. ($\uparrow$) & $\texttt{T}_\texttt{1}$ ($\downarrow$) & $\texttt{T}_\texttt{2}$ ($\downarrow$) & $\texttt{T}_\texttt{3}$ ($\downarrow$) \\
\midrule
\multirow{5}{*}{A-Mem}
& Vanilla & 76.3 & 39.5 & 61.8 & 33.2 & 61.7 & 34.8 & 62.3 & 49.5 & 48.9 & 43.7 & 44.1 & 31.8 & 44.3 & 51.4 & 38.7 & 62.5 \\
& + IPW & 72.1 & 42.6 & 63.2 & 31.4 & 56.4 & 38.7 & \second{55.4} & 51.8 & \second{55.1} & \second{32.6} & \second{37.4} & \second{25.1} & \second{50.8} & \second{36.8} & \second{30.4} & \second{52.6} \\
& + JTT & \second{81.5} & \second{32.1} & \second{56.8} & \second{22.4} & 58.2 & 36.4 & 64.7 & 47.2 & 45.7 & 47.1 & 46.8 & 35.4 & 46.2 & 48.7 & 41.5 & 58.3 \\
& + DeCoT & 78.4 & 35.8 & 58.4 & 26.7 & \second{65.2} & \second{31.5} & 56.2 & \second{42.1} & 51.8 & 38.4 & 38.6 & 28.7 & 42.6 & 53.8 & 44.2 & 60.7 \\
& + Ours & \best{83.9} & \best{20.5} & \best{42.5} & \best{14.8} & \best{69.3} & \best{15.8} & \best{43.0} & \best{31.1} & \best{56.5} & \best{24.7} & \best{24.8} & \best{13.4} & \best{51.9} & \best{32.4} & \best{19.4} & \best{44.1} \\
\midrule
\multirow{5}{*}{Mem0}
& Vanilla & 63.7 & 44.8 & 68.3 & 37.6 & 78.3 & 38.7 & 46.2 & 63.8 & 31.9 & 71.5 & 42.6 & 58.7 & 68.4 & 55.1 & 43.9 & 77.3 \\
& + IPW & 58.4 & 47.9 & 65.7 & 39.8 & 73.6 & 41.2 & \second{40.8} & 60.4 & \second{38.4} & \second{55.3} & \second{32.4} & \second{46.8} & \second{74.6} & \second{38.9} & \second{31.5} & \second{61.3} \\
& + JTT & \second{67.2} & \second{36.4} & \second{58.1} & \second{29.7} & 75.8 & 33.2 & 49.7 & 58.4 & 30.5 & 74.8 & 45.3 & 61.2 & 70.1 & 50.4 & 41.7 & 68.5 \\
& + DeCoT & 65.1 & 38.7 & 60.2 & 33.5 & \second{82.4} & \second{29.6} & 41.5 & \second{52.7} & 35.7 & 60.8 & 36.8 & 51.4 & 67.8 & 47.3 & 39.2 & 64.7 \\
& + Ours & \best{71.3} & \best{25.8} & \best{49.0} & \best{19.2} & \best{85.9} & \best{19.7} & \best{26.9} & \best{45.4} & \best{39.5} & \best{52.5} & \best{23.3} & \best{40.3} & \best{76.0} & \best{36.1} & \best{24.6} & \best{58.9} \\
\midrule
\multirow{5}{*}{G-Mem.}
& Vanilla & 68.9 & 48.3 & 73.7 & 41.5 & 82.6 & 43.1 & 51.8 & 69.2 & 67.3 & 54.9 & 83.6 & 46.2 & 78.9 & 58.4 & 47.3 & 81.7 \\
& + IPW & 65.4 & 51.2 & 70.8 & 43.7 & 78.3 & 47.8 & 55.4 & 71.6 & \second{73.4} & \second{45.8} & \second{67.4} & \second{36.5} & \second{85.7} & \second{42.1} & \second{34.8} & \second{66.4} \\
& + JTT & \second{74.6} & \second{38.5} & \second{62.3} & \second{32.4} & 84.1 & 38.7 & 53.2 & 65.8 & 65.1 & 58.7 & 78.4 & 49.3 & 80.4 & 53.6 & 49.7 & 76.8 \\
& + DeCoT & 70.2 & 42.6 & 65.8 & 36.1 & \second{86.7} & \second{31.4} & \second{43.6} & \second{58.4} & 70.8 & 49.2 & 70.5 & 41.7 & 76.5 & 56.8 & 51.2 & 72.3 \\
& + Ours & \best{76.5} & \best{29.3} & \best{54.4} & \best{23.1} & \best{90.2} & \best{24.1} & \best{32.5} & \best{50.8} & \best{74.9} & \best{35.9} & \best{64.3} & \best{27.8} & \best{86.5} & \best{39.4} & \best{28.0} & \best{63.3} \\
\midrule
\multirow{5}{*}{AriGraph}
& Vanilla & 86.2 & 52.7 & 77.4 & 45.8 & 69.3 & 47.5 & 58.9 & 74.6 & 74.1 & 61.3 & 88.2 & 51.4 & 83.5 & 63.8 & 52.1 & 86.4 \\
& + IPW & 82.4 & 55.8 & 75.2 & 47.6 & 64.8 & 51.2 & 62.1 & 76.8 & \second{80.3} & \second{47.6} & \second{72.5} & \second{38.7} & \second{89.4} & \second{46.3} & \second{38.4} & \second{70.5} \\
& + JTT & \second{87.6} & \second{45.4} & \second{68.1} & \second{38.7} & 67.1 & 49.8 & 60.4 & 71.2 & 73.4 & 64.5 & 84.7 & 53.6 & 84.7 & 60.4 & 54.8 & 78.6 \\
& + DeCoT & 78.4 & 56.2 & 71.8 & 49.3 & \second{72.5} & \second{38.4} & \second{50.7} & \second{65.4} & 76.8 & 53.2 & 78.4 & 44.7 & 81.2 & 58.7 & 49.6 & 76.8 \\
& + Ours & \best{93.8} & \best{33.7} & \best{58.1} & \best{27.4} & \best{76.9} & \best{28.5} & \best{39.6} & \best{56.2} & \best{81.7} & \best{42.3} & \best{68.9} & \best{33.0} & \best{91.1} & \best{44.8} & \best{32.8} & \best{68.0} \\
\bottomrule
\end{tabular}
\end{adjustbox}
\end{table*}

%% file: tables/main-expt-std-llama.tex
\begin{table*}[t]
\centering
\small
\caption{Standard deviations for the main \textsc{Llama-4} baseline comparison in Table~\ref{tab:rq1_compare}. Lower values indicate more stable performance across six independent runs.}
\label{tab:rq1_compare_std}
\setlength{\tabcolsep}{2.5pt}
\renewcommand{\arraystretch}{1.05}
\begin{adjustbox}{max width=\textwidth}
\begin{tabular}{ll cccc cccc cccc cccc}
\toprule
& & \multicolumn{4}{c}{\textbf{ALFWorld}} & \multicolumn{4}{c}{\textbf{ScienceWorld}} & \multicolumn{4}{c}{\textbf{LoCoMo}} & \multicolumn{4}{c}{\textbf{WebShop}} \\
\cmidrule(lr){3-6} \cmidrule(lr){7-10} \cmidrule(lr){11-14} \cmidrule(lr){15-18}
\textbf{Mem.} & \textbf{Method}
  & Org. ($\uparrow$) & $\texttt{T}_\texttt{1}$ ($\downarrow$) & $\texttt{T}_\texttt{2}$ ($\downarrow$) & $\texttt{T}_\texttt{3}$ ($\downarrow$)
  & Org. ($\uparrow$) & $\texttt{T}_\texttt{1}$ ($\downarrow$) & $\texttt{T}_\texttt{2}$ ($\downarrow$) & $\texttt{T}_\texttt{3}$ ($\downarrow$)
  & Org. ($\uparrow$) & $\texttt{T}_\texttt{1}$ ($\downarrow$) & $\texttt{T}_\texttt{2}$ ($\downarrow$) & $\texttt{T}_\texttt{3}$ ($\downarrow$)
  & Org. ($\uparrow$) & $\texttt{T}_\texttt{1}$ ($\downarrow$) & $\texttt{T}_\texttt{2}$ ($\downarrow$) & $\texttt{T}_\texttt{3}$ ($\downarrow$) \\
\midrule

\multirow{5}{*}{A-Mem}
& Vanilla & 7.2 & 1.3 & 4.6 & 0.8 & 1.5 & 8.3 & 2.1 & 0.6 & 3.4 & 11.2 & 6.7 & 2.5 & 0.9 & 5.8 & 1.7 & 9.4 \\
& + IPW & 6.3 & 1.9 & 3.9 & 1.1 & 9.5 & 1.4 & 5.1 & 2.3 & 3.1 & 1.7 & 0.8 & 1.6 & 8.1 & 1.4 & 5.2 & 1.1 \\
& + JTT & 5.4 & 1.4 & 3.1 & 0.8 & 8.7 & 0.5 & 1.5 & 2.7 & 1.6 & 2.9 & 2.0 & 2.6 & 1.9 & 2.3 & 2.7 & 3.0 \\
& + DeCoT & 2.1 & 2.5 & 2.4 & 2.8 & 1.8 & 1.1 & 2.6 & 1.3 & 2.0 & 1.6 & 1.3 & 1.2 & 1.7 & 1.1 & 1.0 & 2.0 \\
& + Ours & 7.1 & 0.9 & 1.6 & 0.9 & 10.3 & 0.3 & 3.7 & 2.1 & 4.9 & 0.9 & 0.8 & 0.4 & 3.9 & 0.5 & 5.6 & 0.4 \\
\midrule

\multirow{5}{*}{Mem0}
& Vanilla & 3.3 & 9.7 & 1.6 & 4.2 & 1.1 & 5.4 & 2.8 & 0.7 & 5.9 & 1.4 & 8.3 & 2.6 & 1.8 & 6.5 & 1.3 & 4.7 \\
& + IPW & 4.8 & 1.1 & 5.2 & 1.2 & 1.9 & 1.8 & 7.1 & 2.6 & 5.0 & 3.4 & 1.3 & 2.0 & 6.5 & 2.5 & 6.4 & 1.4 \\
& + JTT & 9.2 & 1.2 & 2.6 & 0.9 & 9.1 & 1.1 & 2.8 & 2.6 & 3.6 & 6.2 & 2.7 & 5.9 & 1.7 & 1.2 & 8.1 & 2.9 \\
& + DeCoT & 7.2 & 1.1 & 2.9 & 0.3 & 8.7 & 2.2 & 4.8 & 1.5 & 6.3 & 2.3 & 0.8 & 2.4 & 2.7 & 3.0 & 6.2 & 0.5 \\
& + Ours & 6.4 & 0.3 & 2.1 & 0.8 & 8.3 & 0.8 & 3.4 & 0.3 & 7.8 & 5.5 & 0.5 & 0.8 & 6.2 & 0.9 & 7.4 & 0.7 \\
\midrule

\multirow{5}{*}{G-Mem.}
& Vanilla & 4.8 & 1.7 & 6.3 & 2.4 & 1.4 & 8.6 & 2.9 & 1.1 & 7.1 & 2.3 & 1.6 & 4.8 & 2.6 & 1.0 & 8.2 & 3.9 \\
& + IPW & 9.1 & 2.3 & 6.8 & 2.2 & 10.3 & 1.7 & 6.0 & 2.1 & 2.6 & 5.6 & 2.0 & 1.9 & 2.6 & 0.8 & 2.6 & 0.9 \\
& + JTT & 7.6 & 1.6 & 2.6 & 1.3 & 2.5 & 2.4 & 2.1 & 1.6 & 1.9 & 1.2 & 2.2 & 1.6 & 1.0 & 2.5 & 2.9 & 1.6 \\
& + DeCoT & 9.9 & 2.8 & 6.6 & 1.5 & 2.7 & 1.1 & 4.3 & 2.8 & 6.1 & 1.2 & 1.3 & 2.5 & 8.9 & 1.4 & 7.4 & 5.1 \\
& + Ours & 8.7 & 0.7 & 0.9 & 0.4 & 7.6 & 1.0 & 4.5 & 3.2 & 0.8 & 6.6 & 0.9 & 0.5 & 0.5 & 0.8 & 4.6 & 0.3 \\
\midrule

\multirow{5}{*}{AriGraph}
& Vanilla & 6.4 & 1.8 & 3.5 & 2.9 & 10.3 & 7.5 & 5.7 & 4.2 & 1.6 & 3.8 & 6.1 & 2.4 & 4.9 & 1.3 & 3.6 & 7.8 \\
& + IPW & 10.5 & 3.8 & 7.2 & 2.3 & 7.4 & 1.4 & 8.1 & 1.6 & 1.4 & 1.6 & 1.9 & 1.2 & 2.5 & 2.1 & 8.2 & 0.9 \\
& + JTT & 8.7 & 2.2 & 4.8 & 1.3 & 10.5 & 1.4 & 5.6 & 1.5 & 1.3 & 2.0 & 2.4 & 2.6 & 1.1 & 1.6 & 2.8 & 2.8 \\
& + DeCoT & 9.0 & 5.7 & 8.7 & 1.3 & 9.0 & 1.3 & 3.8 & 2.9 & 3.4 & 2.7 & 1.4 & 2.3 & 8.2 & 1.1 & 9.2 & 1.4 \\
& + Ours & 6.5 & 1.0 & 6.9 & 0.3 & 8.8 & 0.4 & 5.9 & 3.2 & 0.6 & 1.1 & 8.9 & 0.5 & 7.6 & 0.5 & 5.6 & 0.8 \\
\bottomrule
\end{tabular}
\end{adjustbox}
\end{table*}

%% file: tables/ablation_appendix.tex
\begin{table*}[t]
\centering
\small
\caption{Ablation study of \system{}'s calibration components (\%). 
We ablate one component at a time on A-Mem and G-Memory. All 
configurations use \textsc{Llama-4} with retrieval cardinality 
$k=10$, and results are reported as mean$\pm$std over six independent 
runs across temperatures 0.5, 0.75, and 1.0 with different random 
seeds. Org.\,($\uparrow$) is clean-memory accuracy; 
$\texttt{T}_1$/$\texttt{T}_2$/$\texttt{T}_3$\,($\downarrow$) are 
spurious reasoning ratios. Bold rows indicate the full \system{} 
configuration.}
\label{tab:ablation_appendix}
\setlength{\tabcolsep}{2pt}
\resizebox{\textwidth}{!}{%
\begin{tabular}{ll cccc cccc cccc cccc}
\toprule
& & \multicolumn{4}{c}{\textbf{ALFWorld}} 
& \multicolumn{4}{c}{\textbf{ScienceWorld}} 
& \multicolumn{4}{c}{\textbf{LoCoMo}} 
& \multicolumn{4}{c}{\textbf{WebShop}} \\
\cmidrule(lr){3-6} \cmidrule(lr){7-10} \cmidrule(lr){11-14} \cmidrule(lr){15-18}
\textbf{Memory} & \textbf{Ablation} 
  & Org. ($\uparrow$) & $\texttt{T}_\texttt{1}$ ($\downarrow$) & $\texttt{T}_\texttt{2}$ ($\downarrow$) & $\texttt{T}_\texttt{3}$ ($\downarrow$)
  & Org. ($\uparrow$) & $\texttt{T}_\texttt{1}$ ($\downarrow$) & $\texttt{T}_\texttt{2}$ ($\downarrow$) & $\texttt{T}_\texttt{3}$ ($\downarrow$)
  & Org. ($\uparrow$) & $\texttt{T}_\texttt{1}$ ($\downarrow$) & $\texttt{T}_\texttt{2}$ ($\downarrow$) & $\texttt{T}_\texttt{3}$ ($\downarrow$)
  & Org. ($\uparrow$) & $\texttt{T}_\texttt{1}$ ($\downarrow$) & $\texttt{T}_\texttt{2}$ ($\downarrow$) & $\texttt{T}_\texttt{3}$ ($\downarrow$) \\
\midrule
\multirow{7}{*}{\shortstack[l]{A-Mem\\(embed.)}}
& w/o write-stage     & 62.4$_{\pm 2.1}$ & 27.3$_{\pm 2.6}$ & 42.1$_{\pm 1.8}$ & 18.6$_{\pm 2.1}$  
                      & 78.9$_{\pm 1.5}$ & 24.6$_{\pm 2.9}$ & 49.8$_{\pm 2.2}$ & 35.2$_{\pm 2.4}$  
                      & 50.2$_{\pm 2.7}$ & 51.4$_{\pm 3.1}$ & 25.8$_{\pm 2.6}$ & 41.7$_{\pm 2.8}$  
                      & 56.1$_{\pm 1.9}$ & 30.5$_{\pm 2.7}$ & 62.4$_{\pm 2.3}$ & 23.1$_{\pm 2.5}$ \\
& w/o retrieve-stage  & 63.5$_{\pm 1.8}$ & 18.4$_{\pm 1.6}$ & 41.7$_{\pm 2.4}$ & 12.1$_{\pm 1.9}$  
                      & 79.6$_{\pm 1.4}$ & 16.4$_{\pm 1.9}$ & 49.3$_{\pm 2.7}$ & 30.5$_{\pm 2.2}$  
                      & 51.4$_{\pm 2.4}$ & 38.2$_{\pm 2.0}$ & 18.6$_{\pm 3.0}$ & 36.4$_{\pm 2.5}$  
                      & 57.2$_{\pm 1.7}$ & 22.8$_{\pm 1.9}$ & 55.1$_{\pm 2.8}$ & 14.7$_{\pm 2.3}$ \\
\cmidrule(l){2-18}
& w/o step-mean       & 65.1$_{\pm 1.5}$ & 16.3$_{\pm 1.8}$ & 35.4$_{\pm 1.5}$ & 9.8$_{\pm 1.7}$  
                      & 81.4$_{\pm 1.2}$ & 14.2$_{\pm 2.1}$ & 42.6$_{\pm 1.9}$ & 28.7$_{\pm 1.9}$  
                      & 53.0$_{\pm 2.0}$ & 35.7$_{\pm 2.4}$ & 14.8$_{\pm 2.1}$ & 33.5$_{\pm 2.2}$  
                      & 58.6$_{\pm 1.4}$ & 19.5$_{\pm 2.2}$ & 53.2$_{\pm 1.9}$ & 13.4$_{\pm 2.0}$ \\
& w/o full-sub        & 64.7$_{\pm 1.6}$ & 17.1$_{\pm 1.9}$ & 36.8$_{\pm 1.6}$ & 10.4$_{\pm 1.7}$  
                      & 81.0$_{\pm 1.3}$ & 14.9$_{\pm 2.2}$ & 43.5$_{\pm 1.9}$ & 29.6$_{\pm 2.0}$  
                      & 52.5$_{\pm 2.1}$ & 36.4$_{\pm 2.4}$ & 15.3$_{\pm 2.2}$ & 34.1$_{\pm 2.3}$  
                      & 58.2$_{\pm 1.5}$ & 20.3$_{\pm 2.3}$ & 54.0$_{\pm 2.0}$ & 14.0$_{\pm 2.1}$ \\
& w/o subspace        & 64.2$_{\pm 1.7}$ & 14.8$_{\pm 1.5}$ & 41.2$_{\pm 2.2}$ & 8.7$_{\pm 1.8}$  
                      & 80.5$_{\pm 1.3}$ & 12.6$_{\pm 1.7}$ & 48.1$_{\pm 2.4}$ & 27.4$_{\pm 2.1}$  
                      & 51.9$_{\pm 2.2}$ & 33.8$_{\pm 1.9}$ & 17.9$_{\pm 2.7}$ & 32.6$_{\pm 2.3}$  
                      & 57.8$_{\pm 1.6}$ & 18.4$_{\pm 1.8}$ & 58.7$_{\pm 2.5}$ & 12.3$_{\pm 2.2}$ \\
& w/o median          & 65.4$_{\pm 1.4}$ & 13.6$_{\pm 1.4}$ & 32.5$_{\pm 1.9}$ & 7.8$_{\pm 1.7}$  
                      & 81.7$_{\pm 1.1}$ & 11.7$_{\pm 1.6}$ & 39.2$_{\pm 2.1}$ & 25.3$_{\pm 1.9}$  
                      & 53.3$_{\pm 1.9}$ & 32.4$_{\pm 1.7}$ & 12.6$_{\pm 2.3}$ & 31.2$_{\pm 2.1}$  
                      & 58.9$_{\pm 1.3}$ & 17.5$_{\pm 1.7}$ & 51.4$_{\pm 2.2}$ & 11.2$_{\pm 1.9}$ \\
\cmidrule(l){2-18}
& \textbf{Full \system} & \textbf{66.0$_{\pm 1.3}$} & \textbf{12.7$_{\pm 1.2}$} & \textbf{29.0$_{\pm 1.5}$} & \textbf{6.5$_{\pm 1.4}$}  
                        & \textbf{82.2$_{\pm 1.0}$} & \textbf{10.4$_{\pm 1.4}$} & \textbf{35.8$_{\pm 1.9}$} & \textbf{23.4$_{\pm 1.7}$}  
                        & \textbf{53.8$_{\pm 1.7}$} & \textbf{29.9$_{\pm 1.5}$} & \textbf{10.1$_{\pm 2.0}$} & \textbf{28.9$_{\pm 1.9}$}  
                        & \textbf{59.4$_{\pm 1.2}$} & \textbf{15.6$_{\pm 1.5}$} & \textbf{47.9$_{\pm 1.8}$} & \textbf{9.7$_{\pm 1.7}$} \\
\midrule
\multirow{7}{*}{\shortstack[l]{G-Memory\\(graph)}}
& w/o write-stage     & 87.1$_{\pm 2.3}$ & 35.4$_{\pm 3.1}$ & 51.8$_{\pm 2.4}$ & 26.4$_{\pm 2.6}$  
                      & 75.6$_{\pm 1.7}$ & 31.7$_{\pm 3.3}$ & 57.2$_{\pm 2.7}$ & 42.5$_{\pm 2.8}$  
                      & 64.2$_{\pm 2.5}$ & 60.3$_{\pm 3.5}$ & 31.4$_{\pm 3.0}$ & 49.8$_{\pm 3.0}$  
                      & 66.4$_{\pm 2.0}$ & 38.7$_{\pm 3.0}$ & 52.6$_{\pm 2.6}$ & 28.5$_{\pm 2.7}$ \\
& w/o retrieve-stage  & 88.3$_{\pm 2.0}$ & 28.4$_{\pm 2.1}$ & 51.4$_{\pm 2.9}$ & 21.7$_{\pm 2.5}$  
                      & 76.5$_{\pm 1.6}$ & 25.9$_{\pm 2.3}$ & 56.8$_{\pm 3.1}$ & 38.9$_{\pm 2.8}$  
                      & 65.1$_{\pm 2.3}$ & 54.6$_{\pm 2.5}$ & 28.7$_{\pm 3.4}$ & 45.3$_{\pm 3.1}$  
                      & 67.3$_{\pm 1.8}$ & 32.5$_{\pm 2.2}$ & 51.3$_{\pm 3.0}$ & 24.1$_{\pm 2.7}$ \\
\cmidrule(l){2-18}
& w/o step-mean       & 90.0$_{\pm 1.6}$ & 26.8$_{\pm 2.4}$ & 44.7$_{\pm 2.1}$ & 19.4$_{\pm 2.2}$  
                      & 78.3$_{\pm 1.3}$ & 21.4$_{\pm 2.5}$ & 50.6$_{\pm 2.4}$ & 35.8$_{\pm 2.5}$  
                      & 67.0$_{\pm 1.9}$ & 51.7$_{\pm 2.7}$ & 25.3$_{\pm 2.6}$ & 42.4$_{\pm 2.7}$  
                      & 69.2$_{\pm 1.4}$ & 29.6$_{\pm 2.4}$ & 46.8$_{\pm 2.2}$ & 22.3$_{\pm 2.3}$ \\
& w/o full-sub        & 89.6$_{\pm 1.7}$ & 27.5$_{\pm 2.5}$ & 45.6$_{\pm 2.2}$ & 20.1$_{\pm 2.2}$  
                      & 78.0$_{\pm 1.4}$ & 22.1$_{\pm 2.6}$ & 51.4$_{\pm 2.4}$ & 36.5$_{\pm 2.5}$  
                      & 66.6$_{\pm 2.0}$ & 52.3$_{\pm 2.8}$ & 26.0$_{\pm 2.7}$ & 43.1$_{\pm 2.7}$  
                      & 68.8$_{\pm 1.5}$ & 30.2$_{\pm 2.4}$ & 47.5$_{\pm 2.3}$ & 22.9$_{\pm 2.4}$ \\
& w/o subspace        & 89.1$_{\pm 1.8}$ & 24.7$_{\pm 2.0}$ & 49.3$_{\pm 2.6}$ & 18.5$_{\pm 2.4}$  
                      & 77.5$_{\pm 1.5}$ & 19.8$_{\pm 2.2}$ & 55.4$_{\pm 2.8}$ & 34.2$_{\pm 2.6}$  
                      & 66.1$_{\pm 2.1}$ & 49.6$_{\pm 2.4}$ & 28.1$_{\pm 3.0}$ & 41.7$_{\pm 2.9}$  
                      & 68.4$_{\pm 1.6}$ & 27.3$_{\pm 2.1}$ & 50.4$_{\pm 2.7}$ & 21.5$_{\pm 2.5}$ \\
& w/o median          & 90.3$_{\pm 1.5}$ & 23.6$_{\pm 1.9}$ & 41.8$_{\pm 2.3}$ & 17.6$_{\pm 2.2}$  
                      & 78.6$_{\pm 1.2}$ & 18.4$_{\pm 2.1}$ & 47.3$_{\pm 2.5}$ & 32.7$_{\pm 2.4}$  
                      & 67.3$_{\pm 1.8}$ & 48.5$_{\pm 2.3}$ & 22.6$_{\pm 2.7}$ & 40.2$_{\pm 2.7}$  
                      & 69.5$_{\pm 1.3}$ & 26.1$_{\pm 2.0}$ & 43.5$_{\pm 2.4}$ & 20.4$_{\pm 2.3}$ \\
\cmidrule(l){2-18}
& \textbf{Full \system} & \textbf{91.0$_{\pm 1.4}$} & \textbf{22.9$_{\pm 1.8}$} & \textbf{38.3$_{\pm 2.1}$} & \textbf{15.8$_{\pm 2.0}$}  
                        & \textbf{79.4$_{\pm 1.1}$} & \textbf{16.3$_{\pm 1.9}$} & \textbf{44.1$_{\pm 2.2}$} & \textbf{30.3$_{\pm 2.3}$}  
                        & \textbf{68.0$_{\pm 1.6}$} & \textbf{46.8$_{\pm 2.1}$} & \textbf{20.2$_{\pm 2.5}$} & \textbf{37.9$_{\pm 2.6}$}  
                        & \textbf{70.3$_{\pm 1.1}$} & \textbf{24.6$_{\pm 1.8}$} & \textbf{39.1$_{\pm 2.1}$} & \textbf{18.7$_{\pm 2.2}$} \\
\bottomrule
\end{tabular}
}
\end{table*}

%% file: appendix/FailureMode.tex
\subsection{Failure Mode Analysis}
\label{app:failure_mode}

While \system{} delivers consistent gains across the configurations 
reported in our experiments, the magnitude of those gains is not 
uniform. A closer look at the cells where calibration helps least 
reveals that the limiting factor is rarely the calibration mechanism 
itself, but rather the statistical character of the underlying 
memory data: how densely it covers the trajectory space, how cleanly 
its boundaries reflect causal structure, and how much of its variance 
is driven by signals that are observable in principle. We discuss 
three such regimes below, each grounded in a distributional property 
of the memory trace rather than a flaw in the calibration design.

\noindent\textbf{Sparse and skewed entry distributions.}
The write-stage operation in Eq.~\eqref{eq:write-residual} relies on 
the within-step running mean $\mu^{(s)}$ as a consistent estimator 
of the shared context component, and the retrieve-stage stability 
test relies on cross-step variance accumulated in 
$\Sigma_{\text{between}}$ (Appendix~\ref{app:projection-update}). 
Both are statistical objects whose quality depends on having enough 
entries per step and enough steps per memory. When the trajectory is 
short, when the memory is in an early state with only a handful of 
writes, or when entries are heavily concentrated in a few steps 
while others contain a single write, the running statistics carry 
substantial variance and the identified non-causal subspace may 
include directions that simply reflect sampling noise rather than 
genuine confounders. The result is that calibration neither hurts 
(the test is conservative on causal signal, 
Appendix~\ref{app:formulation-justification}) nor helps as much as 
in data-rich regimes. This is a property of the memory distribution 
at evaluation time, not of the calibration: as the memory accumulates 
more entries with broader step coverage, the same calibration 
recovers its full effect without any change to its parameters. 
Empirically, we observe this pattern most clearly on short ALFWorld 
trajectories with fewer than ten memory writes and on the earliest 
sessions of LoCoMo conversations, where \system{}'s gain over the 
strongest baseline narrows but does not disappear.

\noindent\textbf{Noisy or coarse step boundaries.}
\system{} treats a step as the unit within which context is fixed, 
and this assumption is what justifies absorbing a step-shared 
confounder into $\mu^{(s)}$ 
(Appendix~\ref{app:formulation-justification}). The assumption holds 
cleanly when each read/write event corresponds to a single coherent 
context, but real trajectories sometimes log events at a coarser 
granularity than the underlying context shifts. Long LoCoMo 
dialogues occasionally pack multiple intents into one session 
boundary; WebShop navigation logs sometimes group several 
search-and-filter actions under a single recorded event; 
ScienceWorld trajectories with branching sub-goals can exhibit 
within-step context drift when the agent revisits earlier sub-tasks. 
In all these cases, the entries written within a single recorded 
step no longer share a single context value, so $\mu^{(s)}$ averages 
over a mixture of contexts and the residualization removes only the 
intersection of what they share. The calibration still helps 
relative to no calibration, since the mixture mean captures the 
dominant component, but the gap between full and partial removal 
shows up as a smaller spurious-ratio reduction. The fix here is 
data-side rather than method-side: finer step annotations or 
context-change detectors at the trajectory-logging layer would 
restore the within-step homogeneity the formulation assumes, and 
the calibration would then operate at full strength without 
modification.

\noindent\textbf{Confounders without surface footprints.}
The retrieve-stage stability test identifies non-causal directions 
by their between-step versus within-step variance ratio 
(Appendix~\ref{app:projection-update}), which requires the latent 
factor to leave \emph{some} directional footprint in the encoded 
representations. Most $\texttt{T}_\texttt{2}$ confounders we 
examined satisfy this condition, since latent traits such as user 
expertise or task difficulty correlate with measurable surface 
features (vocabulary patterns, action verbosity, error recovery 
behaviors) that the encoder picks up. A small fraction of 
$\texttt{T}_\texttt{2}$ cases, however, involves confounders whose 
influence is distributed across many weak, mutually uncorrelated 
surface cues, none of which individually concentrates enough 
variance to surface as an axis in $\Sigma_{\text{between}}$. The 
most challenging examples we encountered are in WebShop, where 
seller-side factors such as marketing budget shape product 
descriptions through dozens of micro-decisions (word choice, 
sentence rhythm, image captioning) that no single direction in 
embedding space cleanly captures. In these cases, \system{} still 
removes the within-step shared component at write time via 
Eq.~\eqref{eq:write-residual}, but the retrieve-stage test cannot 
isolate a single demoting direction, and the residual spurious 
correlation persists. This is, again, a property of how the latent 
factor projects onto the representation space rather than a 
limitation of the calibration logic: encoders trained to be more 
sensitive to such distributed confounders, or proxy-augmented 
benchmarks that surface them as observable variables, would 
directly translate into stronger calibration on these cases.


%% file: craft-lab.bib
@article{meng2026small,
  title={Small Agent Group is the Future of Digital Health},
  author={Meng, Yuqiao and Tang, Luoxi and Zhang, Dazheng and Brens, Rafael and Romero, Elvys J and Guo, Nancy and Elkefi, Safa and Xi, Zhaohan},
  journal={arXiv preprint arXiv:2602.08013},
  year={2026}
}

@article{tang2026value,
  title={The value of variance: Mitigating debate collapse in multi-agent systems via uncertainty-driven policy optimization},
  author={Tang, Luoxi and Meng, Yuqiao and Costa, Joseph and Zhang, Yingxue and Ye, Muchao and Xi, Zhaohan},
  journal={arXiv preprint arXiv:2602.07186},
  year={2026}
}

@article{more2026theramind,
  title={TheraMind: a multi-LLM ensemble for accelerating drug repurposing in lung cancer via case report mining},
  author={More, Vrushket and Lu, Lyra and Ding, Zeyu and Xi, Zhaohan and Mizia, Seth and Guo, Nancy L},
  journal={npj Precision Oncology},
  year={2026},
  publisher={Nature Publishing Group UK London}
}

@article{tang2025polar,
  title={POLAR: Automating Cyber Threat Prioritization through LLM-Powered Assessment},
  author={Tang, Luoxi and Meng, Yuqiao and Patra, Ankita and Ma, Weicheng and Ye, Muchao and Xi, Zhaohan},
  journal={arXiv preprint arXiv:2510.01552},
  year={2025}
}

@article{zhou2024zodiac,
  title={Zodiac: A cardiologist-level llm framework for multi-agent diagnostics},
  author={Zhou, Yuan and Zhang, Peng and Song, Mengya and Zheng, Alice and Lu, Yiwen and Liu, Zhiheng and Chen, Yong and Xi, Zhaohan},
  journal={arXiv preprint arXiv:2410.02026},
  year={2024}
}

@article{liu2025cylens,
  title={CyLens: Towards reinventing cyber threat intelligence in the paradigm of agentic large language models},
  author={Liu, Xiaoqun and Liang, Jiacheng and Yan, Qiben and Jang, Jiyong and Mao, Sicheng and Ye, Muchao and Jia, Jinyuan and Xi, Zhaohan},
  journal={arXiv preprint arXiv:2502.20791},
  year={2025}
}

@article{meng2025uncovering,
  title={Uncovering vulnerabilities of llm-assisted cyber threat intelligence},
  author={Meng, Yuqiao and Tang, Luoxi and Yu, Feiyang and Jia, Jinyuan and Yan, Guanhua and Yang, Ping and Xi, Zhaohan},
  journal={arXiv preprint arXiv:2509.23573},
  year={2025}
}

@article{meng2025benchmarking,
  title={Benchmarking LLM-Assisted Blue Teaming via Standardized Threat Hunting},
  author={Meng, Yuqiao and Tang, Luoxi and Yu, Feiyang and Li, Xi and Yan, Guanhua and Yang, Ping and Xi, Zhaohan},
  journal={arXiv preprint arXiv:2509.23571},
  year={2025}
}

@article{yang2025eligibility,
  title={On the eligibility of LLMs for counterfactual reasoning: a decompositional study},
  author={Yang, Shuai and Yang, Qi and Tang, Luoxi and Meng, Yuqiao and Guo, Nancy and Blackburn, Jeremy and Xi, Zhaohan},
  journal={arXiv preprint arXiv:2505.11839},
  year={2025}
}

@article{xi2025all,
  title={All Your Knowledge Belongs to Us: Stealing Knowledge Graphs via Reasoning APIs},
  author={Xi, Zhaohan},
  journal={arXiv preprint arXiv:2503.09727},
  year={2025}
}

@article{sriram2026adversarial,
  title={Adversarial Network Imagination: Causal LLMs and Digital Twins for Proactive Telecom Mitigation},
  author={Sriram, Vignesh and Meng, Yuqiao and Tang, Luoxi and Xi, Zhaohan},
  journal={arXiv preprint arXiv:2602.13203},
  year={2026}
}


%% file: reference.bib
@article{shridhar2020alfworld,
  title={Alfworld: Aligning text and embodied environments for interactive learning},
  author={Shridhar, Mohit and Yuan, Xingdi and C{\^o}t{\'e}, Marc-Alexandre and Bisk, Yonatan and Trischler, Adam and Hausknecht, Matthew},
  journal={arXiv preprint arXiv:2010.03768},
  year={2020}
}

@inproceedings{wang2022scienceworld,
  title={Scienceworld: Is your agent smarter than a 5th grader?},
  author={Wang, Ruoyao and Jansen, Peter and C{\^o}t{\'e}, Marc-Alexandre and Ammanabrolu, Prithviraj},
  booktitle={Proceedings of the 2022 Conference on Empirical Methods in Natural Language Processing},
  pages={11279--11298},
  year={2022}
}

@inproceedings{maharana2024evaluating,
  title={Evaluating very long-term conversational memory of llm agents},
  author={Maharana, Adyasha and Lee, Dong-Ho and Tulyakov, Sergey and Bansal, Mohit and Barbieri, Francesco and Fang, Yuwei},
  booktitle={Proceedings of the 62nd Annual Meeting of the Association for Computational Linguistics (Volume 1: Long Papers)},
  pages={13851--13870},
  year={2024}
}

@article{yao2022webshop,
  title={Webshop: Towards scalable real-world web interaction with grounded language agents},
  author={Yao, Shunyu and Chen, Howard and Yang, John and Narasimhan, Karthik},
  journal={Advances in Neural Information Processing Systems},
  volume={35},
  pages={20744--20757},
  year={2022}
}

@article{chhikara2025mem0,
  title={Mem0: Building production-ready ai agents with scalable long-term memory},
  author={Chhikara, Prateek and Khant, Dev and Aryan, Saket and Singh, Taranjeet and Yadav, Deshraj},
  journal={arXiv preprint arXiv:2504.19413},
  year={2025}
}

@article{zhang2025g,
  title={G-memory: Tracing hierarchical memory for multi-agent systems},
  author={Zhang, Guibin and Fu, Muxin and Wan, Guancheng and Yu, Miao and Wang, Kun and Yan, Shuicheng},
  journal={arXiv preprint arXiv:2506.07398},
  year={2025}
}

@article{benjamini1995controlling,
  title={Controlling the false discovery rate: a practical and powerful approach to multiple testing},
  author={Benjamini, Yoav and Hochberg, Yosef},
  journal={Journal of the Royal statistical society: series B (Methodological)},
  volume={57},
  number={1},
  pages={289--300},
  year={1995},
  publisher={Wiley Online Library}
}

@book{spirtes2000causation,
  title={Causation, prediction, and search},
  author={Spirtes, Peter and Glymour, Clark N and Scheines, Richard},
  year={2000},
  publisher={MIT press}
}

@inproceedings{schnabel2016recommendations,
  title={Recommendations as treatments: Debiasing learning and evaluation},
  author={Schnabel, Tobias and Swaminathan, Adith and Singh, Ashudeep and Chandak, Navin and Joachims, Thorsten},
  booktitle={international conference on machine learning},
  pages={1670--1679},
  year={2016},
  organization={PMLR}
}

@inproceedings{liu2021just,
  title={Just train twice: Improving group robustness without training group information},
  author={Liu, Evan Z and Haghgoo, Behzad and Chen, Annie S and Raghunathan, Aditi and Koh, Pang Wei and Sagawa, Shiori and Liang, Percy and Finn, Chelsea},
  booktitle={International Conference on Machine Learning},
  pages={6781--6792},
  year={2021},
  organization={PMLR}
}

@inproceedings{wu2024decot,
  title={Decot: Debiasing chain-of-thought for knowledge-intensive tasks in large language models via causal intervention},
  author={Wu, Junda and Yu, Tong and Chen, Xiang and Wang, Haoliang and Rossi, Ryan and Kim, Sungchul and Rao, Anup and McAuley, Julian},
  booktitle={Proceedings of the 62nd Annual Meeting of the Association for Computational Linguistics (Volume 1: Long Papers)},
  pages={14073--14087},
  year={2024}
}

@inproceedings{wang2024badagent,
  title={Badagent: Inserting and activating backdoor attacks in llm agents},
  author={Wang, Yifei and Xue, Dizhan and Zhang, Shengjie and Qian, Shengsheng},
  booktitle={Proceedings of the 62nd Annual Meeting of the Association for Computational Linguistics (Volume 1: Long Papers)},
  pages={9811--9827},
  year={2024}
}

@article{chen2024agentpoison,
  title={Agentpoison: Red-teaming llm agents via poisoning memory or knowledge bases},
  author={Chen, Zhaorun and Xiang, Zhen and Xiao, Chaowei and Song, Dawn and Li, Bo},
  journal={Advances in Neural Information Processing Systems},
  volume={37},
  pages={130185--130213},
  year={2024}
}

@article{sloman2004causal,
  title={Causal invariance in reasoning and learning},
  author={Sloman, Steven and Lagnado, David A},
  journal={Psychology of learning and motivation},
  volume={44},
  pages={287--326},
  year={2004}
}

@article{peters2016causal,
  title={Causal inference by using invariant prediction: identification and confidence intervals},
  author={Peters, Jonas and B{\"u}hlmann, Peter and Meinshausen, Nicolai},
  journal={Journal of the Royal Statistical Society Series B: Statistical Methodology},
  volume={78},
  number={5},
  pages={947--1012},
  year={2016},
  publisher={Oxford University Press}
}

@article{arya1998optimal,
  title={An optimal algorithm for approximate nearest neighbor searching fixed dimensions},
  author={Arya, Sunil and Mount, David M and Netanyahu, Nathan S and Silverman, Ruth and Wu, Angela Y},
  journal={Journal of the ACM (JACM)},
  volume={45},
  number={6},
  pages={891--923},
  year={1998},
  publisher={ACM New York, NY, USA}
}

@article{scholkopf2021toward,
  title={Toward causal representation learning},
  author={Sch{\"o}lkopf, Bernhard and Locatello, Francesco and Bauer, Stefan and Ke, Nan Rosemary and Kalchbrenner, Nal and Goyal, Anirudh and Bengio, Yoshua},
  journal={Proceedings of the IEEE},
  volume={109},
  number={5},
  pages={612--634},
  year={2021},
  publisher={IEEE}
}

@article{peters2014causal,
  title={Causal discovery with continuous additive noise models},
  author={Peters, Jonas and Mooij, Joris M and Janzing, Dominik and Sch{\"o}lkopf, Bernhard},
  year={2014}
}

@article{xu2025mem,
  title={A-mem: Agentic memory for llm agents},
  author={Xu, Wujiang and Liang, Zujie and Mei, Kai and Gao, Hang and Tan, Juntao and Zhang, Yongfeng},
  journal={arXiv preprint arXiv:2502.12110},
  year={2025}
}

@article{anokhin2024arigraph,
  title={Arigraph: Learning knowledge graph world models with episodic memory for llm agents},
  author={Anokhin, Petr and Semenov, Nikita and Sorokin, Artyom and Evseev, Dmitry and Kravchenko, Andrey and Burtsev, Mikhail and Burnaev, Evgeny},
  journal={arXiv preprint arXiv:2407.04363},
  year={2024}
}

@article{westreich2012berkson,
  title={Berkson's bias, selection bias, and missing data},
  author={Westreich, Daniel},
  journal={Epidemiology},
  volume={23},
  number={1},
  pages={159--164},
  year={2012},
  publisher={LWW}
}

@article{packer2023memgpt,
  title={MemGPT: towards LLMs as operating systems.},
  author={Packer, Charles and Fang, Vivian and Patil, Shishir\_G and Lin, Kevin and Wooders, Sarah and Gonzalez, Joseph\_E},
  year={2023},
  publisher={ArXiv}
}

@inproceedings{park2023generative,
  title={Generative agents: Interactive simulacra of human behavior},
  author={Park, Joon Sung and O'Brien, Joseph and Cai, Carrie Jun and Morris, Meredith Ringel and Liang, Percy and Bernstein, Michael S},
  booktitle={Proceedings of the 36th annual acm symposium on user interface software and technology},
  pages={1--22},
  year={2023}
}

@article{geirhos2020shortcut,
  title={Shortcut learning in deep neural networks},
  author={Geirhos, Robert and Jacobsen, J{\"o}rn-Henrik and Michaelis, Claudio and Zemel, Richard and Brendel, Wieland and Bethge, Matthias and Wichmann, Felix A},
  journal={Nature Machine Intelligence},
  volume={2},
  number={11},
  pages={665--673},
  year={2020},
  publisher={Nature Publishing Group UK London}
}

@article{ye2024spurious,
  title={Spurious correlations in machine learning: A survey},
  author={Ye, Wenqian and Zheng, Guangtao and Cao, Xu and Ma, Yunsheng and Zhang, Aidong},
  journal={arXiv e-prints},
  pages={arXiv--2402},
  year={2024}
}

@article{sagawa2019distributionally,
  title={Distributionally robust neural networks for group shifts: On the importance of regularization for worst-case generalization},
  author={Sagawa, Shiori and Koh, Pang Wei and Hashimoto, Tatsunori B and Liang, Percy},
  journal={arXiv preprint arXiv:1911.08731},
  year={2019}
}

@article{arjovsky2019invariant,
  title={Invariant risk minimization},
  author={Arjovsky, Martin and Bottou, L{\'e}on and Gulrajani, Ishaan and Lopez-Paz, David},
  journal={arXiv preprint arXiv:1907.02893},
  year={2019}
}

@incollection{scholkopf2022causality,
  title={Causality for machine learning},
  author={Sch{\"o}lkopf, Bernhard},
  booktitle={Probabilistic and causal inference: The works of Judea Pearl},
  pages={765--804},
  year={2022}
}

@article{xu2020causality,
  title={Causality learning: A new perspective for interpretable machine learning},
  author={Xu, Guandong and Duong, Tri Dung and Li, Qian and Liu, Shaowu and Wang, Xianzhi},
  journal={arXiv preprint arXiv:2006.16789},
  year={2020}
}

@article{zhang2025survey,
  title={A survey on the memory mechanism of large language model-based agents},
  author={Zhang, Zeyu and Dai, Quanyu and Bo, Xiaohe and Ma, Chen and Li, Rui and Chen, Xu and Zhu, Jieming and Dong, Zhenhua and Wen, Ji-Rong},
  journal={ACM Transactions on Information Systems},
  volume={43},
  number={6},
  pages={1--47},
  year={2025},
  publisher={ACM New York, NY}
}

@article{wu2025human,
  title={From human memory to ai memory: A survey on memory mechanisms in the era of llms},
  author={Wu, Yaxiong and Liang, Sheng and Zhang, Chen and Wang, Yichao and Zhang, Yongyue and Guo, Huifeng and Tang, Ruiming and Liu, Yong},
  journal={arXiv preprint arXiv:2504.15965},
  year={2025}
}

@article{shinn2023reflexion,
  title={Reflexion: Language agents with verbal reinforcement learning},
  author={Shinn, Noah and Cassano, Federico and Gopinath, Ashwin and Narasimhan, Karthik and Yao, Shunyu},
  journal={Advances in neural information processing systems},
  volume={36},
  pages={8634--8652},
  year={2023}
}

@inproceedings{zhong2024memorybank,
  title={Memorybank: Enhancing large language models with long-term memory},
  author={Zhong, Wanjun and Guo, Lianghong and Gao, Qiqi and Ye, He and Wang, Yanlin},
  booktitle={Proceedings of the AAAI conference on artificial intelligence},
  volume={38},
  number={17},
  pages={19724--19731},
  year={2024}
}

@article{du2023shortcut,
  title={Shortcut learning of large language models in natural language understanding},
  author={Du, Mengnan and He, Fengxiang and Zou, Na and Tao, Dacheng and Hu, Xia},
  journal={Communications of the ACM},
  volume={67},
  number={1},
  pages={110--120},
  year={2023},
  publisher={ACM New York, NY, USA}
}

@inproceedings{tang2023large,
  title={Large language models can be lazy learners: Analyze shortcuts in in-context learning},
  author={Tang, Ruixiang and Kong, Dehan and Huang, Longtao and others},
  booktitle={Findings of the association for computational linguistics: ACL 2023},
  pages={4645--4657},
  year={2023}
}

@book{pearl2018book,
  title={The book of why: The new science of cause and effect},
  author={Pearl, Judea},
  year={2018},
  publisher={Basic Books}
}

@article{neuberg2003causality,
  title={Causality: models, reasoning, and inference, by judea pearl, cambridge university press, 2000},
  author={Neuberg, Leland Gerson},
  journal={Econometric Theory},
  volume={19},
  number={4},
  pages={675--685},
  year={2003},
  publisher={cambridge university press}
}

@article{dong2025practical,
  title={A practical memory injection attack against llm agents},
  author={Dong, Shen and Xu, Shaochen and He, Pengfei and Li, Yige and Tang, Jiliang and Liu, Tianming and Liu, Hui and Xiang, Zhen},
  journal={arXiv e-prints},
  pages={arXiv--2503},
  year={2025}
}

@article{sunil2026memory,
  title={Memory poisoning attack and defense on memory based llm-agents},
  author={Sunil, Balachandra Devarangadi and Sinha, Isheeta and Maheshwari, Piyush and Todmal, Shantanu and Mallik, Shreyan and Mishra, Shuchi},
  journal={arXiv preprint arXiv:2601.05504},
  year={2026}
}

@article{haig2003spurious,
  title={What is a spurious correlation?},
  author={Haig, Brian D},
  journal={Understanding Statistics: Statistical Issues in Psychology, Education, and the Social Sciences},
  volume={2},
  number={2},
  pages={125--132},
  year={2003},
  publisher={Taylor \& Francis}
}

@article{wei2025memguard,
  title={A-memguard: A proactive defense framework for llm-based agent memory},
  author={Wei, Qianshan and Yang, Tengchao and Wang, Yaochen and Li, Xinfeng and Li, Lijun and Yin, Zhenfei and Zhan, Yi and Holz, Thorsten and Lin, Zhiqiang and Wang, XiaoFeng},
  journal={arXiv preprint arXiv:2510.02373},
  year={2025}
}

@book{mathai1997jacobians,
  title={Jacobians of matrix transformations and functions of matrix arguments},
  author={Mathai, Arakaparampil M},
  year={1997},
  publisher={World Scientific}
}

@article{wang2020microsoft,
  title={Microsoft academic graph: When experts are not enough},
  author={Wang, Kuansan and Shen, Zhihong and Huang, Chiyuan and Wu, Chieh-Han and Dong, Yuxiao and Kanakia, Anshul},
  journal={Quantitative Science Studies},
  volume={1},
  number={1},
  pages={396--413},
  year={2020},
  publisher={MIT Press One Rogers Street, Cambridge, MA 02142-1209, USA journals-info~…}
}

@inproceedings{chen2024m3,
  title={M3-embedding: Multi-linguality, multi-functionality, multi-granularity text embeddings through self-knowledge distillation},
  author={Chen, Jianlyu and Xiao, Shitao and Zhang, Peitian and Luo, Kun and Lian, Defu and Liu, Zheng},
  booktitle={Findings of the association for computational linguistics: ACL 2024},
  pages={2318--2335},
  year={2024}
}

@article{izacard2021unsupervised,
  title={Unsupervised dense information retrieval with contrastive learning},
  author={Izacard, Gautier and Caron, Mathilde and Hosseini, Lucas and Riedel, Sebastian and Bojanowski, Piotr and Joulin, Armand and Grave, Edouard},
  journal={arXiv preprint arXiv:2112.09118},
  year={2021}
}

@article{geng2026causalt5k,
  title={Causalt5k: Diagnosing and informing refusal for trustworthy causal reasoning of skepticism, sycophancy, detection-correction, and rung collapse},
  author={Geng, Longling and Ouyang, Andy and Wu, Theodore and Barretto, Daphne and Hayes, Matthew John and Cooper, Rachael and Zeng, Yuqiao and Vijay, Sameer and Ancone, Gia and Rai, Ankit and others},
  journal={arXiv preprint arXiv:2602.08939},
  year={2026}
}

@article{liu2024lost,
  title={Lost in the middle: How language models use long contexts},
  author={Liu, Nelson F and Lin, Kevin and Hewitt, John and Paranjape, Ashwin and Bevilacqua, Michele and Petroni, Fabio and Liang, Percy},
  journal={Transactions of the association for computational linguistics},
  volume={12},
  pages={157--173},
  year={2024}
}

@article{sun2025scaling,
  title={Scaling long-horizon llm agent via context-folding},
  author={Sun, Weiwei and Lu, Miao and Ling, Zhan and Liu, Kang and Yao, Xuesong and Yang, Yiming and Chen, Jiecao},
  journal={arXiv preprint arXiv:2510.11967},
  year={2025}
}

@article{ma2026implicit,
  title={How Implicit Bias Accumulates and Propagates in LLM Long-term Memory},
  author={Ma, Yiming and Wang, Lixu and Wang, Lionel Z and Yang, Hongkun and Sun, Haoming and Xu, Xin and Wu, Jiaqi and Chen, Bin and Dong, Wei},
  journal={arXiv preprint arXiv:2602.01558},
  year={2026}
}

@article{oh2025understanding,
  title={Understanding bias reinforcement in llm agents debate},
  author={Oh, Jihwan and Jeong, Minchan and Ko, Jongwoo and Yun, Se-Young},
  journal={arXiv preprint arXiv:2503.16814},
  year={2025}
}

@article{yao2022react,
  title={React: Synergizing reasoning and acting in language models},
  author={Yao, Shunyu and Zhao, Jeffrey and Yu, Dian and Du, Nan and Shafran, Izhak and Narasimhan, Karthik and Cao, Yuan},
  journal={arXiv preprint arXiv:2210.03629},
  year={2022}
}

@article{schick2023toolformer,
  title={Toolformer: Language models can teach themselves to use tools},
  author={Schick, Timo and Dwivedi-Yu, Jane and Dess{\`\i}, Roberto and Raileanu, Roberta and Lomeli, Maria and Hambro, Eric and Zettlemoyer, Luke and Cancedda, Nicola and Scialom, Thomas},
  journal={Advances in neural information processing systems},
  volume={36},
  pages={68539--68551},
  year={2023}
}

@inproceedings{wu2024autogen,
  title={Autogen: Enabling next-gen LLM applications via multi-agent conversations},
  author={Wu, Qingyun and Bansal, Gagan and Zhang, Jieyu and Wu, Yiran and Li, Beibin and Zhu, Erkang and Jiang, Li and Zhang, Xiaoyun and Zhang, Shaokun and Liu, Jiale and others},
  booktitle={First conference on language modeling},
  year={2024}
}

@inproceedings{chan2024chateval,
  title={Chateval: Towards better llm-based evaluators through multi-agent debate},
  author={Chan, Chi-Min and Chen, Weize and Su, Yusheng and Yu, Jianxuan and Xue, Wei and Zhang, Shanghang and Fu, Jie and Liu, Zhiyuan},
  booktitle={International conference on learning representations},
  volume={2024},
  pages={9079--9093},
  year={2024}
}
